\documentclass[10pt,twocolumn,letterpaper]{article}

\usepackage[pagenumbers]{cvpr} 
\usepackage[ruled,vlined,linesnumbered]{algorithm2e}
\definecolor{cvprblue}{rgb}{0.21,0.49,0.74}
\usepackage[pagebackref,breaklinks,colorlinks,allcolors=cvprblue]{hyperref}
\usepackage{multirow}
\usepackage{array}
\usepackage{xcolor,colortbl}
\usepackage{booktabs}
\usepackage{pifont}
\usepackage[table]{xcolor}
\usepackage{tocloft}

\usepackage{soul}
\usepackage[font=small,labelfont=bf,tableposition=top]{caption}
\usepackage{titletoc}

\definecolor{cvprblue}{rgb}{0.21,0.49,0.74}
\definecolor{emphasize}{RGB}{42,69,189}
\definecolor{linkcolor}{RGB}{231,68,149}
\definecolor{red}{RGB}{223,105,80}
\definecolor{blue}{RGB}{92,102,240}
\definecolor{green}{RGB}{47,170,148}
\definecolor{lightgreen}{RGB}{81, 177, 141}
\definecolor{coolest}{RGB}{204,234,224}
\definecolor{red_background}{RGB}{244,207,199}
\definecolor{orange_background}{RGB}{255,232,210}
\definecolor{gray_background}{rgb}{0.93,0.93,0.93}
\definecolor{lightgray}{rgb}{0.85,0.85,0.85}
\definecolor{green_background}{RGB}{224,250,237}

\usepackage{soul}

\SetCommentSty{mycommfont}

\def\confName{CVPR}
\def\confYear{2026}
\newcommand{\ours}{EmbodiedSplat\xspace}
\newcommand{\oursfast}{EmbodiedSplat-\textit{fast}\xspace}

\title{EmbodiedSplat: Online Feed-Forward Semantic 3DGS \\ for Open-Vocabulary 3D Scene Understanding}

\author{Seungjun Lee \qquad Zihan Wang \qquad Yunsong Wang \qquad Gim Hee Lee \vspace{2pt} \\
National University of Singapore \\
{\tt\small seungjun.lee@u.nus.edu, gimhee.lee@nus.edu.sg} \\ \vspace{-5pt} \\ Project page: \href{https://0nandon.github.io/EmbodiedSplat/}{\textcolor{linkcolor}{EmbodiedSplat.io}}}

\begin{document}

\twocolumn[{
\renewcommand\twocolumn[1][]{#1}
\maketitle
\vspace{-30pt}
\begin{center}
    \captionsetup{type=figure}
    \includegraphics[width=\textwidth]{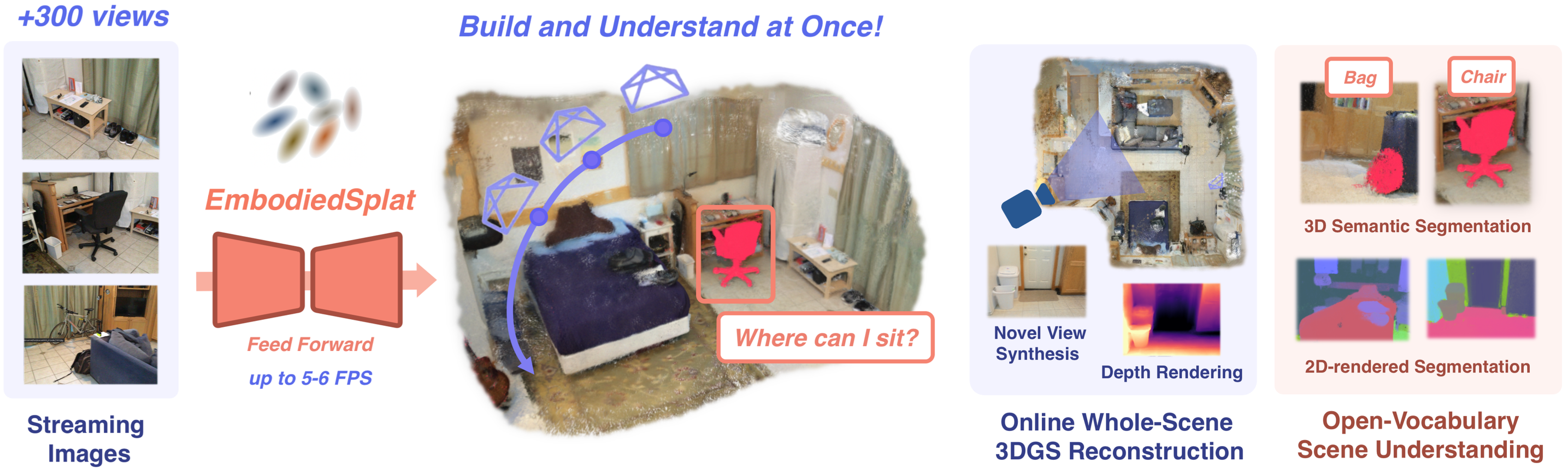}
    \vspace{-20pt}
    \caption{\textbf{Build and understand at Once.} By taking over 300 streaming images, our \ours reconstructs whole-scene open-vocabulary 3DGS in online manner at up to 5-6 FPS per-frame processing time. Reconstructed scene supports diverse perception tasks such as open-vocabulary 3D semantic segmentation, 2D-rendered semantic segmentation and novel-view color synthesis with depth rendering.}
    \label{fig:teaser}
\end{center}
}]

\begin{abstract}

Understanding a 3D scene immediately with its exploration is essential for embodied tasks, where an agent must construct and comprehend the 3D scene in an online and nearly real-time manner. In this study, we propose \textbf{\ours}, an online feed-forward 3DGS for open-vocabulary scene understanding that enables simultaneous online 3D reconstruction and 3D semantic understanding from the streaming images. Unlike existing open-vocabulary 3DGS methods which are typically restricted to either offline or per-scene optimization setting, our objectives are two-fold: 1) Reconstructs the semantic-embedded 3DGS of the entire scene from over 300 streaming images in an online manner. 2) Highly generalizable to novel scenes with feed-forward design and supports nearly real-time 3D semantic reconstruction when combined with real-time 2D models. To achieve these objectives, we propose an Online Sparse Coefficients Field with a CLIP Global Codebook where it binds the 2D CLIP embeddings to each 3D Gaussian while minimizing memory consumption and preserving the full semantic generalizability of CLIP. Furthermore, we generate 3D geometric-aware CLIP features by aggregating the partial point cloud of 3DGS through 3D U-Net to compensate the 3D geometric prior to 2D-oriented language embeddings. Extensive experiments on diverse indoor datasets, including ScanNet, ScanNet++, and Replica, demonstrate both the effectiveness and efficiency of our method. Code will be publicly available on project website.
\end{abstract}   
\vspace{-30pt}
\section{Introduction}
\label{sec:intro}
\vspace{-1mm}

Embodied tasks such as robotic manipulation and navigation         ~\citep{krantz2020beyond,hm3d-ovon, mousavian20196,zhang20233d, ze2023gnfactor, chaplot2020object,wang2023gridmm, wang2025g3d, wang2025d3d} require an agent to perceive the 3D scene immediately with its exploration. Specifically, the embodied agent equipped with a precise SLAM system collects posed RGB or RGB-D images to understand the 3D scene, follow human instructions, and make autonomous decisions based on its own action. In these embodied scenarios, we expect the 3D perception model to satisfy~\citep{xu2024embodiedsam}: 1) \textit{\textbf{Online}}: The model should process streaming images synchronously with its exploration rather than relying on pre-collected data. 2) \textit{\textbf{Real-time}}: High inference speed is required to stay synchronized with its exploration process. 3) \textit{\textbf{Highly-generalizable}}: The model should be generalizable to diverse types of scenes. 4) \textit{\textbf{Whole-scene Understanding}}: Reconstructing and interpreting large-scale 3D scenes are demanded to perform long-term actions. 5) \textit{\textbf{Open-Vocabulary Understanding}}: The model needs to perceive a wide-range of objects described with diverse linguistic forms.

In this paper, our objective is to develop an embodied perception model that meets the above five conditions by leveraging 3D Gaussian Splatting (3DGS)~\citep{kerbl20233d}. 3DGS is the recent 3D representation that supports real-time novel view synthesis with explicit structure which existing representations such as point clouds and NeRF~\citep{mildenhall2021nerf} fail to provide. Owing to its strong capability for high-fidelity real-world digitization, 3DGS has attracted growing interest in the robotics community~\citep{li2025activesplat, jin2025activegs, lee2025diet}, naturally motivating the exploration of open-vocabulary scene understanding with 3DGS. Several pioneering approaches~\citep{qin2024langsplat, shi2024language, zhou2024feature, zhang2025econsg, zuo2025fmgs} distill the foundational knowledge of 2D models~\citep{radford2021learning, li2022language} into 3DGS by rendering the 2D features map with rasterization function. Despite its promise, these methods must render multiple feature maps to interpret the 3D scene, leading to heavy computation and long inference time~\citep{jun2025dr}. To this end, more recent approaches enable direct reference to the 3D space without relying on the heavy rendering function, employing clustering-based methods~\citep{wu2024opengaussian, li2025instancegaussian, jiang2025votesplat} or direct feature-lifting approaches~\citep{jun2025dr, cheng2024occam, marrie2025ludvig, lee2025cf3}. Nevertheless, all of them share two limitations in embodied scenarios: 1) They require per-scene optimization that cannot be generalized to novel scenes without additional training. 2) They cannot be adapted to the online setting. Only a few studies address the partial aspects of embodied perception. Online-LangSplat~\citep{katragadda2025online} and EA3D~\citep{zhou2025ea3d} introduce an online reconstruction framework for semantic 3DGS by leveraging a 3DGS-based SLAM~\citep{matsuki2024gaussian, gao2024hicom}. However, it still requires heavy per-scene optimization, failing to achieve real-time semantic reconstruction ($<$ 2FPS). LSM~\citep{fan2024large} and SIU3R~\citep{xu2025siu3r} propose language-embedded feed-forward 3DGS that can be easily generalized to the new scene. Nonetheless, it does not support online settings or full-scene reconstruction as it operates with only two or a few input views.

To this end, we propose \ours, a novel \textbf{\textit{online}} framework to endow pretrained feed-forward 3DGS~\citep{wang2025freesplat++} with open-vocabulary capability. We integrate two types of CLIP features into newly generated 3D Gaussians for each time step: 1) \textbf{2D CLIP features} are directly produced from current frame and projected onto 3D Gaussians. However, binding original 2D features to all Gaussians incurs huge memory overhead. To this end, we propose a novel \textit{\textbf{Online Sparse Coefficient Field}} with \textit{\textbf{CLIP Global Codebook}} which stores per-Gaussian semantics in a memory-efficient manner. In contrast to existing memory-compression methods such as Auto-encoder~\citep{qin2024langsplat}, Product Quantization (PQ) Index~\citep{jun2025dr} and per-scene optimized codebook~\citep{wu2024opengaussian, li2025instancegaussian}, our approach requires no pretraining or per-scene optimization, preserves the full semantic capability of CLIP, and supports online updates. 2) \textbf{3D CLIP features} are further defined to compensate the 3D geometric prior by aggregating the feature point cloud of 3DGS through 3D U-Net~\citep{choy20194d}. Combining those two types of features enable mutual compensation between rich semantics from CLIP and 3D geometric prior from 3D module, resulting in clear performance improvement. To enable near real-time inference speed, we further propose the faster variant of our \ours, which achieves 5-6 FPS of processing time.

We validate the effectiveness of our \ours in 3D semantic segmentation on diverse indoor scene datasets~\citep{dai2017scannet, rozenberszki2022language, yeshwanth2023scannet++, straub2019replica}. Extensive experimental results show that our \ours largely outperforms existing baselines on segmentation performance and scene reconstruction time. Our contributions are as follows:
\begin{itemize}
    \item 
    Novel framework for embodied 3D perception which enables \textcolor{red}{\textbf{\textit{online}}}, \textcolor{red}{\textbf{\textit{whole-scene}}} reconstruction for 
    \textcolor{red}{\textbf{\textit{language-embedded 3DGS}}} with up to \textcolor{red}{\textbf{\textit{5-6 FPS}}} inference speed.
    \item Combination of 2D CLIP Features with rich semantic capabilities and 3D CLIP Features with geometric prior.
    \item Sparse Coefficient Field with CLIP Global Codebook to store the per-Gaussian language embeddings compactly.
    \item Experiment results show that our framework significantly surpasses the existing semantic 3DGS in terms of segmentation performance and scene reconstruction time.
\end{itemize}

\vspace{-5pt}
\section{Related Works}
\vspace{-2pt}
\noindent \textbf{Open-vocabulary 3D scene understanding.} Understanding 3D scene with free-form language has advanced significantly by bridging the diverse 3D representations such as point clouds, neural radiance fields (NeRF)~\cite{mildenhall2021nerf}, and 3D Gaussian Splatting (3DGS)~\cite{kerbl20233d} with natural language. Point-based methods either leverage 2D open-vocabulary models~\cite{radford2021learning, ghiasi2022scaling, li2022language} to interpret point clouds by associating 3D points with 2D pixels through the camera projection matrix~\cite{nguyen2023open3dis, takmaz2023openmask3d, huang2023openins3d}, or directly distill features from 2D foundation models into a 3D neural network~\cite{lee2024segment, peng2023openscene, ding2023pla, yang2023regionplc, ding2023lowis3d}. NERF-based methods~\cite{engelmann2024opennerf, kerr2023lerf, rashid2023language, koch2025relationfield, kobayashi2022decomposing} follow the feature distillation approach, where the semantic embeddings from CLIP~\cite{radford2021learning}, LSeg~\cite{li2022language} and DINO~\cite{caron2021emerging} are transferred to NERF feature space through 2D rendering function. Although effective, NeRF requires long training and rendering time due to the multiple MLP layers inside. Furthermore, its implicit representation hinders the direct referring of 3D space. In contrast, 3DGS supports real-time novel view synthesis with an explicit point-based structure which motivates the community to understand the 3D scene on top of 3DGS representation. Following this trend, our \ours proposes the first feed-forward 3DGS that enables the online open-vocabulary 3D perception.

\smallskip
\noindent \textbf{Open-vocabulary 3DGS.} Early approaches in semantic 3DGS~\cite{qin2024langsplat, shi2024language, zhou2024feature, zhang2025econsg, zuo2025fmgs, katragadda2025online, fan2024large} follow a framework similar to NeRF-based methods, exploiting the 2D rendering function to align per-Gaussian features with language embeddings from 2D foundation models~\cite{radford2021learning}. Learnable semantic features are attached to each Gaussian and trained by minimizing the distance between 2D-rendered feature maps and image features extracted from the original 2D images. Another line of work~\citep{wu2024opengaussian, li2025instancegaussian, cen2025segment, jiang2025votesplat} proposes clustering-based methods that group Gaussians into instance-level by exploiting 2D segmentation masks from SAM~\cite{kirillov2023segment} and then classify each instance-level Gaussians group. More recent approaches~\cite{jun2025dr, cheng2024occam, marrie2025ludvig, lee2025cf3} directly lift the 2D CLIP features into 3D Gaussians, bypassing the feature distillation. They associate the 2D pixels with 3D Gaussians based on the 2D rendering function and directly attach the langauge embeddings to each Gaussian. Although effective, none of them are readily adaptable to embodied scenarios due to their focus on per-scene optimization and offline setting. In contrast, our approach adopts fully feed-forward design without any per-scene optimization, enabling: 1) online, whole-scene semantic 3DGS reconstruction at near real-time speed, and 2) memory-efficient semantic representations via a compact, scene-agnostic codebook that is dynamically updated during online exploration.

\section{Preliminaries}

\noindent \textbf{3D Gaussian Splatting.} 
3DGS~\cite{kerbl20233d} explicitly models 3D scene as a collection of Gaussian primitives, where each is defined by a mean vector $\mu$, a 3D covariance matrix $\Sigma$, the opacity value $\alpha$ and the color $\mathbf{c}$. To ensure positive definiteness, the covariance matrix is decomposed into $\Sigma = \mathbf{R}\mathbf{S}\mathbf{S}^{\top}\mathbf{R}^{\top}$, where $\mathbf{S}$ is scaling matrix and $\mathbf{R}$ is rotation matrix. Given the $M$ number of 3D Gaussians $\Theta = \{\mu_i, \mathbf{S}_i, \mathbf{R}_i, \alpha_i, \mathbf{c}_i\}^{M}_{i=1}$, the color of 2d pixels $\hat{\mathbf{c}}$ can be rendered with point-based alpha-blending on each ray:
\begin{equation}
\vspace{-2mm}
\label{eq:3dgs}
\small
\hat{\mathbf{c}}=\sum^{N}_{i=1}T_i\tilde{\alpha_i}\mathbf{c}_i, \quad \tilde{\alpha_i} = \alpha_i \mathrm{exp}(-\frac{1}{2}\mathbf{d}^{\top}\Sigma^{-1}_{2D}\mathbf{d}).
\end{equation}
Here, $T_i$ denotes the transmittence and $\mathbf{d} \in \mathbb{R}^{2 \times 1}$ is the pixel distance between the target pixel and the projected point of the Gaussian center. $\Sigma_{2D}$ is the 2D covariance matrix obtained by projecting the 3D covariance $\Sigma$ onto the image plane according to $\Sigma_{2D} = \mathbf{J}\mathbf{W}\Sigma\mathbf{W}^{\top}\mathbf{J}^{\top}$, where $\mathbf{W}$ is the world-to-camera transformataion matrix and $\mathbf{J}$ is the projection Jacobian. The Gaussian parameters $\Theta$ are optimized by minimizing the photometric loss, which estimates the difference between rendered images and observed images in the same camera positions.

\smallskip
\noindent \textbf{FreeSplat++.}
We build our \ours on pretrained FreeSplat++~\cite{wang2025freesplat++}, a feed-forward 3DGS model for scene-level reconstruction. 
Since FreeSplat++ is designed for offline use, we modify its inference pipeline to enable \emph{online} perception from streaming images: \textbf{\textit{1) Input selection.}} Given the current frame $I^{t}\in\mathbb{R}^{H\times W\times 3}$, we select $N$ past frames from time steps $t{-}N$ to $t{-}1$ to reflect the online setting.
\textbf{\textit{2) Per-Frame Encoding.}}
With $I^{t}$ and the $N$ reference views as input, the CNN encoder $\mathcal{E}$ predicts pixel-wise local Gaussian triplets $\Theta^t_l=\{\mu^t_l,\omega^t_l,\mathbf{f}^t_l\}^{H \times W}$ and a depth map $d^t$. 
Here, $\omega^t_l$ are per-pixel confidence scores and $\mathbf{f}^t_l$ are Gaussian latents. The 3D centers $\mu^{t}_l$ are obtained by unprojecting 2D pixels into 3D using predicted depth $d^t$.
\textbf{\textit{3) Pairing for Online Fusion.}}
We fuse the new local triplets $\Theta^t_l$ with the global triplets from the previous step $\Theta^{t-1}_g=\{\mu^{t-1}_g,\omega^{t-1}_g,\mathbf{f}^{t-1}_g\}$ to reduce the 3DGS redundancy if they are overlapped in the 3D space. Specifically, we follow \textit{broader fusion} technique of~\cite{wang2025freesplat++} where triplet pairs $\mathcal{P}^t$ between $\Theta^t_l$ and $\Theta^{t-1}_g$ are constructed if: 
i) two Gaussians $\Theta^t_l(i)$ and $\Theta^{t-1}_g(m_i)$ project to the same pixel in frame $I^t$, and 
ii) the minimum depth difference on frame $I^t$ exceeds a predefined threshold.
\textbf{\textit{4) Fusion rule.}}
For each pair $(i,m_i)\in\mathcal{P}^t$, we combine the two Gaussians using a confidence-weighted update for positions $\mu$ and confidences $\omega$ while latents $\mathbf{f}$ are fused via lightweight GRU:
\begin{subequations}\label{eq:fusion}
\small
\begin{align}
\mu^t_g(m_i) &= \frac{\omega^t_l(i)\,\mu^t_l(i) + \omega^{t-1}_g(m_i)\,\mu^{t-1}_g(m_i)}{\omega^t_l(i) + \omega^{t-1}_g(m_i)}, \\
\omega^t_g(m_i) &= \omega^t_l(i) + \omega^{t-1}_g(m_i), \\
\mathbf{f}^t_g(m_i) &= \mathrm{GRU}\!\big(\mathbf{f}^t_l(i),\, \mathbf{f}^{t-1}_g(m_i)\big).
\end{align}
\end{subequations}
Local Gaussians which have no valid match with global Gaussians are just appended to the global set unchanged.
\textbf{\textit{5) Decoding.}}
After processing all frames, we decode the final global latents $\mathbf{f}^T_g$ into Gaussian parameters $\{\Sigma,\alpha,\mathbf{c}\}$ using an MLP decoder $\mathcal{D}$, where $T$ is the last time step. Kindly refer to original paper~\citep{wang2025freesplat++} or Sec.~\ref{sec:supple_embodiedsplat} of Supplementary material for more details.

\begin{figure*}[t]
\centering
\includegraphics[width=1.0\linewidth]{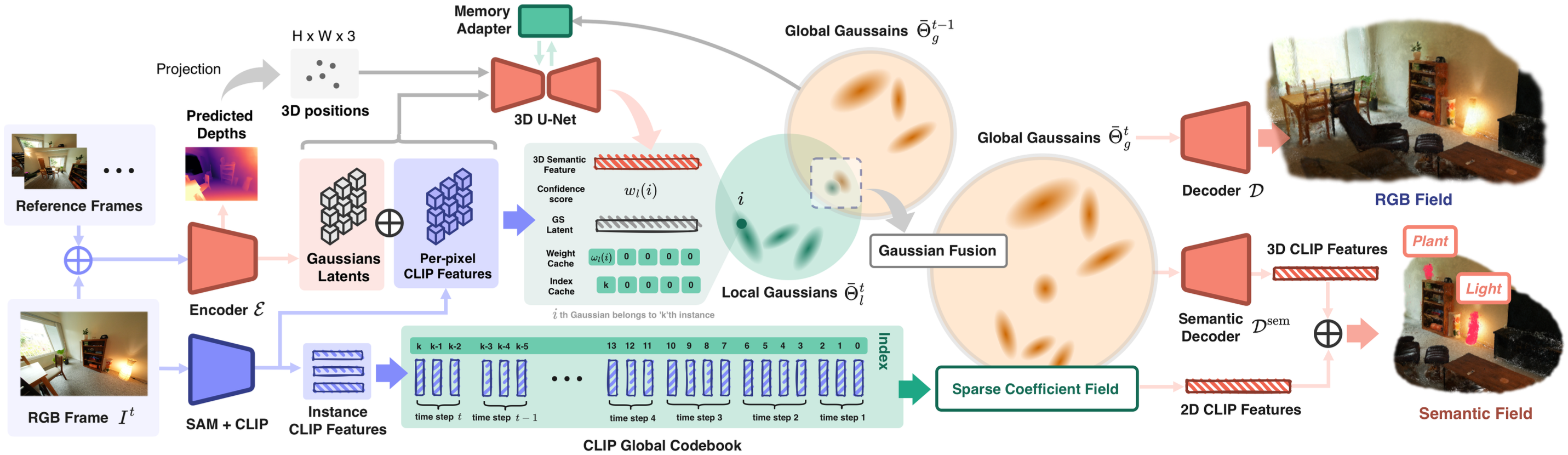}
  \vspace{-20pt}
  \caption{\textbf{Overall framework of our \ours.} We endow feed-forward 3DGS with semantic understanding capabilities by binding the two types of CLIP features: 1) \textbf{2D Semantic Features} are attached to each Gaussian via \textbf{\textit{Sparse Coefficient Field}} with \textbf{\textit{CLIP Global Codebook}}, effectively reducing memory consumption while preserving semantic generalizability of CLIP. 2) \textbf{3D Geometric-aware Features} are produced by aggregating the feature point cloud of 3DGS through 3D U-Net~\citep{choy20194d} and temporal-aware memory adapter~\citep{xu2024memory}. These two types of features enable mutual compensation between semantic and 3D geometry, which results in superior understanding capabilities compared to the existing baselines.}
  \label{fig:framework}
  \vspace{-15pt}
\end{figure*}

\section{Our Methods}

\noindent \textbf{Overview.}
We learn a mapping function $f_{\theta}$ with learnable parameters $\theta$ that transforms a posed image stream $\{I^t\}_{t=1}^{T}$ (typically $T>300$) into a 3D Gaussian field with per-Gaussian semantics:
$\{\mu_i, \mathbf{S}_i, \mathbf{R}_i, \alpha_i, \mathbf{c}_i, \mathbf{s}_i\}_{i=1}^{M}$.
Here, $M$ is the number of Gaussians and $\mathbf{s}_i$ is the language embedding for Gaussian $i$.
Sec.~\ref{sec:method_embodedsplat} introduces \textbf{\ours}, which is a feed-forward 3DGS framework endowed with online semantic reasoning.
Sec.~\ref{sec:method_embodedsplat_fast} presents \textbf{\ours-\textit{fast}}, which is a faster, lightweight variant of \ours for near real-time inference. Fig.~\ref{fig:framework} shows the overall architecture of our framework.

\subsection{EmbodiedSplat}
\label{sec:method_embodedsplat}

\noindent \textbf{Lifting 2D Semantic Features.}
Prior semantic 3DGS methods~\cite{qin2024langsplat, shi2024language, wu2024opengaussian, jun2025dr, li2025instancegaussian} do \textbf{\textit{3D-to-2D}}: associate 2D features to 3D Gaussians by rasterizing them into the image plane using Eq.~\ref{eq:3dgs}. 
We take the opposite route of \textbf{\textit{2D-to-3D}}: during online reconstruction, pixel-wise 2D features are directly unprojected into 3D space along the local Gaussian triplets $\Theta^t_l$. 
Concretely, we augment each triplet $\Theta^t_l=\{\mu^t_l,\omega^t_l,\mathbf{f}^t_l\}$ with $D$-dimensional pixel-level CLIP features $\mathbf{s}^t_l \in \mathbb{R}^{H \times W \times D}$ to form \emph{Gaussian quadruplets} $\tilde{\Theta}^t_l=\{\mu^t_l,\omega^t_l,\mathbf{f}^t_l,\mathbf{s}^t_l\}$. During the Gaussian fusion, local $\mathbf{s}^t_l(i)$ is combined with paired global CLIP feature $\mathbf{s}^{t-1}_g(m_i)$ by following the confidence-weighted average from Eq.~\ref{eq:fusion}\textcolor{cvprblue}{a}:
\begin{equation}
\vspace{-2mm}
\label{eq:clip_fusion}
\mathbf{s}^t_g(m_i)=\frac{\omega^t_l(i)\,\mathbf{s}^t_l(i)+\omega^{t-1}_g(m_i)\,\mathbf{s}^{t-1}_g(m_i)}{\omega^t_l(i)+\omega^{t-1}_g(m_i)}.
\end{equation}

\smallskip
\noindent \textbf{Global Codebook.}
Binding a full CLIP features to every Gaussian is memory-intensive, especially for scenes with millions of Gaussians. 
Prior works compress CLIP with encoder–decoder networks~\citep{qin2024langsplat, katragadda2025online, zhang2025econsg} or Product Quantization (PQ)~\citep{jun2025dr}. 
These approaches require extra pretraining and often lose information due to dimensionality reduction~\cite{jun2025dr, li2025instancegaussian}. 
Another line of works learns per-scene optimized codebooks~\citep{wu2024opengaussian, li2025instancegaussian} which cannot be adapted to generalizable setting.

We instead present a \textbf{\textit{CLIP Global Codebook}} with an \textbf{\textit{Online Sparse Coefficient Field}}. 
This design: 1) does not require any pretraining or per-scene optimization, 2) effectively reduces memory while preserving original open-vocabulary semantics of CLIP, and 3) supports real-time online updates. The key observation is that the number of unique semantics in a scene is far smaller than the number of Gaussians~\cite{li2025instancegaussian}. 
We therefore discretize the scene into instance-level entities and represent the semantic feature of each Gaussian as a \emph{sparse} linear combination of instance features observed across multiple views.

Given the current frame $I^t$ and pixel-wise CLIP features, we compute instance-level CLIP features $\hat{\mathbf{s}}^t \in \mathbb{R}^{M^t \times D}$ using an image segmentation model~\citep{kirillov2023segment, zhao2023fast} followed by average pooling, where $M^t$ is the number of instances in $I^t$. These instance features are accumulated to the global codebook $\mathbf{C}$ along the time steps: $\mathbf{C}^t=\mathrm{concat}(\mathbf{C}^{t-1}, \hat{\mathbf{s}}^t)$ where $\mathbf{C}^{t-1}$ is codebook from previous time step $t{-}1$.
Each entry in codebook receives a monotonically increasing index for fast lookup. Every pixel-aligned local Gaussian is then paired with its owning instance via this index. Overall, the codebook at step $t$ is defined as $\mathbf{C}^t=[\hat{\mathbf{s}}^{1:K}]$, where $K=\sum_{i=1}^{t} M^i$ is the total number of accumulated instance features from $I^1$ to $I^t$. 
These codebook vectors act as global basis functions for semantic Gaussians, where their sparse coefficients are maintained by online fusion described next.

\smallskip
\noindent \textbf{Sparse Caches and Reconstruction.}
Instead of storing a dense CLIP vector $\mathbf{s}^t_l \in \mathbb{R}^D$ in the quadruplets $\tilde{\Theta}^t_l$, we attach two vectors with length $L$: 
1) An \emph{index cache} $\mathbf{I}^t_l \in \mathbb{R}^L$ that provides association with the global codebook; 2) A \emph{weight cache} $\Omega^t_l \in \mathbb{R}^L$ that stores sparse coefficients for the corresponding indices. 
We initialize both caches with zeros. 
For each $i$th local Gaussian $\tilde{\Theta}^t_l(i)$, we insert the codebook index of its paired instance feature into the first slot of $\mathbf{I}^t_l(i)$, such that $\mathbf{I}^t_l(i,0)=k$ where the Gaussian belongs to the $k$-th instance in the codebook. 
We also copy the confidence score of $\omega^t_l(i)\in[0,1]$ into the first slot of weight cache: $\Omega^t_l(i,0)=\omega^t_l(i)$. 
This yields pixel-wise \emph{Gaussian quintuplets} $\widetilde{\Theta}^t_l=\{\mu^t_l,\omega^t_l,\mathbf{f}^t_l,\mathbf{I}^t_l,\Omega^t_l\}$ for frame $I^t$, where each Gaussian is linked to a single codebook entry via $\mathbf{I}^t_l(i,0)$ with corresponding weight $\Omega^t_l(i,0)$. We provide the toy example of sparse caches initialization in Fig.~\ref{fig:supplee_spc_init} to further aid understanding.

\smallskip
\noindent \textbf{Online Update.} During the Gaussian fusion, initialized caches $\{\mathbf{I}, \Omega\}$ for local Gaussians are combined with those of their paired global Gaussians using our proposed online update strategy, Algorithm~\ref{alg:sparse_coeff}. Here, we reformulate the confidence-weighted fusion in Eq.~\ref{eq:clip_fusion} for the sparse coefficient field. Specifically, it accumulates related codebook indices and weights over the time steps for each Gaussian (Lines 1-4) while the weights are updated based on the confidence-weighted fusion (Lines 3-4) to progressively aggregate evidence from new incoming views. 
After each fusion, it keeps only the top $L{-}1$ entries by confidence $\Omega^t_g$ (Lines 5–6) where it leads to two benefits: 1) It removes noisy indices with low-confidence, which sharpens the semantic reconstruction in Eq.~\ref{eq:clip_recovery}. 2) It \emph{fixes} the cache size at $2(L{-}1)$ numbers per Gaussian. 
We set $L{=}6$ and thus $2(L{-}1){=}10$ in our experiments, which is far smaller than a full CLIP vector (512 or 768), yielding substantial memory savings (\cf Tab.~\ref{tab:ablation_memory_size} and Tab.~\ref{tab:supple_exp_memory}).

After final step $T$, we renormalize each weight cache to sum to one:
$\Omega^T_g(i) \leftarrow \Omega^T_g(i) / \sum_{j=1}^{L-1}\Omega^T_g(i,j)$. For semantic reasoning, per-Gaussian CLIP feature can be reconstructed as a sparse linear combination of codebook vectors:
\begin{equation}
\vspace{-2mm}
\label{eq:clip_recovery}
\small
\mathbf{s}^T_g(i)=\sum_{j=1}^{L-1}\Omega^T_g(i,j)\,\mathbf{C}^T\!\big(\mathbf{I}^T_g(i,j)\big), 
\quad \sum_{j=1}^{L-1}\Omega^T_g(i,j)=1.
\end{equation}
Here, $\mathbf{C}^T(\mathbf{I}^T_g(i,\cdot))$ serves as a local basis vectors and $\Omega^T_g(i,\cdot)$ are its sparse coefficients. 
Since the codebook stores \emph{original} instance-level CLIP features, we retain the full open-vocabulary semantics of CLIP. We further provide the detailed illustration of Algorithm~\ref{alg:sparse_coeff} with toy example in Fig.~\ref{fig:supple_online_fuse_toy-5} for better understanding.

\smallskip
\noindent \textbf{Geometry-aware 3D Semantic Features.}
2D CLIP features $\mathbf{s}^T_g$ are semantically rich but lack explicit 3D priors since they are originally derived from images. 
Inspired by~\cite{lee2024segment}, we therefore construct \emph{geometry-aware} 3D features to improve perception in 3D space: 
\textbf{\textit{1) Inputs to the 3D Backbone:}}
Given the local Gaussian quintuplets $\widetilde{\Theta}^t_l$ at time $t$, we feed the 3D coordinates $\mu^t_l \in \mathbb{R}^{(H\times W)\times 3}$ and semantic-aware latents $\mathbf{g}^t_l$ into a 3D sparse U-Net~\cite{choy20194d} equipped with a memory-based adapter~\cite{xu2024memory}. Here, $\mathbf{g}^t_l = \mathbf{f}^t_l + \mathrm{proj}(\mathbf{s}^t_l)$,
where $\mathbf{f}^t_l$ are Gaussian latents and $\mathrm{proj}(\cdot)$ projects pixel-wise CLIP features $\mathbf{s}^t_l$ to match the feature dimensionality.
\textbf{\textit{2) 3D Aggregation with Memory.}}
The 3D sparse U-Net aggregates features over the point cloud and injects geometric priors from previously reconstructed scenes via the memory-based adapter. 
It outputs compact 3D features $\hat{\mathbf{g}}^t_l \in \mathbb{R}^{(H\times W)\times D^s}$, which we append to the local quintuplets $\widetilde{\Theta}^t_l$. We keep $D^s=64$ to preserve memory efficiency.
\textbf{\textit{2) Gaussian Fusion.}}
For each matched Gaussian pair, we fuse local and global 3D features with an additional GRU network:
$\hat{\mathbf{g}}^t_g(m_i) = \mathrm{GRU}\!\big(\hat{\mathbf{g}}^t_l(i),\, \hat{\mathbf{g}}^{t-1}_g(m_i)\big)$, following Eq.~\ref{eq:fusion}\textcolor{cvprblue}{c}.
\textbf{\textit{3) Open-Vocabulary 3D Reasoning.}}
When open-vocabulary 3D perception is required, we project the global 3D features $\hat{\mathbf{g}}^t_g$ into CLIP space using a lightweight MLP decoder $\mathcal{D}^{\mathrm{sem}}$.

\begin{algorithm}[t]
\small
\caption{Online Fusion of Index/Weight Caches}
\label{alg:sparse_coeff}

\SetKwInOut{Input}{In}
\SetKwInOut{Output}{Out}

\LinesNotNumbered
\tcc{\textit{Note:} Both $\mathbf{I}$ and $\Omega$ have fixed length $L$.}
\Input{$(i,m_i)\in\mathcal{P}^t$;~~~~~~~~~~~~~~~~~~~~~~~~~~~~~~~~~~~~~~~\tcp{Gaussian pair} \\
$\{\mathbf{I}^t_l(i),\,\Omega^t_l(i),\,\omega^t_l(i)\}$;~~~~~~~~~~~~~~~~~~~~~~~~~~~~\tcp{Local caches}\\
$\{\mathbf{I}^{t-1}_g(m_i),\,\Omega^{t-1}_g(m_i),\,\omega^{t-1}_g(m_i)\}$~\tcp{Global caches}\\
}
\Output{$\{\mathbf{I}^t_g(m_i),\,\Omega^t_g(m_i)\}$ \tcp*{New global Gaussian $m_i$}}

\LinesNumbered

\tcc{Append the first entry of local index cache to the end of the previous global index cache}
$\mathbf{I}^{t-1}_g(m_i,{-}1) \gets \mathbf{I}^t_l(i,0)$\;

\tcc{Start the new global index cache from the updated global cache $\mathbf{I}^{t-1}_g(m_i)$}
$\mathbf{I}^t_g(m_i) \gets \mathbf{I}^{t-1}_g(m_i)$\;

\tcc{Confidence-weighted carry-over of previous global weight cache $\Omega^{t-1}_g(m_i)$}
$\displaystyle \Omega^t_g(m_i) \gets \frac{\omega^{t-1}_g(m_i)}{\omega^t_l(i)+\omega^{t-1}_g(m_i)} \cdot \Omega^{t-1}_g(m_i)$\;

\tcc{Inject the first entry of local weight cache, scaled by coefficient from weighted-sum}
$\displaystyle \Omega^t_g(m_i,{-}1) \gets \frac{\omega^t_l(i)}{\omega^t_l(i)+\omega^{t-1}_g(m_i)} \cdot \Omega^t_l(i,0)$\;

\tcc{Keep the strongest contributors: sort by weight in descending order}
$I \gets \mathrm{argsort}\!\left(\Omega^t_g(m_i),\, \mathrm{descending}{=} \mathrm{True}\right)$; \\
$\Omega^t_g(m_i) \gets \Omega^t_g(m_i, I)$; \quad $\mathbf{I}^t_g(m_i) \gets \mathbf{I}^t_g(m_i, I)$;\;

\tcc{Prune to top $L{-}1$ entries and zero tail slot}
$\Omega^t_g(m_i,{-}1) \gets 0$; \quad $\mathbf{I}^t_g(m_i,{-}1) \gets 0$; 

\Return $\mathbf{I}^t_g(m_i),\,\Omega^t_g(m_i)$\;
\end{algorithm}

\begin{table*}[h!]
    \renewcommand{\arraystretch}{1.5}
    \centering
    \footnotesize
    \resizebox{\linewidth}{!}{
    \begin{tabular}{l|c|cc|cc|cc|cc|cc|c|c|c}
    \toprule
    \multirow{3}{*}{\textbf{Method}} & \multirow{3}{*}{\shortstack{\textbf{Search}\\\textbf{Domain}}} & \multicolumn{6}{c|}{\textbf{ScanNet}~\cite{dai2017scannet}} & \multicolumn{2}{c|}{\textbf{ScanNet200}~\cite{rozenberszki2022language}} & \multicolumn{2}{c|}{\textbf{ScanNet++}~\cite{yeshwanth2023scannet++}} & \multirow{3}{*}{\shortstack{\textbf{Scene-reconstruction}\\\textbf{Time}\\\\(363 images)}} & \multirow{3}{*}{\shortstack{\textcolor{gray}{\textbf{Per-Scene}} 
    \textbf{/}\\ \textcolor{green}{\textbf{Generalizable}}}} & \multirow{3}{*}{\textbf{\textcolor{red}{on} / \textcolor{blue}{off}}} \\
    \cline{3-12}
     &  & \multicolumn{2}{c|}{10 classes} & \multicolumn{2}{c|}{15 classes} & \multicolumn{2}{c|}{19 classes} & \multicolumn{2}{c|}{70 classes} & \multicolumn{2}{c|}{20 classes} & & & \\
     &  & mIoU & mACC & mIoU & mACC & mIoU & mACC & mIoU & mACC & mIoU & mACC & & & \\
     \midrule \noalign{\vskip -2pt}
     LangSplat~\cite{qin2024langsplat} & \multirow{3}{*}{2D} & 6.52 & 20.11 & 3.25 & 13.16 & 1.34 & 7.44 & 0.72 & 4.39 & 2.21 & 10.18 & $\sim$ 6 hr & \multirow{3}{*}{\textcolor{gray}{\textbf{Per-Scene}}} & \multirow{2}{*}{\textcolor{blue}{\textbf{Offline}}} \\
     LEGaussians~\cite{shi2024language} & & 6.79 & 18.13 & 4.13 & 15.49 & 2.53 & 5.67 & 1.39 & 5.45 & 2.93 & 9.34 & $\sim$ 6 hr & \\
     \cline{15-15}
     Online-LangSplat~\cite{katragadda2025online} & & 7.13 &  21.56 & 3.89 & 14.52 & 3.45 & 8.97 & 2.45 & 4.12 & 4.51 & 11.34 & \cellcolor{orange_background} 5.4 min (1.12 FPS) & & \textcolor{red}{\textbf{Online}} \\
     \midrule
     OpenGaussian~\cite{wu2024opengaussian} & \multirow{4}{*}{3D} & 29.50 & 44.61  & 23.74 & 39.14 & 22.52 & 35.02 & 15.15 & 25.66 & 25.65 & 37.03 & $\sim$ 2.5 hr & \multirow{4}{*}{\textcolor{gray}{\textbf{Per-Scene}}} & \multirow{4}{*}{\textcolor{blue}{\textbf{Offline}}} \\
     Occam's LGS~\cite{cheng2024occam} & & 42.14 & 70.28 & 35.04 & 63.71 & 30.49 & 57.91 & 20.32 & 40.49 & 34.08 & 61.19 & $\sim$ 2 hr & & \\
     Dr. Splat~\cite{jun2025dr} & & 39.21 & 66.66 & 31.84 & 60.58 & 28.38 & 55.85 & 19.29 & 33.84 & 39.85 & 58.34 & $\sim$ 2 hr & & \\
     InstanceGaussian~\cite{li2025instancegaussian} & & 29.77 & 52.32 & 28.79 & 50.07 & 26.57 & 48.63 & 23.20 & 38.32 & 29.98 & 47.47 & $\sim$ 3 hr & & \\
     \midrule
     \textbf{EmbodiedSplat} (RGB) & \multirow{2}{*}{3D} & \cellcolor{red_background} \textbf{49.81} & \cellcolor{red_background} \textbf{76.13} & \cellcolor{red_background}  \textbf{49.23} & \cellcolor{red_background} \textbf{75.47} & \cellcolor{red_background} \textbf{46.22} & \cellcolor{red_background} \textbf{70.37} & \cellcolor{red_background} \textbf{31.16} & \cellcolor{orange_background} \underline{48.38} & \cellcolor{orange_background} \underline{41.93} & \cellcolor{orange_background} \underline{61.50} & \underline{8 min (0.75 FPS)} & \multirow{2}{*}{\textcolor{green}{\textbf{Generalizable}}} & \multirow{2}{*}{\textcolor{red}{\textbf{Online}}} \\
     \textbf{EmbodiedSplat}-\textit{fast} (RGB) & &  \cellcolor{orange_background} \underline{47.86} & \cellcolor{orange_background} \underline{77.62} & \cellcolor{orange_background} \underline{43.21} & \cellcolor{orange_background} \underline{73.85} & \cellcolor{orange_background} \underline{41.03} & \cellcolor{orange_background} \underline{70.12} & \cellcolor{orange_background} \underline{30.46} & \cellcolor{red_background} \textbf{55.31} & \cellcolor{red_background} \textbf{45.53} & \cellcolor{red_background} \textbf{71.42} & \cellcolor{red_background} \textbf{1 min 10 sec (5.18 FPS)} & & \\
     \midrule
     \cellcolor{gray_background} \textbf{EmbodiedSplat} (RGB-D) & \multirow{2}{*}{3D} & \cellcolor{gray_background} 57.41 & \cellcolor{gray_background} 82.45 & \cellcolor{gray_background} 55.18 & \cellcolor{gray_background} 80.27 & \cellcolor{gray_background} 52.12 & \cellcolor{gray_background} 75.66 & \cellcolor{gray_background} 34.75 & \cellcolor{gray_background} 52.36 & \cellcolor{gray_background} 44.03 & \cellcolor{gray_background} 66.27 & \cellcolor{gray_background} 8 min (0.75 FPS) & \multirow{2}{*}{\textcolor{green}{\textbf{Generalizable}}} & \multirow{2}{*}{\textcolor{red}{\textbf{Online}}} \\
     \cellcolor{gray_background} \textbf{EmbodiedSplat}-\textit{fast} (RGB-D) & & \cellcolor{gray_background} 51.05 & \cellcolor{gray_background} 80.15 & \cellcolor{gray_background} 46.92 & \cellcolor{gray_background} 77.15 & \cellcolor{gray_background} 43.89 & \cellcolor{gray_background} 72.73 & \cellcolor{gray_background} 32.43 & \cellcolor{gray_background} 58.14 & \cellcolor{gray_background} 51.09 & \cellcolor{gray_background} 78.68 & \cellcolor{gray_background} 1 min 10 sec (5.18 FPS)  &  &  \\
    \bottomrule
    \end{tabular}}
    \newline
    \vspace{2pt}
    \caption{\textbf{Quantitative comparisons on 3D Semantic Segmentation across ScanNet~\cite{dai2017scannet}, ScanNet200~\cite{rozenberszki2022language} and ScanNet++~\cite{yeshwanth2023scannet++}.} We compare the performance of our \ours with existing semantic 3DGS methods on 3D semantic segmentation. Our \ours achieves the best performance across all benchmarks while maintaining the shortest reconstruction time. The best results are in \colorbox{red_background}{\textbf{bold}} while the second best results are \colorbox{orange_background}{\underline{underscored}}.}
    \label{tab:quantitative}
    \vspace{-20pt}
\end{table*}

\noindent \textbf{Training.} Given the $T$ number of views, our \ours reconstructs the semantic Gaussians field $\bar{\Theta}^T_g = \{\mu^T_g, \omega^T_g, \mathbf{f}^T_g, \mathbf{I}^T_g, \Omega^T_g, \hat{\mathbf{g}}^T_g\}$ with global codebook $\mathbf{C}^{T}$. Finally, the model is trained by minimizing the cosine similarity between 2D CLIP features and geometric-aware 3D CLIP features which can be formulated as:
\begin{equation}
\vspace{-2mm}
\label{eq:loss}
\mathcal{L}_{\mathrm{cos}} = 1 - \mathrm{cos}(\mathbf{s}^T_g, \mathcal{D}^{\mathrm{sem}}(\hat{\mathbf{g}}^T_g)),
\end{equation}
where $\mathrm{cos}(\cdot, \cdot)$ denotes cosine similarity with L2 feature normalization. During training, the parameters for the feed-forward 3DGS are initialized with the pretrained FreeSplat++~\cite{wang2025freesplat++} and kept fixed, while the remaining components, such as the 3D U-Net and the memory adapter are optimized by Eq.~\ref{eq:loss}. Here, 2D features $\mathbf{s}^T_g$ serve as supervision while ground-truth class labels are not used.

\smallskip
\noindent \textbf{2D-3D Ensemble.} During the inference, 2D semantic features $\mathbf{s}^T$ and 3D geometric features $\hat{\mathbf{g}}^T$ are exploited to obtain respective classification probability $\mathbf{P}^{\mathrm{2D}} = p(\mathrm{cos}(\mathbf{t}, \mathbf{s}^T))$ and $\mathbf{P}^{\mathrm{3D}} = p(\mathrm{cos}(\mathbf{t}, \mathcal{D}^{\mathrm{sem}}(\hat{\mathbf{g}}^T)))$, where $p(\cdot, \cdot)$ denotes softmax operation and $\mathbf{t}$ is text features from CLIP text encoder. The final probability is yielded by the geometric mean between $\mathbf{P}^{\mathrm{2D}}$ and $\mathbf{P}^{\mathrm{3D}}$~\cite{peng2023openscene, lee2024segment}:
\begin{equation}
\vspace{-2mm}
\mathbf{P} = \mathrm{max}(\mathbf{P}^{\mathrm{2D}}, \mathbf{P}^{\mathrm{3D}})^{\tau} \cdot \mathrm{min}(\mathbf{P}^{\mathrm{2D}}, \mathbf{P}^{\mathrm{3D}})^{1-\tau},
\end{equation}
where $\tau$ is the exponent to increase confidence.

\subsection{EmbodiedSplat-\textit{fast}}
\label{sec:method_embodedsplat_fast}

Real-time reconstruction speed per frame is crucial for embodied tasks, where it enables immediate understanding and interaction with 3D scene~\cite{xu2024embodiedsam}. To this end, we introduce the variant of our \ours to support the nearly real-time online reconstruction of semantic Gaussians. We add three variations to our \ours architecture: 1) We replace heavy 2D foundation models~\cite{radford2021learning, ghiasi2022scaling, kirillov2023segment} with real-time 2D perception models~\cite{li2025mask, zhao2023fast}. 2) We remove the 3D U-Net with memory adapter by only using the 2D CLIP features to improve the inference speed. 3) Finally, we further propose an efficient 3D search strategy to calculate the classification probability $\mathbf{P}^{\mathrm{2D}}$ faster, which is detailed in the following section. We name our variant model that achieves 5-6 FPS processing time as \ours-\textit{fast}.

\smallskip
\noindent \textbf{Codebook-based Cosine Similarity.}
Computing cosine similarity for every Gaussians against a text query is costly. With a single prompt, the complexity of naive per-Gaussian cosine similarity is $O(MD)$, where $M$ is the number of Gaussians (often $>10^6$) and $D$ is the CLIP dimension.
To reduce this cost, we exploit the linear combination in Eq.~\ref{eq:clip_recovery}.

By assuming that the codebook vectors are unit-normalized and defining the normalized per-Gaussian feature
$\bar{\mathbf{s}}^T_g(i) = \mathbf{s}^T_g(i) / \left\|\mathbf{s}^T_g(i)\right\|_2$, 
we can rewrite Eq.~\ref{eq:clip_recovery} as:
\begin{equation}
\label{eq:clip_recovery_rethink}
\footnotesize
\bar{\mathbf{s}}^T_g(i) \;\approx\; \sum_{j=1}^{L-1} \Omega^T_g(i,j)\; \mathbf{C}^T\!\big(\mathbf{I}^T_g(i,j)\big),
\quad \left\|\mathbf{C}^T(k)\right\|_2 = 1.
\end{equation}
For unit vectors, cosine similarity equals inner product, and inner products are linear. 
Hence, for a unit-normalized text embedding $\mathbf{t}$:
{\scriptsize
\vspace{-8pt}
\begin{equation}
\label{eq:cos_rethink}
\begin{aligned}
\mathrm{cos}\!\big(\mathbf{t}, \bar{\mathbf{s}}^T_g(i)\big)
= \big\langle \mathbf{t}, \bar{\mathbf{s}}^T_g(i) \big\rangle
&\approx \sum_{j=1}^{L-1}\Omega^T_g(i,j)\, \big\langle \mathbf{t}, \mathbf{C}^T\!\big(\mathbf{I}^T_g(i,j)\big) \big\rangle \\
&\approx \sum_{j=1}^{L-1}\Omega^T_g(i,j)\, \mathrm{cos}\!\big(\mathbf{t}, \mathbf{C}^T\!\big(\mathbf{I}^T_g(i,j)\big)\big). 
\end{aligned}\vspace{-5pt}
\end{equation}
}
We precompute $\mathrm{cos}\!\big(\mathbf{t}, \mathbf{C}^T(k)\big)$ for all $k$ in the global codebook resulting in a cost of $O(KD)$, where $K$ is the codebook size. Using these precomputed values, the cosine similarity of each Gaussian reduces to a sparse weighted sum over at most $L{-}1$ entries via Eq.~\ref{eq:cos_rethink}, resulting in a total cost of $O\!\big(KD + M(L{-}1)\big)$.
Since $K \ll M$ and $L$ is small, this is substantially cheaper than $O(MD)$ which leads to much faster 3D search (\cf Tab.~\ref{tab:ablation_cosine_similarity}).

\vspace{-2pt}
\section{Experiments}
\label{sec:exp}
\vspace{-3pt}

\noindent \textbf{Implementation Details.} Following~\cite{xu2024embodiedsam, xu2024memory}, we train \ours in two stages. 1) To warm up the model, we first train our \ours as single-view perception model by using the individual RGB frames with no time step. Specifically, the model is trained for 100,000 iterations without memory adapter. 2) We further finetune the model with memory adapter by feeding the streaming RGB images for 300,000 iterations. Specifically, we randomly sample 8 to 10 consecutive frames from the whole multi-view images. To avoid GPU OOM, all inputs are resized to 384 $\times$ 512 and batch size is set to 1. During inference, we sample the keyframes that cover the whole scene by following~\cite{duzceker2021deepvideomvs, sayed2022simplerecon} and reconstruct the semantic 3DGS for the entire scene in online fashion. For image segmentation model, we use FastSAM~\cite{zhao2023fast} to improve the inference speed by following~\cite{wang2025dynam3d}. OpenSeg~\cite{ghiasi2022scaling} is further used as pixel-level CLIP for \ours by following~\cite{zhang2025econsg, peng2023openscene, lee2024segment, lee2026segment} while Mask-Adpater~\cite{li2025mask} is adopted as instance-level CLIP for \ours-\textit{fast} to enable the near real-time reconstruction speed. All experiments are conducted on single NVIDIA RTX 6000 Ada GPU.

\begin{figure*}[t]
\centering
\includegraphics[width=1.0\linewidth]{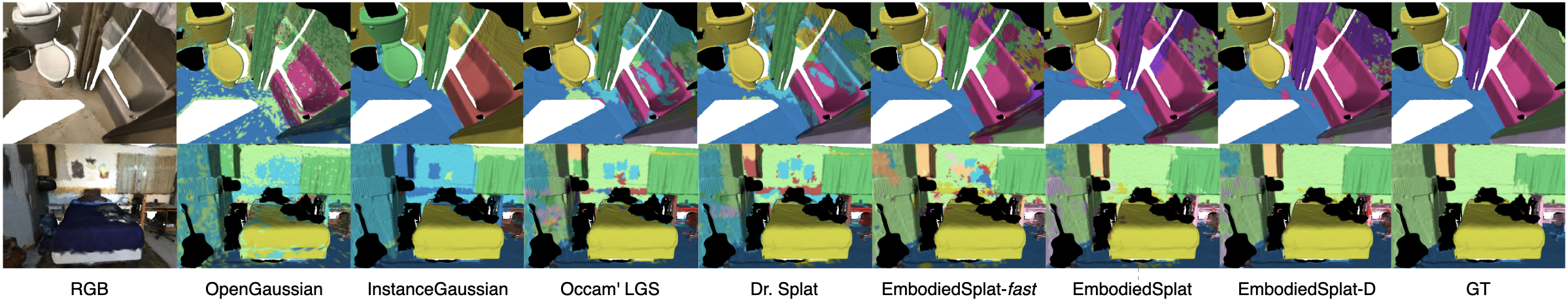}
  \vspace{-17pt}
  \caption{\textbf{Quantitative comparisons on 3D semantic segmentation.}}
  \label{fig:qualitative}
  \vspace{-13pt}
\end{figure*}

\smallskip
\noindent \textbf{Datasets.} We select three real-world indoor datasets for the experiments: ScanNetv2~\cite{dai2017scannet}, ScanNet200~\cite{shi2024language} and ScanNet++~\cite{yeshwanth2023scannet++} and one synthetic indoor dataset: Replica~\cite{straub2019replica}. ScanNetv2~\cite{dai2017scannet} is a large-scale RGB-D and point clouds dataset with 1,513 indoor scenes which also provides semantic annotations for 20 classes. By following~\cite{wang2024freesplat, wang2025freesplat++}, we use 100 scenes for training and sample 10 scenes for testing. We evaluate the performance on 3D semantic segmentation with varying number of classes where 10 classes, 15 classes and 19 classes are provided as candidate labels to the model. ScanNet200~\cite{rozenberszki2022language} is a fine-grained annotated version of ScanNetv2 which contains 200 classes. We sample 70 classes present in the 10 testing scenes and use them for evaluation. ScanNet++~\cite{yeshwanth2023scannet++} is a high-quality indoor dataset with 450 indoor scenes. We use the official training split for the training and select 4 scenes for the evaluation. We sample 20 classes for ScanNet++ that share a similar class configuration with ScanNetv2 and evaluate the semantic segmentation. Replica~\cite{straub2019replica} is a synthetic indoor dataset annotated with 48 classes. Following~\cite{lee2024segment}, we evaluate on 8 scenes in Replica for open-set semantic segmentation.

\smallskip
\noindent \textbf{Baselines.} We formulate the baselines with semantic 3DGS methods in two categories. The first category is rasterization-based methods, where they render the 2D feature maps to understand the 3D scene. We name this baseline as \textbf{2D} methods where LangSplat~\cite{qin2024langsplat}, LEGaussians~\cite{shi2024language} and Online-LangSplat~\cite{katragadda2025online} are chosen for this category. Following~\cite{jun2025dr}, we modify their inference strategy to support direct 3D referring operation without rendering 2D feature maps. The second cateogry is named as \textbf{3D} methods where their search operation is directly conducted in 3D space with semantic Gaussians. OpenGaussian~\cite{wu2024opengaussian}, Occam's LGS~\cite{cheng2024occam}, Dr. Splat~\cite{jun2025dr} and InstanceGaussian~\cite{li2025instancegaussian} are selected for this category.

\smallskip
\noindent \textbf{Experimental Settings.} For the per-scene optimization baselines~\cite{qin2024langsplat, shi2024language, katragadda2025online, wu2024opengaussian, li2025instancegaussian, jun2025dr, cheng2024occam}, we initialize the 3DGS with ground-truth point clouds and camera poses given by the dataset and optimize the Gaussians to each testing scene. Given the optimized 3DGS with semantic features, we assign the labels to ground-truth point clouds by aggregating the contribution of individual Gaussians to each 3D point based on Mahalanobis distance~\cite{de2000mahalanobis} following~\cite{huang2024gaussianformer, huang2025gaussianformer}. Annotated point clouds are used to evaluate the performance on 3D semantic segmentation with mIoU and mACC.

\vspace{-4pt}
\subsection{Experimental Results.}

\noindent \textbf{3D Semantic Segmentation.} Tab.~\ref{tab:quantitative} shows the evaluation on 3D semantic segmentation across three real-world indoor datasets with varying number of classes. We made the following observations: 1) 2D methods exhibit significantly inferior performance in a direct 3D referring evaluation compared to the 3D methods. This gap mainly arises from intermediate rendering step of 2D methods. Since the learnable features for each gaussian are linearly interpolated to render 2D feature map, transfer of CLIP's capability to each Gaussian is largely weakened during the training~\citep{jun2025dr}. 2) Our \ours demonstrates the best performance across all benchmarks with the shortest reconstruction time due to the feed-forward design. 3) The \textcolor{gray}{\textit{\textbf{gray}}}-colored columns indicate the performance of our \ours when the depth maps obtained from the sensors are used instead of the model predictions. It further boosts the segmentation performance by better aligning the Gaussians with surface of the scene.
4) By combining the real-time 2D models, our \ours-\textit{fast} shows nearly real-time reconstruction speed that is 5-6 FPS of per-frame processing time. Fig.~\ref{fig:qualitative} further presents the qualitative comparison on 3D semantic segmentation. Clustering-based methods~\cite{wu2024opengaussian, li2025instancegaussian} such as InstanceGaussian~\cite{li2025instancegaussian} demonstrate the high-quality of instance-level Gaussian clusters. However, they often misclassify instances, particularly when the objects are in close proximity to one another. Direct feature lifting methods~\cite{jun2025dr, cheng2024occam} also exhibits noisy segmentation results, particularly in background categories such as \textit{wall} and \textit{floor}. Our \ours demonstrates better segmentation quality compared to the baselines, while combining the ground-truth depths (\ours-D) further enhances the quality of the visualization.

\begin{table}[t]
    \renewcommand{\arraystretch}{1.5}
    \centering
    \footnotesize
    \resizebox{\linewidth}{!}{
    \begin{tabular}{l|cc|cc|cc}
    \toprule
    \multirow{3}{*}{\textbf{Method}} & \multicolumn{2}{c|}{\textbf{ScanNet++}~\cite{yeshwanth2023scannet++} \ding{220} \textbf{ScanNet}~\cite{dai2017scannet}} & \multicolumn{2}{c|}{\textbf{ScanNet}~\cite{dai2017scannet} \ding{220} \textbf{ScanNet++}~\cite{yeshwanth2023scannet++}} & \multicolumn{2}{c}{\textbf{ScanNet}~\cite{dai2017scannet} \ding{220} \textbf{Replica}~\cite{straub2019replica}} \\
    \cline{2-7}
     & \multicolumn{2}{c|}{19 classes} & \multicolumn{2}{c|}{20 classes} & \multicolumn{2}{c}{48 classes} \\
     & $\quad\quad$mIoU$\quad\;\;$ & mACC & $\quad\quad$mIoU$\quad\;\;$ & mACC & $\quad\quad$mIoU & mACC \\
     \midrule \noalign{\vskip -2pt}
     OpenGaussian~\cite{wu2024opengaussian} & $\quad\quad$22.52$\quad\;\;$ & 35.02  & $\quad\quad$25.65$\quad\;\;$ & 37.03 & $\quad\quad$11.59 & 18.12 \\
     Occam's LGS~\cite{cheng2024occam} & $\quad\quad$30.49$\quad\;\;$ & 57.91 & $\quad\quad$34.08$\quad\;\;$ & \cellcolor{red_background} \textbf{61.19} & $\quad\quad$\cellcolor{red_background} \textbf{16.19} & \cellcolor{red_background}\textbf{30.14} \\
     Dr. Splat~\cite{jun2025dr} & $\quad\quad$28.38$\quad\;\;$ & 55.85 & $\quad\quad$\cellcolor{red_background} \textbf{39.85}$\quad\;\;$ & 58.34 & $\quad\quad$\cellcolor{orange_background}\underline{14.47} & \cellcolor{orange_background}\underline{27.12} \\
     InstanceGaussian~\cite{li2025instancegaussian} & $\quad\quad$26.57$\quad\;\;$ & 48.63 & $\quad\quad$29.98$\quad\;\;$ & 47.47 & $\quad\quad$12.45 & 18.34 \\
     \midrule
     \textbf{EmbodiedSplat} (RGB) & $\quad\quad$\cellcolor{red_background} \textbf{45.32}$\quad\;\;$ & \cellcolor{orange_background} \underline{67.56} & $\quad\quad$30.65$\quad\;\;$ & 48.72 & $\quad\quad$ 9.88 & 17.72 \\
     \textbf{EmbodiedSplat}-\textit{fast} (RGB) &  $\quad\quad$\cellcolor{orange_background} \underline{41.24}$\quad\;\;$ & \cellcolor{red_background} \textbf{69.44} & $\quad\quad$\cellcolor{orange_background} \underline{34.60}$\quad\;\;$ & \cellcolor{orange_background} \underline{60.13} & $\quad\quad$ 10.45 & 19.45 \\
     \midrule
     \cellcolor{gray_background} \textbf{EmbodiedSplat} (RGB-D)  & $\quad\quad$\cellcolor{gray_background} 50.80$\quad\;\;$ & \cellcolor{gray_background} 73.15 & $\quad\quad$\cellcolor{gray_background} 44.14$\quad\;\;$ & \cellcolor{gray_background} 66.63 & $\quad\quad$\cellcolor{gray_background}11.42 & \cellcolor{gray_background} 20.10 \\
     \cellcolor{gray_background} \textbf{EmbodiedSplat}-\textit{fast} (RGB-D) & $\quad\quad$\cellcolor{gray_background} 47.59$\quad\;\;$ & \cellcolor{gray_background} 76.43 & $\quad\quad$ \cellcolor{gray_background}51.66$\quad\;\;$ & \cellcolor{gray_background} 77.81 & $\quad\quad$\cellcolor{gray_background} 14.38 & \cellcolor{gray_background} 23.92 \\
    \bottomrule
    \end{tabular}}
    \newline
    \vspace{2pt}
    \caption{\textbf{Quantitative comparisons on cross-domain 3D semantic segmentation across ScanNet~\cite{dai2017scannet}, ScanNet++~\cite{yeshwanth2023scannet++} and Replica~\cite{straub2019replica}.}}
    \label{tab:cross_domain_quantitative}
    \vspace{-13pt}
\end{table}

\smallskip
\noindent \textbf{Cross-domain 3D Semantic Segmentation.} Tab.~\ref{tab:cross_domain_quantitative} further evaluates the generalizability of our model in a cross-domain segmentation setting, where the model is trained on ScanNet and evaluated on ScanNet++ and vice versa. Our model shows strong semantics generalizability in ScanNet++ $\rightarrow$ ScanNet transfer with performance degradation remaining below 1 mIoU compared to ScanNet $\rightarrow$ ScanNet setting in Tab.~\ref{tab:quantitative}. In contrast, ScanNet $\rightarrow$ ScanNet++ shows a clear performance drop (-11.28 mIoU) compared to the in-distribution setting. This gap mainly arised from poor depth estimation. Since ScanNet++ often includes challenging regions for depth estimation such as \textit{ceilings} which are largely absent in ScanNet, the model trained on ScanNet frequently struggles to predict accurate depth maps when evaluated on ScanNet++. When this issue is mitigated with RGB-D inputs (\textcolor{gray}{\textit{\textbf{gray}}}-column), the performance becomes similar to the in-distribution setting (44.14 mIoU vs 44.03 mIoU). We also simulate a more challenging scenario, where the model is trained with ScanNet and evaluated on the Replica dataset (\textit{Real-2-Sim}). Due to the huge domain gap between the real-world and synthetic dataset, our \ours fails to achieve the best results compared to the per-scene optimization methods. These results are expected since the baseline methods are initialized with the ground-truth point clouds and optimized for each individual scene.
Nevertheless, our model achieves performance comparable to clustering-based baselines~\cite{li2025instancegaussian, wu2024opengaussian}. When combined with sensor-estimated depth maps, our \ours-\textit{fast} even attains results on par with feature-lifting methods~\cite{jun2025dr, cheng2024occam} such as Dr. Splat~\cite{jun2025dr}.

\begin{table}[t]
    \renewcommand{\arraystretch}{1.5}
    \centering
    \footnotesize
    \resizebox{\linewidth}{!}{
    \begin{tabular}{c|c|c|c|c|c|c}
    \toprule
     \multirow{3}{*}{\shortstack{\textbf{2D Features} \\ $\mathbf{s}^T_g$}} & \multirow{3}{*}{\shortstack{\textbf{3D Features} \\ $\hat{\mathbf{g}}^T_g$}} & \multicolumn{3}{c|}{\textbf{ScanNet}~\cite{dai2017scannet}} & \textbf{ScanNet200}~\cite{rozenberszki2022language} & \textbf{ScanNet++}~\cite{yeshwanth2023scannet++} \\
    \cline{3-7}
     & & 10 classes & 15 classes & 19 classes & 70 classes & 20 classes \\
     \cline{3-7}
     & & \multicolumn{5}{c}{mIoU} \\
     \midrule \noalign{\vskip -2pt}
     \ding{51} & \ding{56} & 48.93 & 48.38 & 45.09 & 29.56 & 40.86\\
     \midrule
     \ding{56} & \ding{51} & 49.24 & 48.78 & 45.39 & 30.36 & 41.36\\
     \midrule
     \ding{51} &  \ding{51} & \textbf{49.81} & \textbf{49.23} & \textbf{46.22} & \textbf{31.16} & \textbf{41.93} \\
    \bottomrule
    \end{tabular}}
    \newline
    \caption{\textbf{Ablations on 3D CLIP features $\hat{\mathbf{g}}^T_g$}}
    \label{tab:ablation_3d_clip}
    \vspace{-15pt}
\end{table}

\begin{table}[t]
    \renewcommand{\arraystretch}{1.5}
    \centering
    \footnotesize
    \resizebox{\linewidth}{!}{
    \begin{tabular}{l|c|c|c}
    \toprule
    \textbf{Cosine similarity} & \textbf{Time (ms)} & \textbf{Complexity} & \textbf{Note} \\
    \midrule
    Per-Gaussian & 14.35 & $O(MD)$ & \multirow{2}{*}{\shortstack{$M$ = 3.2M, $K$ = 8.7K, \\ $D$ = 768, $L$ = 6 }} \\
    \cmidrule{1-3}
    Codebook-based (\textit{Ours}) & \textbf{1.18} & $O(KD + M(L-1))$ &  \\
    \bottomrule
    \end{tabular}}
    \newline
    \caption{\textbf{Ablations on codebook-based cosine similarity}}
    \label{tab:ablation_cosine_similarity}
    \vspace{-15pt}
\end{table}

\begin{table}[t]
    \renewcommand{\arraystretch}{1.5}
    \centering
    \footnotesize
    \resizebox{\linewidth}{!}{
    \begin{tabular}{l|c|c|c|c}
    \toprule
    \textbf{Methods} & \textbf{Type / Feature dimension} & \textbf{Size (MB)} & \textbf{pretraining} & \textbf{information loss} \\
    \midrule
    LangSplat~\cite{qin2024langsplat} & Auto-encoder / 3 & \textbf{30} & \ding{51} & \ding{51} \\
    Dr. Splat~\cite{jun2025dr} & PQ Index / 130 & 173 & \ding{51} & \ding{51} \\
    Occam's LGS~\cite{cheng2024occam} & \ding{56} / 512 & 2295 & \ding{56} & \ding{56} \\
    \midrule
    \ours & CLIP Global Codebook / 10 & \underline{148} & \ding{56} & \ding{56} \\
    \bottomrule
    \end{tabular}}
    \newline
    \vspace{2pt}
    \caption{\textbf{Comparisons on memory size for semantic features.}}
    \label{tab:ablation_memory_size}
    \vspace{-10pt}
\end{table}

\subsection{Ablation Studies.}

\vspace{-2pt}
\noindent \textbf{Ablations on 3D CLIP features.} Tab.~\ref{tab:ablation_3d_clip} demonstrates the effectiveness of combining 3D geometric-aware CLIP features $\hat{\mathbf{g}}^T_g$, where it leads to performance improvement in mIoU across all indoor benchmarks (3rd row). The 2D CLIP feature $\mathbf{s}^T_g$ preserves rich semantic generalization but lacks 3D geometric priors as it is derived from 2D images. In contrast, the 3D CLIP feature $\hat{\mathbf{g}}^T_g$ encodes geometric priors from 3D point clouds, but loses part of the original semantic richness due to the inevitable information loss from the distillation process (\cf Eq.~\ref{eq:loss}). Combining both features allows mutual compensation between semantics and geometry to yield the best overall performance (3rd row).

\begin{table}[t]
    \centering
    \footnotesize
    \resizebox{0.8\linewidth}{!}{
    \begin{tabular}{l|c|c|c|c|c|c}
    \toprule
    \textbf{Dataset} & \textbf{Classes} & \textbf{Metric} & $L=2$ & $L=4$ & $L=6$ & $L=11$ \\
    \midrule
    ScanNet~\cite{dai2017scannet} & 19 classes & mIoU & 44.38 & 45.01 & \textbf{45.09} & 45.08 \\
    \bottomrule
    \end{tabular}}
    \newline
    \vspace{2pt}
    \caption{\textbf{Ablations on cache size $L$.}}
    \label{tab:ablation_cache_size}
    \vspace{-18pt}
\end{table}

\begin{figure}[t]
\centering
\includegraphics[width=1.0\linewidth]{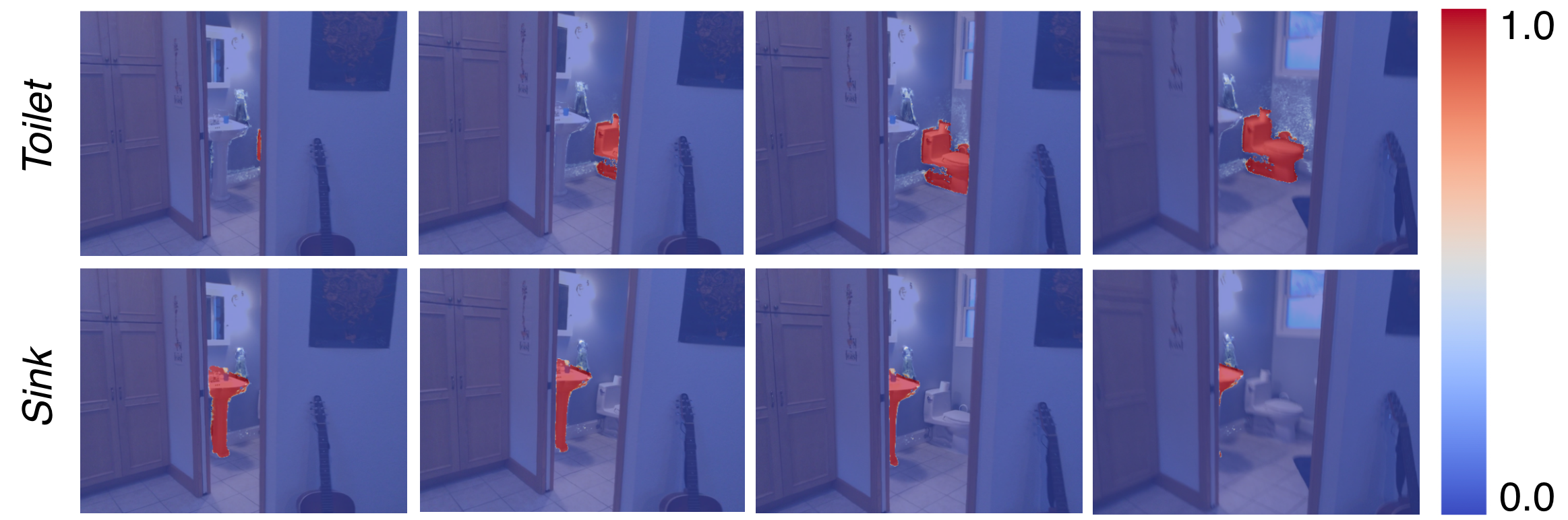}
  \vspace{-10pt}
  \caption{\textbf{2D-rendered object search.}}
  \label{fig:2d_sem}
  \vspace{-5pt}
\end{figure}

\begin{figure}[t]
\centering
\includegraphics[width=1.0\linewidth]{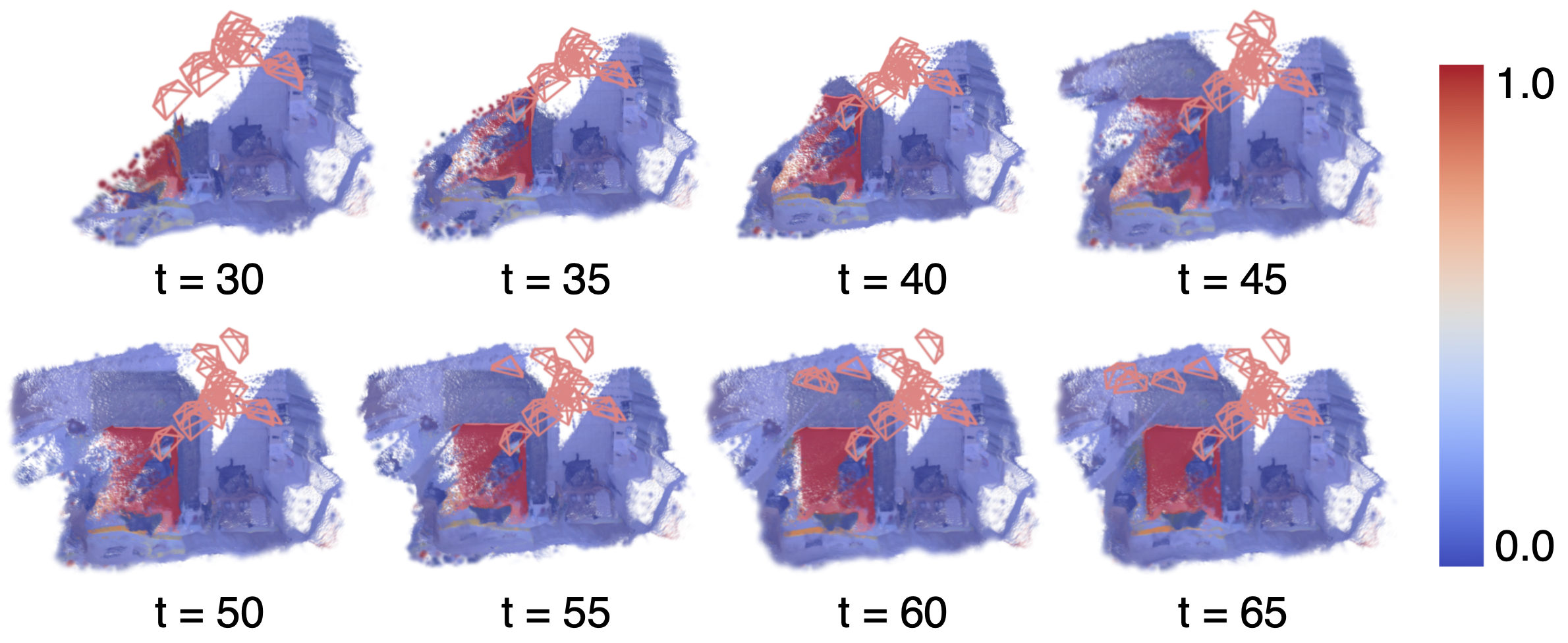}
  \vspace{-18pt}
  \caption{\textbf{Online 3D reasoning for class ``\textit{Bed}''.}}
  \label{fig:online_sem}
\end{figure}

\vspace{1mm}
\noindent \textbf{Ablations on codebook-based cosine similarity.} As shown in Tab.~\ref{tab:ablation_cosine_similarity}, our codebook-based cosine similarity (2nd row) significantly improves efficiency, achieving nearly 14× faster processing speed compared to the naive per-Gaussian cosine similarity computation (1st row). The experiment is conducted on \texttt{scene0000\_01} of the ScanNet dataset, where our \ours produces $M$ = 3.2M number of Gaussians while the global codebook stores $K$=8.7K of the instance-level CLIP features ($K \ll M$). 

\vspace{1mm}
\noindent \textbf{Comparisons on memory size for semantic features.} Tab.~\ref{tab:ablation_memory_size} compares the memory efficiency of our sparse coefficient field $\{\mathbf{I}, \Omega\}$ and CLIP global codebook with the other memory compression methods. The experiment is conducted on \texttt{scene0000\_01} of ScanNet dataset and the total memory consumption is estimated by summing the sizes of the compressor components (\eg, Auto-encoder~\cite{qin2024langsplat} and PQ Index~\cite{jun2025dr}) and the compressed semantic features. LangSplat~\cite{qin2024langsplat} achieves the lowest memory consumption by aggressively compressing the CLIP feature dimension (512 $\rightarrow$ 3) using a pretrained auto-encoder. However, it suffers from significant information loss~\cite{jun2025dr, li2025instancegaussian} due to the 
severe dimensionality reduction and requires an additional pretraining stage. Similarly, the PQ index of Dr.~Splat~\cite{jun2025dr} shares the same limitations.
Occam’s LGS preserves the full semantic capability of CLIP by attaching original features to each Gaussian, but incurs heavy memory overhead (2295 MB) due to the high dimensionality of CLIP.
In contrast, our \ours\ preserves the original CLIP information without requiring any pretraining stage and achieves high memory efficiency (148 MB) through a sparse coefficient field with a CLIP global codebook.

\vspace{1mm}
\noindent \textbf{Ablations on cache size $L$.} Tab.~\ref{tab:ablation_cache_size} provides the ablation on cache size $L$ in ScanNet~\cite{dai2017scannet} dataset. For each time step, we prune top $L-1$ indices by the confidence scores as described in Algorithm.~\ref{alg:sparse_coeff}. $L=2$ denotes that each Gaussian only select one instance CLIP feature with the highest weight from the codebook. Aggregating the multiple instance features from multi-view images leads to performance improvement ($L=2$ vs $L=4,6,11$).

\vspace{1mm}
\noindent \textbf{Qualitative results on 2D-rendered object search.} Although the main focus of our paper is scene understanding with direct 3D referring, we also present qualitative samples on 2D-rendered object search via heatmap visualization. Fig.~\ref{fig:2d_sem} shows that our \ours can produce multi-view consistent segmentation results.

\vspace{1mm}
\noindent \textbf{Qualitative results on online 3D reasoning.} Fig.~\ref{fig:online_sem} presents the Bird's-Eye View heatmap visualization for class ``\textit{Bed}'' during the online semantic reconstruction. Our \ours progressively searches for the ``\textit{Bed}'' along its exploration. More results are in video visualization showcased in our project website.

\vspace{-5pt}
\section{Conclusions}
\vspace{-5pt}

Immediate understanding of a 3D scene during online exploration is crucial for embodied tasks, where agents must incrementally construct and interpret their environment in real time. Although several works~\citep{xu2024embodiedsam, tang2025onlineanyseg, du2025moonseg3r} have proposed effective online 3D perception frameworks, the majority rely on point cloud representations, leaving 3DGS-based approaches largely underexplored. Only a few recent studies~\citep{katragadda2025online, zhou2025ea3d} have investigated online 3D understanding with 3DGS within SLAM-based frameworks~\citep{matsuki2024gaussian, gao2024hicom}. However, they struggle to achieve real-time reconstruction due to the computationally intensive per-scene optimization inherent in SLAM pipelines. Given the increasing adoption of 3DGS in the robotics community driven by its high-fidelity real-world digitization, developing an online 3D perception system with real-time speed under a 3DGS representation is both timely and essential. 

In this paper, we introduce the novel framework to endow the pretrained feed-forward 3DGS with online open-vocabulary capability. Our Sparse Coefficient Field, together with a CLIP-based Global Codebook enables memory-efficient semantic representations for each 3D Gaussian while fully preserving semantic richness of CLIP. Importantly, our approach eliminates the need for per-scene optimization or additional pretraining to obtain memory compressor, showing the clear advantage compared to previous semantic 3DGS. By incorporating the real-time 2D VLM, our \ours even achieves near real-time per-frame processing time, satisfying the practical requirements of embodied agent. We believe our framework provides the pioneering effort on 3DGS-based 3D perception model for embodied scenarios.

{   
    \small
    \bibliographystyle{ieeenat_fullname}
    \bibliography{main}
}
\clearpage
\setcounter{page}{1}
\maketitlesupplementary

\section*{Table of Contents}
\startcontents[appendices]
\printcontents[appendices]{l}{1}{\setcounter{tocdepth}{3}}

\section{Additional Explanations.} 

In this section, we supplement the additional details about our study including the experimental settings (Sec.~\ref{sec:supple_imple}) and further details for the overall framework of our \ours (Sec.~\ref{sec:supple_embodiedsplat}) and \oursfast (Sec.~\ref{sec:supple_embodiedsplat_fast}).

\subsection{Implementation Details.} 
\label{sec:supple_imple}

\vspace{1mm}
\noindent \textbf{Model \& Training Details.} \ours is built on top of pretrained FreeSplat++~\citep{wang2025freesplat++}, which is the feed-forward 3DGS that supports whole-scene 3D reconstruction. 3D sparse U-Net~\citep{choy20194d} with temporal-aware memory adapter~\citep{xu2024memory} are attached to obtain 3D geometric-aware features $\hat{\mathbf{g}}$ where Minkowski Res16UNet18A~\citep{choy20194d} is adopted as 3D U-Net. The training of \ours consists of two stages which are explained in Sec. \textcolor{cvprblue}{5}. Memory-based adapter is trained only in the second stage, where it is zero-initialized to enable a smooth fine-tuning, following~\citep{xu2024memory}. We adopt Adam~\cite{adam2014method} optimizer with an initial learning rate of 1$e$-4 followed by cosine decay for both stages.

\vspace{1mm}
\noindent \textbf{Keyframe Selection.} For the experiments, we select keyframes for each testing scene that cover the entire scene from the full set of multi-view images, following ~\cite{duzceker2021deepvideomvs, sayed2022simplerecon, hou2019multi}. Specifically, we calculate the pose distance between two cameras as follows:
\begin{equation}
\vspace{-2mm}
\label{eq:supple_pose_distance}
dist(\mathbf{T}_{\mathrm{rel}}) = \sqrt{||\mathbf{t}_{\mathrm{rel}}||^2 + \frac{2}{3}\mathrm{tr}(\mathbb{I} - \mathbf{R}_{\mathrm{rel}})},
\end{equation}
where $\mathbf{T}_{\mathrm{rel}} = [\mathbf{R}_{\mathrm{rel}}|\mathbf{t}_{\mathrm{rel}}]$ denotes the relative pose between two cameras. If the pose distance between current frame and last keyframe exceeds 0.1, we append the current frame to the list of keyframes. Collected keyframes are treated as streaming images for the experiment of our \ours. Tab.~\ref{tab:supple_bencmark_config} presents the list of testing scenes for each benchmark with the number of selected keyframes. Note that the same set of keyframes is used to train all per-scene optimized baselines~\citep{qin2024langsplat, jun2025dr, li2025instancegaussian, wu2024opengaussian, katragadda2025online, shi2024language, cheng2024occam} to ensure fair comparisons.

\begin{table}[t]
    \centering
    \footnotesize
    \resizebox{\linewidth}{!}{
    \begin{tabular}{c|l|c|c}
    \toprule
    \textbf{Dataset} & \textbf{Scene ID} & \textbf{Num of multi-view images} & \textbf{Num of keyframes} \\
    \midrule
    \multirow{10}{*}{ScanNet~\citep{dai2017scannet}} & \texttt{scene0000\_01} & 5920 & \textbf{363} \\
    & \texttt{scene0046\_00} & 2480 & \textbf{280} \\
    & \texttt{scene0079\_00} & 1196 & \textbf{119} \\
    & \texttt{scene0158\_00} & 1920 & \textbf{83} \\
    & \texttt{scene0316\_00} & 770 & \textbf{45} \\
    & \texttt{scene0389\_00} & 1415 & \textbf{205} \\
    & \texttt{scene0406\_00} & 1414 & \textbf{120} \\
    & \texttt{scene0521\_00} & 1566 & \textbf{84} \\
    & \texttt{scene0553\_00} & 1500 & \textbf{61} \\
    & \texttt{scene0616\_00} & 3027 & \textbf{226} \\
    \midrule
    \multirow{4}{*}{ScanNet++~\citep{yeshwanth2023scannet++}} & \texttt{09c1414f1b} & 2391 & \textbf{428} \\
    & \texttt{9071e139d9} & 1221 & \textbf{310} \\
    & \texttt{a24f64f7fb} & 652 & \textbf{138} \\
    & \texttt{c49a8c6cff} & 1188 & \textbf{253} \\
    \midrule
    \multirow{8}{*}{Replica~\citep{straub2019replica}} & \texttt{office0} & 900 & \textbf{230} \\
    & \texttt{office1} & 900 & \textbf{230} \\
    & \texttt{office2} & 900 & \textbf{230} \\
    & \texttt{office3} & 900 & \textbf{230} \\
    & \texttt{office4} & 900 & \textbf{230} \\
    & \texttt{room0} & 900 & \textbf{230} \\
    & \texttt{room1} & 900 & \textbf{230} \\
    & \texttt{room2} & 900 & \textbf{230} \\
    \bottomrule
    \end{tabular}}
    \newline
    \caption{\textbf{Configurations of the three benchmarks~\citep{dai2017scannet, yeshwanth2023scannet++, straub2019replica}.}}
    \label{tab:supple_bencmark_config}
    \vspace{-10pt}
\end{table}

\vspace{1mm}
\noindent \textbf{Point clouds Annotation.} To evaluate the performance on 3D semantic segmentation, we assign text labels to the ground-truth point clouds using the optimized semantic 3DGS, following~\citep{huang2024gaussianformer, huang2025gaussianformer}. Given $M$ semantic Gaussians and $C$ text labels, we first compute the per-Gaussian semantic logits $\mathbf{P} \in \mathbb{R}^{M \times C}$. To obtain the semantic logit for a point $\mathbf{p} \in \mathbb{R}^3$, we aggregate the semantic logits of multiple Gaussians weighted by their Mahalanobis distance~\citep{de2000mahalanobis} to $\mathbf{p}$. Specifically, the scaled semantic logit contributed by $i$-th Gaussian to point $\mathbf{p}$ is formulated as:
\begin{equation}
\vspace{-2mm}
\label{eq:supple_mahalanobis}
\hat{\mathbf{P}}_i = \mathrm{exp}(-\frac{1}{2}(\mathbf{p}-\mu_i)^{\top}\Sigma_i^{-1}(\mathbf{p}-\mu_i))\mathbf{P}_i,
\end{equation}
where $\mathbf{P}_i$, $\mu_i$ and $\Sigma_i$ are the semantic logits, 3D mean vector and 3D covariance of $i$-th Gaussian, respectively. The final semantic logits for a point $\mathbf{p}$ is then computed by summing the weighted logits from every neighboring Gaussians that are located sufficiently close to $\mathbf{p}$, as formulated in:
\begin{equation}
\vspace{-2mm}
\dot{\mathbf{P}} = \sum_{i \in \mathcal{N}(\mathbf{p})} \hat{\mathbf{P}}_i,
\end{equation}
where $\mathcal{N}(\mathbf{p})$ denotes the set of neighboring Gaussians of point $\mathbf{p}$. Annotated point clouds are compared to the ground-truth semantic labels to evaluate the 3D semantic segmentation performance. Since \ours and all of the baselines exploit the camera poses given by the dataset, the resulting 3DGS is already aligned with the ground-truth point clouds which enables the proper aggregation of 3DGS to given 3D points via Eq.~\ref{eq:supple_mahalanobis}.

\begin{figure}[t]
\centering
\includegraphics[width=1.0\linewidth]{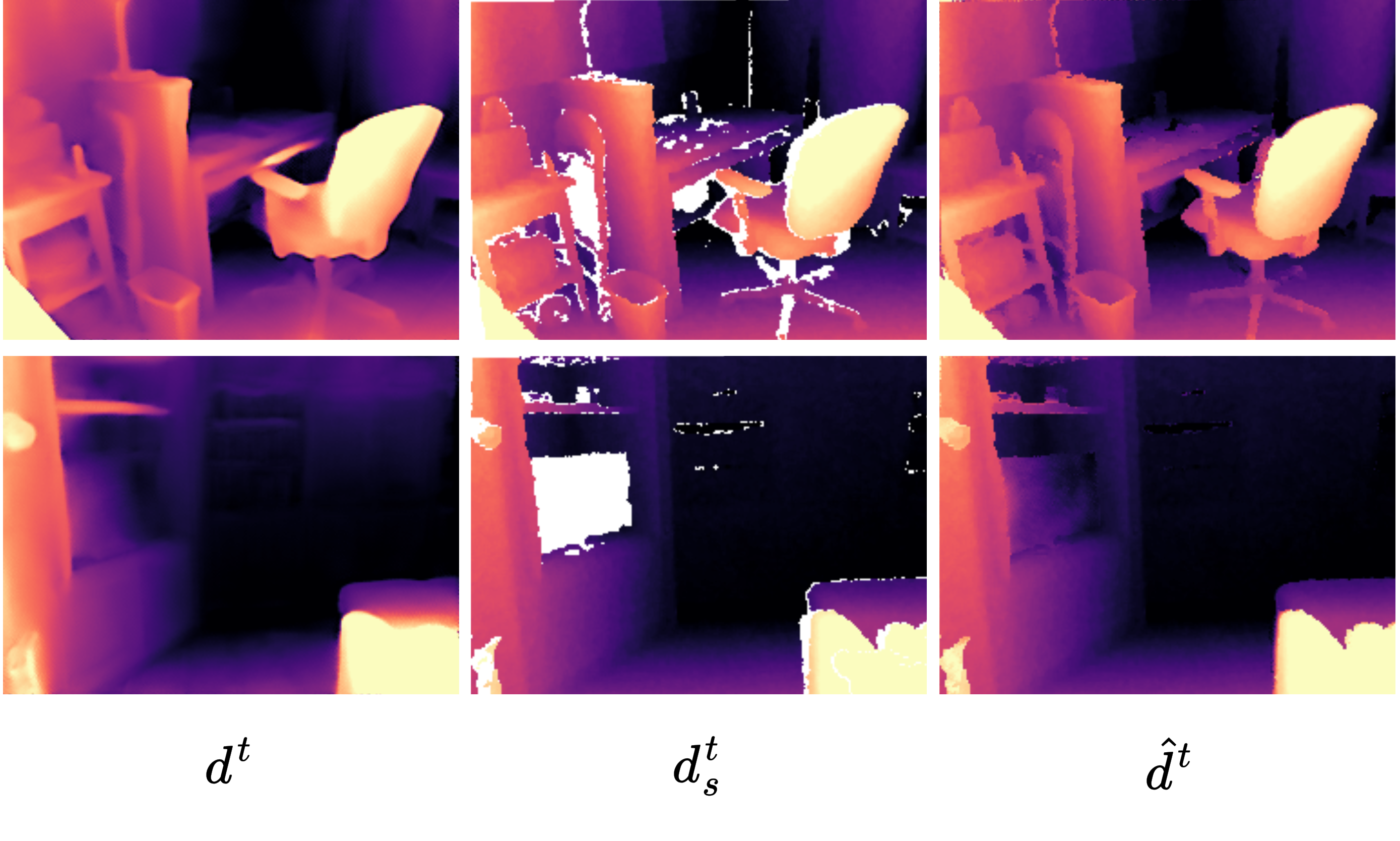}
  \vspace{-25pt}
  \caption{\textbf{Depth visualizations with RGB-D inputs.}}
  \label{fig:supple_depth_vis}
\end{figure}

\vspace{1mm}
\noindent \textbf{Using Ground-Truth Depth Maps.} Embodied agent equipped with depth sensor can acquire RGB-D inputs instead of RGB alone. Therefore, we also evaluate the performance of our \ours with RGB-D instead of using depth predictions from the model (\cf \textcolor{gray}{\textbf{\textit{gray}}}-colored rows in Tab. \textcolor{cvprblue}{1}-\textcolor{cvprblue}{2}). However, depths estimated from sensor often contains \textit{holes} which represents missing or invalid depth values. These typically arise from the surfaces that are reflective, transparent, too dark, too thin, or outside the valid measurement range. Hence, we fill the missing regions of the sensor depth $d^t_s$ with depth predictions $d^t$ from our model, formulated as:
\begin{equation}
\label{eq:supple_boraderfusion}
\footnotesize
\hat{d}^t(i) =
\begin{cases}
\displaystyle d^t_s(i),  
& \text{if } d^t_s(i) > 1e-3 \text{ and } d^t_s(i) < 10, \\ d^t(i), & \text{otherwise}
\end{cases}
\end{equation}
where $i \in \{1, \cdots, H \times W\}$. Fig.~\ref{fig:supple_depth_vis} shows the $d^t$, $d^t_s$ and $\hat{d}^t$, respectively. Sensor depths $d^t_s$ show large missing parts indicated with white, where those parts are filled by predicted depths, resulting in a smooth and complete depth maps $\hat{d}^t$.

\subsection{\ours}
\label{sec:supple_embodiedsplat}

The main objective of our \ours is to map the $T$ number of posed streaming images into open-vocabulary 3DGS which supports diverse perception tasks such as 1) 3D semantic segmentation, 2) 2d-rendered semantic segmentation and 3) noviel-view synthesis with depth rendering, while achieving near real time reconstruction speed. Here, we explain further details for inference pipeline of our \ours.

\vspace{1mm}
\noindent \textbf{Warm-up Stage.} Our \ours first collects the $N=30$ number of images and reconstructs the initial semantic 3DGS as warm-up stage. The warm-up stage is performed offline, requiring approximately 33–34 seconds in the \ours setting and 4–5 seconds in the \oursfast setting. Constructed semantic 3DGS serves as starting point which is then progressively expanded in an online manner as the model continues to explore the scene.

\begin{figure}[t]
\centering
\includegraphics[width=1.0\linewidth]{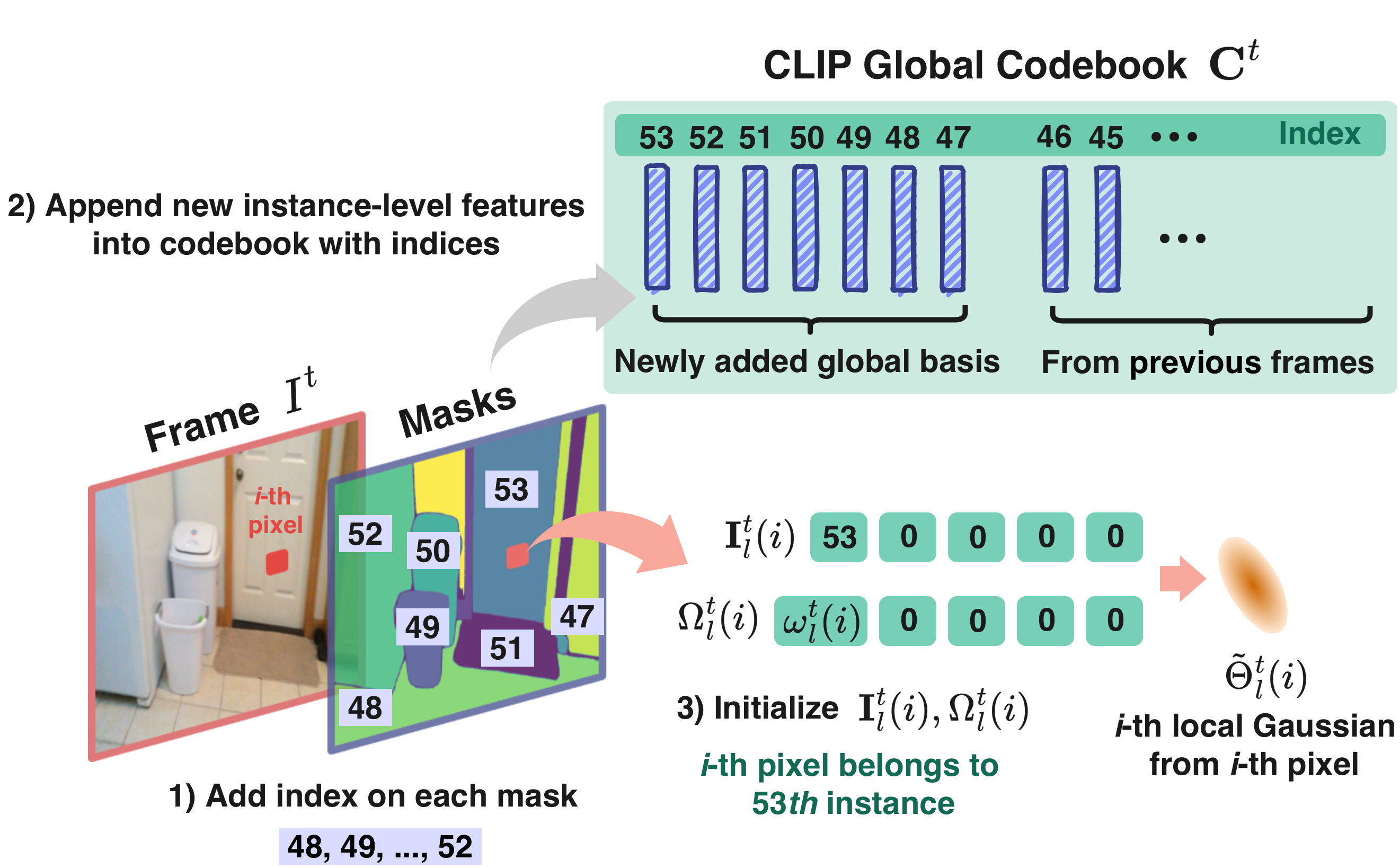}
  \vspace{-18pt}
  \caption{\textbf{Toy example of sparse coefficient field initialization at $i$th pixel.} \textbf{1)} We first add index to each instance-level mask. Index starts from 47 since the latest index in global codebook from previous time step is 46. \textbf{2)} Instance-level features extracted from current frame $\mathbf{I}^t$ are appended into codebook with attached indices, producing updated codebook $\mathbf{C}^t$. \textbf{3)} We initialize the index cache $\mathbf{I}^t_l(i)$ and weight cache $\Omega^t_l(i)$ of $i$-th local Gaussian aligned with $i$-th pixel. Since $i$-th pixel belongs to instance `53', index `53' is added to first entry of $\mathbf{I}^t_l(i)$: $\mathbf{I}^t_l(i, 0)=53$. Corresponding confidence value $\omega^t_l(i)$ is further inserted into first entry of $\Omega^t_l(i)$.}
  \label{fig:supplee_spc_init}
\end{figure}

\vspace{1mm}
\noindent \textbf{Local Semantic Gaussians Field.} Given the current frame $I^t$ with time step $t$, we select the $N=30$ past frames from the previous time steps $t-N$ to $t-1$. $N$ reference views and $I^t$ are converted to the local semantic Gaussians field $\bar{\Theta}^t_l = \{\mu^t_l, \omega^t_l, \mathbf{f}^t_l, \mathbf{I}^t_l, \Omega^t_l, \hat{\mathbf{g}}^t_l\} $ with updated global codebook $\mathbf{C}^t$ through the process described below:
\begin{itemize}
    \item \textbf{Local gaussian triplets $\{\mu^t_l, \omega^t_l, \mathbf{f}^t_l\}$}: We follow the FreeSplat++ to obtain the local gaussian triplets by leveraging the CNN-based network $\mathcal{E}$. Specifically, the backbone features of $I^t$ and $N$ reference views are extracted using a shared 2D backbone, after which a cost volume is constructed between them via plane sweep stereo~\cite{collins1996space, im2019dpsnet}. Obtained cost volume is then processed by UNet++~\cite{zhou2018unet++}-like decoder, outputting the depth map $d^t \in \mathbb{R}^{H \times W}$, pixel-wise Gaussian latents $\mathbf{f}^t_l \in \mathbb{R}^{H \times W \times D}$ and confidence scores $\omega^t_l \in \mathbb{R}^{H \times W}$ for each Gaussian latent. Finally, 2D pixels are unprojected to 3D space by using depth prediction $d^t$, resulting in 3D points $\mu^t_l \in \mathbb{R}^{H \times W \times 3}$. Here, $H \times W$ denotes the resolution of input images and $D$ represents feature dimension of $\mathbf{f}^t_l$. Kindly refer to FreeSplat++~\citep{wang2025freesplat++} for more details.
    \item \textbf{Sparse Coefficient Field $\{\mathbf{I}^t_l, \Omega^t_l\}$ and codebook $\mathbf{C}^t$}: We lift the original 2D CLIP features to each Gaussian through our novel sparse coefficient field, preserving both memory efficiency and full semantic capability of CLIP. We provide the illustration of toy example for local sparse coefficient field initialization in Fig.~\ref{fig:supplee_spc_init} for better understanding.
    \item \textbf{3D CLIP Features $\hat{\mathbf{g}}^t_l$}: Given the pixel-aligned Gaussian latents $\mathbf{f}^t_l$ and pixel-wise CLIP features $\mathbf{s}^t_l$, semantic-aware Gaussian latents $\mathbf{g}^t_l$ are obtained by doing $\mathbf{g}^t_l = \mathbf{f}^t_l + \mathrm{proj}(\mathbf{s}^t_l)$, where MLP layer $\mathrm{proj}(\cdot)$ is adopted to match the feature dimension between them. Then, $\mathbf{g}^t_l \in \mathbb{R}^{H \times W \times D}$ and 3D points $\mu^t_l \in \mathbf{R}^{H \times W \times 3}$ construct the local feature point clouds which is subsequently processed by 3D U-Net and memory-based adapter. Memory-based adapter retrieves the global Gaussian latents $\mathbf{f}^{t-1}_g$ and their 3D coordinates $\mu^{t-1}_g$ from the previous time step, and selects the latents that are spatially close to the local feature point clouds in 3D space. While the 3D U-Net processes the local point clouds as input, selected global latents are injected to its intermediate layers. This design enables the network to aggregate geometric priors not only from the local point clouds of the current frame, but also from the previously reconstructed global scene. Resulting 3D features $\hat{\mathbf{g}}^t_l \in \mathbb{R}^{H \times W \times D^s}$ compensate the 3D geometric prior to 2D CLIP features, leading to the clear performance improvement in 3D scene understanding (\cf Tab. \textcolor{cvprblue}{3}).  Kindly refer to~\cite{xu2024memory} for more details about memory-based adapter.
\end{itemize}

\vspace{1mm}
\noindent \textbf{Gaussians Fusion.} Obtained local semantic Gaussians $\bar{\Theta}^t_l$ are then fused with the global Gaussians $\bar{\Theta}^{t-1}_g$ to produce the updated global set $\bar{\Theta}^{t}_g$ at step $t$. Gaussian fusion is applied only to valid Gaussians pairs between the local and global sets where the valid pairs are determined according to the rules described below.

For $i$-th local Gaussian $\bar{\Theta}^t_l(i)$ which is aligned with $i$-th pixel of current frame $I^t$, we first obtain a set of Gaussians $\mathcal{S}^t_i$ from global set $\bar{\Theta}^{t-1}_g$ whose 3D coordinates project onto pixel $i$ in frame $I^t$. Subsequently, we search the valid match for $\bar{\Theta}^t_l(i)$ within the $\mathcal{S}^t_i$ based on the \textit{broader fusion} technique proposed by~\citep{wang2025freesplat++}:
\begin{equation}
\label{eq:supple_boraderfusion}
\footnotesize
m_i =
\begin{cases}
\displaystyle \arg\min_{j \in \mathcal{S}^t_i} d^{t}_{g}(j), 
& \text{if } d^{t}_{l}(i) - \min_{j \in \mathcal{S}^t_i} d^{t}_{g}(j) > -\delta, \\ \qquad
\varnothing, & \text{otherwise}
\end{cases}
\end{equation}
where $\delta$ is a threshold and $d^t_g(j)$ is the depth value of $j$-th global Gaussian in $\mathcal{S}^t_i$. Similarly, $d^t_l(i)$ is the predicted depth value of $i$-th local Gaussian on frame $I^t$. Local Gaussians which have no valid match ($\varnothing$) are directly appended to the global set without modification. In contrast, resulting valid Gaussians pairs $(i, m_i) \in \mathcal{P}^t$ are fused according to the following fusion rule:
\begin{subequations}\label{eq:supple_fusion}
\footnotesize
\begin{align}
\mu^t_g(m_i) &= \frac{\omega^t_l(i)\,\mu^t_l(i) + \omega^{t-1}_g(m_i)\,\mu^{t-1}_g(m_i)}{\omega^t_l(i) + \omega^{t-1}_g(m_i)}, \\
\omega^t_g(m_i) &= \omega^t_l(i) + \omega^{t-1}_g(m_i), \\
\mathbf{f}^t_g(m_i) &= \mathrm{GRU}\!\big(\mathbf{f}^t_l(i),\, \mathbf{f}^{t-1}_g(m_i)\big), \\
\hat{\mathbf{g}}^t_g(m_i) &= \mathrm{GRU}\!\big(\hat{\mathbf{g}}^t_l(i), \hat{\mathbf{g}}^{t-1}_g(m_i)\big), \\
\mathbf{I}^t_g(m_i), \Omega^t_g(m_i) &= \mathcal{F}\!\big(\mathbf{I}^t_l(i), \Omega^{t-1}_g(m_i)\big).
\end{align}
\end{subequations}
Here, $\mathcal{F}(\cdot, \cdot)$ denotes our proposed online fusion algorithm for sparse coefficient field which is described in Algorithm. \textcolor{cvprblue}{1}. Next, we provide the further details about our sparse coefficient field with its online fusion algorithm $\mathcal{F}(\cdot, \cdot)$.

\vspace{1mm}
\noindent \textbf{Motivation of Sparse Coefficient Field.} At the beginning of Sec.~\ref{sec:method_embodedsplat}, we introduce a naive approach for lifting the pixel-level 2D CLIP features to each Gaussian. Eq.~\textcolor{cvprblue}{3} subsequently fuses the local CLIP features with paired global features based on the confidence-weighted average. It can be rewritten as:
\begin{equation}
\label{eq:supple_clip_fusion}
\mathbf{s}^t_g(m_i) = \alpha\cdot\mathbf{s}^t_l(i) + (1-\alpha)\cdot\mathbf{s}^{t-1}_g(m_i),
\end{equation}
where $\alpha = \frac{\omega^t_l(i)}{\omega^t_l(i) + \omega^{t-1}_g(m_i)}$ denotes the coefficient of linear combination. During the exploration from step 1 to $T$, the 2D CLIP features of $m_i$-th global Gaussian may be produced by fusing the $k \geq 1$ number of local features by repeating the weighted-sum of Eq.~\ref{eq:supple_clip_fusion} for $k-1$ times. For example, if $k=3$, the final CLIP features $\mathbf{s}^T_g(m_i)$ of $m_i$-th global Gaussian can be formulated as:
\begin{equation}
\label{eq:supple_clip_fusion_rethink}
\footnotesize
\begin{aligned}
\mathbf{s}^T_g(m_i) &= (1-\alpha_2)\!\big((1-\alpha_1)\cdot\mathbf{s}_l(m_i, 0) + \alpha_1\cdot\mathbf{s}_l(m_i, 1)\big) \\ 
& \qquad + \alpha_2\cdot\mathbf{s}_l(m_i, 2) \\
&= (1-\alpha_2)(1-\alpha_1)\cdot\mathbf{s}_l(m_i, 0) + (1-\alpha_2)\alpha_1\cdot\mathbf{s}_l(m_i, 1) \\ 
& \qquad + \alpha_2\cdot\mathbf{s}_l(m_i, 2) \\
&= \sum^{k-1}_{j=0}\beta_j\cdot\mathbf{s}_l(m_i, j), \quad \text{$k$=3,}
\end{aligned}
\end{equation}
where $\mathbf{s}_l(m_i)$ is the set of $k$ CLIP features collected across $k$ different views to produce $\mathbf{s}^T_g(m_i)$. $\alpha_i$ denotes the coefficent of $i$-th fusion operation from Eq.~\ref{eq:supple_clip_fusion}. Eq.~\ref{eq:supple_clip_fusion_rethink} shows that $\mathbf{s}^T_g(m_i)$ can be represented as linear combination of $\mathbf{s}_l(m_i)$, where $\sum^{k-1}_{j=0}\beta_j = 1$. Here, $\mathbf{s}_l(m_i)$ serves as local basis for $\mathbf{s}^T_g(m_i)$ and $\beta$ denotes coefficients for each local basis. Note that Eq.~\ref{eq:supple_clip_fusion_rethink} can be generalized to any number of $k$.

The main idea of our sparse coefficient field $\{\mathbf{I}, \Omega\}$ is to leverage the weighted combination of local basis in Eq.~\ref{eq:supple_clip_fusion_rethink} to store per-Gaussian CLIP features in memory-efficient manner. \textit{\textbf{Pixel-level to Instance-level}}: We replace the \textit{pixel-level} local basis $\mathbf{s}_l(m_i)$ with \textit{instance-level} representations and allow every Gaussians to share the same global basis dictionary $\mathbf{C}$ through lookup indices. Specifically, index cache $\mathbf{I}$ stores the indices where they are linked with corresponding instance-level CLIP features stacked in the global codebook $\mathbf{C}$. These instance features serve as global basis function for semantic features of each Gaussian. Weight cache $\Omega$ further stores the coefficients $\beta$ to retrieve the contribution of each basis. Finally, the original CLIP features $\mathbf{s}^T_g(m_i)$ can be restored by performing linear combination with sparse coefficient field via Eq.~\textcolor{cvprblue}{4}.

\vspace{1mm}
\noindent \textbf{Online fusion of Sparse Coefficient Field.}
The number of local basis $k$ in Eq.~\ref{eq:supple_clip_fusion_rethink} may increase if exploration continues with collecting more views while the coefficients list $\beta$ may be also updated based on the confidence-weighted average of Eq.~\ref{eq:supple_clip_fusion}. Our online fusion Algorithm.~\textcolor{cvprblue}{1} tracks the growth of local basis $\mathbf{s}^T_g(m_i)$ and updates of coefficients $\beta$ by using our sparse coefficient field along the exploration. To aid the understanding of our online update process, Fig.~\ref{fig:supple_online_fuse_toy-5} presents a toy example that illustrates Algorithm.~\textcolor{cvprblue}{1}. It continuously collects the new evidence from incoming view by accumulating the index of new local basis into global index cache. Coefficients for each local basis in weight cache $\Omega$ are also updated based on the confidence-weighted average. To limit the maximum number of local basis for each Gaussian, Algorithm.~\textcolor{cvprblue}{1} only keeps the top $L-1$ basis with the highest coefficients. It removes the basis with low confidence scores, thereby sharpening the semantic Gaussian representations and preserving the memory efficiency.

\vspace{1mm}
\noindent \textbf{Post Refinement.} After processing all of the $T$ images, we perform \textit{floater removal} proposed by~\citep{wang2025freesplat++} as a post refinement. It effectively removes the floater which improves the rendering quality while only taking 2-3 seconds in both \ours and \oursfast.

\subsection{\oursfast.}
\label{sec:supple_embodiedsplat_fast}

\oursfast is introduced as a faster and lighter variant of our \ours to satisfy the near real-time per-frame processing time. Three modifications are adopted to \ours: 1) Replacing the heavy 2D VLM into real-time models, 2) removing the 3D CLIP features to improve the inference speed and 3) proposing the efficient 3D search strategy to obtain per-Gaussian cosine similarities faster. Since \oursfast doesn't use 3D modules such as 3D U-Net and memory adapter, and relies solely on 2D CLIP features with the sparse coefficient field, it can be directly built on top of pretrained feed-forward 3DGS~\cite{wang2025freesplat++} without any additional training. This training-free approach allows the direct combination with various types of 2D VLMs, highlighting the broad applicability of \oursfast.

\vspace{1mm}
\noindent \textbf{Codebook-based Cosine Similarity.} In \oursfast, we further introduce the efficient inference strategy to compute the per-Gaussian cosine similarities. The main idea is to precompute the cosine similarities between instance-level CLIP features stored in codebook $C^T$ and text prompts, and then reuse these values to obtain the per-Gaussian cosine similarities through the linear combination derived from sparse coefficient field (\cf Eq. \textcolor{cvprblue}{8}). Since the number of features stored in global codebook is much smaller than the total number of Gaussians (\cf Tab.~\ref{tab:supple_exp_memory}), it significantly improves the 3D search latency. Specifically, our codebook-based cosine similarity results in $O\!\big(KD + M(L-1)\big)$ complexity while naive computation of per-Gaussian cosine similarities incurs $O(MD)$, where $K$ denotes the codebook size, $M$ is the number of Gaussians, $D$ is the CLIP dimension and $L$ is the cache size. We show the complexity comparison between these two approaches:
\begin{equation}
\small
\begin{aligned}
O\!\big(KD + M(L-1)\big) \approx O(KD + M) \\
\ll O(MD + M) \approx O\!\big(M(D+1)\big) \approx O(MD),
\end{aligned}
\end{equation}
where $K \ll M$ and $L=6$. Tab. \textcolor{cvprblue}{4} further reports the real inference time comparisons on NVIDIA 6000 Ada GPU.

\begin{figure}[t]
\centering
\includegraphics[width=1.0\linewidth]{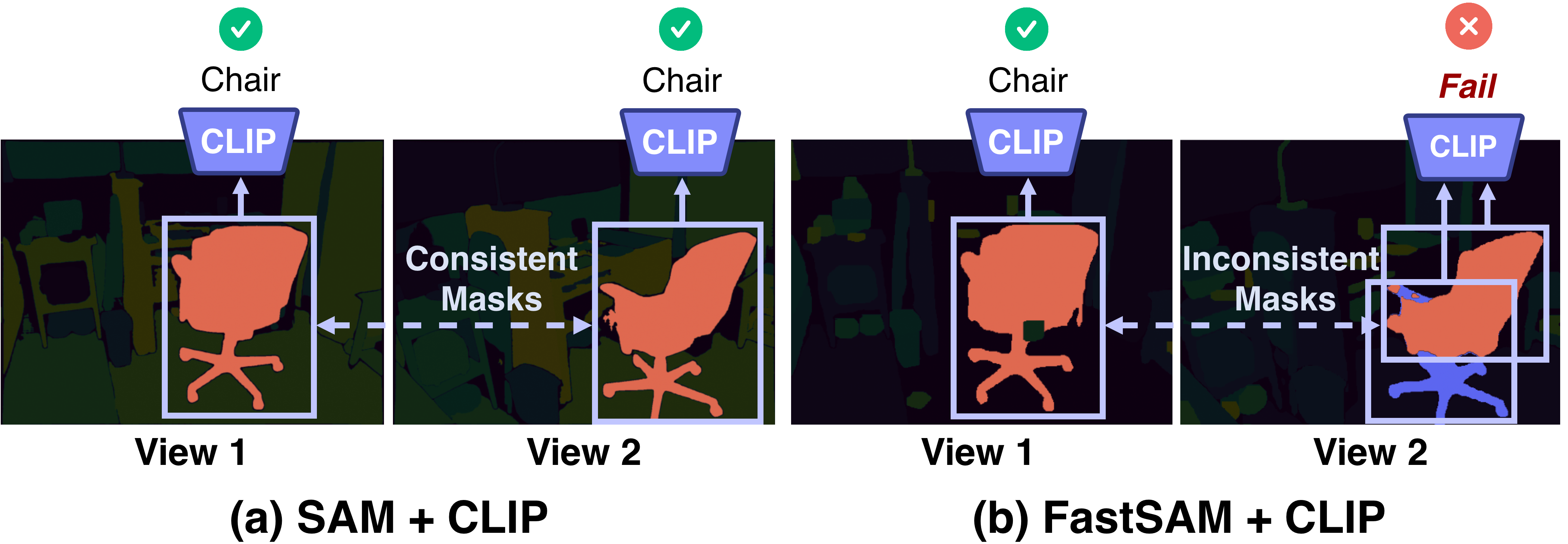}
  \vspace{-18pt}
  \caption{\textbf{Multi-view inconsistent masks from FastSam~\citep{zhao2023fast}.}}
  \label{fig:supple_2dvlm}
  \vspace{-5pt}
\end{figure}

\begin{table}[t]
    \centering
    \footnotesize
    \resizebox{\linewidth}{!}{
    \begin{tabular}{l|c|c}
    \toprule
    \textbf{2D VLM Configurations} & \textbf{Contextual Information} & \textbf{Speed} \\
    \midrule
    SAM~\citep{kirillov2023segment, qin2024langsplat} + CLIP~\citep{radford2021learning} & \ding{56} & 23220 ms \\
    FastSAM~\cite{zhao2023fast} + CLIP~\citep{radford2021learning} & \ding{56}  & 31.5 ms \\
    FastSAM~\citep{zhao2023fast} + OpenSeg~\citep{ghiasi2022scaling} (\textit{Ours}) & \ding{51} &  991.3 ms \\
    FastSAM~\citep{zhao2023fast} + Mask-Adapter~\citep{li2025mask} (\textit{Ours}) & \ding{51} & 43.3 ms \\
    \bottomrule
    \end{tabular}}
    \newline
    \caption{\textbf{Comparisons on different 2D VLM configurations.}}
    \label{tab:supple_2dvlm_comparisons}
    \vspace{-10pt}
\end{table}

\section{Discussions.}

In this section, we supplement additional discussions: Sec.~\ref{sec:supple_discussions_2dvlm} explores the diverse configurations of 2D VLM in the embodied scenarios, justifying our choice of 2D models within \ours framework. Sec.~\ref{sec:supple_discussions_baselines} explains additional works that are closely related to our study but are not included in the main paper. Finally, Sec.~\ref{sec:supple_limitations} discusses the limitation of our \ours.

\subsection{2D VLM}
\label{sec:supple_discussions_2dvlm}
Here, we discuss the motivation behind our choice of a 2D VLM. Most of the exisitng semantic 3DGS~\citep{qin2024langsplat, wu2024opengaussian, li2025instancegaussian, cheng2024occam, jun2025dr} exploit the combination of SAM~\citep{kirillov2023segment} and CLIP~\citep{radford2021learning} to extract the open-vocabulary cues from 2D images. Specifically, LangSplat~\citep{qin2024langsplat} customizes the original SAM to output three levels of masks with different granularity by leveraging the SAM's inherent property of producing coarse-to-fine segmentations. After obtaining the instance-level masks, corresponding image patches are cropped and subsequently fed into CLIP to generate open-vocabulary features. Following works~\cite{wu2024opengaussian, li2025instancegaussian, cheng2024occam, jun2025dr} adopt the same strategy by simply adopting the same customized SAM from~\citep{qin2024langsplat}. However, this simple combination of SAM and CLIP has two limitations in the embodied scenarios: 1) Customized SAM~\citep{qin2024langsplat} + CLIP incurs prolonged inference time (23.22 seconds per image) which hinders the near real-time capability of the model. Main bottleneck arises from the heavy post-processing of customized SAM to obtain clean object-level masks for the entire image. 2) Feeding the cropped image patches to CLIP ignores the background part of the objects which is crucial to understand the contextual information.

First limitation can be easily addressed by simply adopting the real-time 2D models such as FastSAM~\cite{zhao2023fast}. However, combining the FastSAM with CLIP further exacerbates the second issue. We empirically find that FastSAM frequently fails to generate masks with a consistent level of granularity; for example, it may produce an object-level mask in one view while generating part-level masks in another view with same object. Since the cropped patches of part-level masks lack the surrounding regions necessary to preserve object-level semantics, CLIP often produces incorrect semantic predictions for these patches which is illustrated in Fig.~\ref{fig:supple_2dvlm}. To alleviate this issue, we adopt the combination of FastSAM and pixel-level CLIP models such as OpenSeg~\cite{ghiasi2022scaling} and LSeg~\cite{li2022language}. Because these models operate on the full image, each pixel-level CLIP feature inherently preserves global contextual information. The resulting per-pixel features are then pooled using the masks produced by FastSAM to obtain instance-level representations. To further improve the inference speed, we adopt Mask-Adpater~\cite{li2025mask} into \oursfast, which pools the instance-level CLIP features from MaskCLIP~\citep{zhou2022maskclip}-based architecture. Tab.~\ref{tab:supple_2dvlm_comparisons} summarizes the comparisons among different 2D VLM configurations. Furthermore, Tab.~\ref{tab:supple_exp_3dss} and Sec.~\ref{sec:supple_exp_3dss} examines the performance of diverse 2D VLM within our \ours framework.

\subsection{More Baselines.}
\label{sec:supple_discussions_baselines}
We discuss more recent baselines or concurrent works in semantic 3DGS which are closely related to our study but not mentioned in the main paper. Tab.~\ref{tab:supple_baseline_comparisons} shows the overall comparison between our \ours and the additional baselines which are described next.

\vspace{1mm}
\noindent \textbf{3D methods.} In the main paper, we adopt clustering-based methods~\citep{wu2024opengaussian, li2025instancegaussian} and feature-lifting approaches~\citep{cheng2024occam, jun2025dr} as a \textit{3D} baseline that supports direct 3d referring. Here, we introduce more baselines which fall in this \textit{3D} category. VoteSplat~\citep{jiang2025votesplat} is another clustering-based method where it groups the gaussians by exploiting Hough voting algorithm to reduce the training cost. Since their code is not publicly available, we do not compare the performance with VoteSplat. LUDVIG~\citep{marrie2025ludvig} is another recent work in feature-lifting approach, where they directly lift the pixel-wise CLIP features into per-scene optimized 3DGS without feature distillation process. Specifically, they collect the multiple CLIP features for each Gaussian across multi-view images and aggregate them based on the rendering weights obtained from rasterization function. CF$^3$~\citep{lee2025cf3} follows the LUDVIG to directly bind the 2D features into 3DGS while more focusing on reducing the number of Gaussians. Since the way they lift the features are highly overlapped with Occam's LGS~\citep{cheng2024occam}, we do not include LUDVIG and CF$^3$ in Tab.~\textcolor{cvprblue}{1} of the main paper. However, Tab.~\ref{tab:supple_exp_3dss} provides the further comparison with them on 3D semantic segmentation.

\begin{table}[t]
    \centering
    \footnotesize
    \resizebox{\linewidth}{!}{
    \begin{tabular}{l|c|c|c|c}
    \toprule
    \textbf{Method} & \textbf{Venue} & \textbf{Generalizable} & \textbf{Online} & \textbf{Whole-Scene} \\
    \midrule
    VoteSplat~\cite{jiang2025votesplat} & ICCV'25 & \ding{56} & \ding{56} & \ding{51} \\
    LUDVIG~\cite{marrie2025ludvig} & ICCV'25 & \ding{56} & \ding{56} & \ding{51} \\
    CF$^3$~\cite{lee2025cf3} & ICCV'25 & \ding{56} & \ding{56} & \ding{51} \\
    \midrule
    LSM~\cite{fan2024large} & NeurIPS'24 & \ding{51} & \ding{56} & \ding{56} \\
    OVGaussian~\cite{chen2024ovgaussian} & arXiv'24 & \ding{51} & \ding{56} & \ding{56} \\
    SLGaussian~\cite{chen2025slgaussian} & arXiv'24 & \ding{51} & \ding{56} & \ding{56} \\
    GSemSplat~\cite{wang2024gsemsplat} & arXiv'24 & \ding{51} & \ding{56} & \ding{56} \\
    SemanticSplat~\cite{li2025semanticsplat} & arXiv'25 & \ding{51} & \ding{56} & \ding{56} \\
    UniForward~\cite{tian2025uniforward} & arXiv'25 & \ding{51} & \ding{56} & \ding{56} \\
    Gen-LangSplat~\cite{saxena2025gen} & arXiv'25 & \ding{51} & \ding{56} & \ding{56} \\
    SIU3R~\cite{xu2025siu3r} & NeurIPS'25 & \ding{51} & \ding{56} & \ding{56} \\
    \midrule
    EA3D~\cite{zhou2025ea3d} & NeurIPS'25 & \ding{56} & \ding{51} & \ding{51} \\
    \midrule
    \cellcolor{green_background} \textbf{\ours} (\textit{Ours}) & \cellcolor{green_background} - & \cellcolor{green_background} \ding{51} & \cellcolor{green_background} \ding{51} & \cellcolor{green_background} \ding{51} \\
    \bottomrule
    \end{tabular}}
    \newline
    \caption{\textbf{Comparison between \ours and additional baselines.}}
    \label{tab:supple_baseline_comparisons}
\end{table}

\vspace{1mm}
\noindent \textbf{Feed-forward semantic 3DGS.} LSM~\citep{fan2024large} is a pioneering work that introduces semantic feed-forward 3DGS. They add an additional semantic head on the feed-forward 3DGS. 2D CLIP features obtained from multi-view images are then distilled into feed-forward model through semantic head via 2D rendering function. Although effective, LSM only performs with only two or a few input views, lacking the capability to understand the whole scene. Furthermore, LSM focuses on understanding the scene by rendering the 2D feature maps rather than directly referring the 3D Gaussians. Since our study focuses on whole-scene understanding with direct 3D inference which is crucial in embodied scenarios, we do not include LSM as baseline in Tab.~\textcolor{cvprblue}{1}. Several literatures~\cite{tian2025uniforward, li2025semanticsplat, wang2024gsemsplat, chen2025slgaussian, chen2024ovgaussian, saxena2025gen, xu2025siu3r} follow the similar framework with LSM, proposing diverse variants of semantic feed-forward 3DGS. However, 1) all of them do not provide the open-sourced code except for SIU3R~\citep{xu2025siu3r}. 2) They don't address the whole-scene semantic reconstruction. 3) Finally, they only discuss the offline setting, solely relying on pre-collected multi-view images which deviates from embodied scenarios.

\begin{table*}[t]
    \renewcommand{\arraystretch}{1.5}
    \centering
    \footnotesize
    \resizebox{\linewidth}{!}{
    \begin{tabular}{l|c|cc|cc|cc|cc|c}
    \toprule
    \multirow{3}{*}{\textbf{Method}} & \multirow{3}{*}{\textbf{Inputs}} & \multicolumn{6}{c|}{\textbf{ScanNet}~\cite{dai2017scannet}} & \multicolumn{2}{c|}{\textbf{ScanNet200}~\cite{rozenberszki2022language}} & \multirow{3}{*}{\textbf{\textcolor{red}{Online} / \textcolor{blue}{Offline}}} \\
    \cline{3-10}
     & & \multicolumn{2}{c|}{10 classes} & \multicolumn{2}{c|}{15 classes} & \multicolumn{2}{c|}{19 classes} & \multicolumn{2}{c|}{70 classes} &   \\
    & & mIoU & mACC & mIoU & mACC & mIoU & mACC & mIoU & mACC &  \\
     \midrule \noalign{\vskip -2pt}
     Occam's LGS~\cite{cheng2024occam} & RGB & 42.14 & 70.28 & 35.04 & 63.71 & 30.49 & 57.91 & 20.32 & 40.49 & \textbf{\textcolor{blue}{Offline}} \\
     Dr. Splat~\cite{jun2025dr} & RGB & 39.21 & 66.66 & 31.84 & 60.58 & 28.38 & 55.85 & 19.29 & 33.84 & \textbf{\textcolor{blue}{Offline}} \\
     LUDVIG~\cite{marrie2025ludvig} & RGB & 41.11 & 68.34 & 33.73 & 62.90 & 29.34 & 56.98 & 21.23 & 39.87 & \textbf{\textcolor{blue}{Offline}} \\
     CF$^3$~\cite{lee2025cf3} & RGB & 38.14 & 65.13 & 30.13 & 59.13 & 26.34 & 52.43 & 18.23 & 30.11 & \textbf{\textcolor{blue}{Offline}} \\
     \midrule
     \cellcolor{green_background} EmbodiedSplat (SAM~\citep{kirillov2023segment, qin2024langsplat} + CLIP~\citep{radford2021learning}) & RGB & 45.56 & 75.13 & 37.90 & 66.82 & 33.12 & 59.03 & 21.98 & 41.11 & \textbf{\textcolor{red}{Online}} \\
     \cellcolor{green_background} EmbodiedSplat (FastSAM~\citep{zhao2023fast} + LSeg~\citep{li2022language}) & RGB & 57.48 & 77.63 & 51.84 & 68.78 & 42.23 & 58.12 & 23.13 & 35.18 & \textbf{\textcolor{red}{Online}} \\
     \cellcolor{green_background} EmbodiedSplat (FastSAM~\citep{zhao2023fast} + OpenSeg~\citep{ghiasi2022scaling}) & RGB & 49.81 & 76.13 & 49.23 & 75.47 & 46.22 & 70.37 & 31.16 & 48.38 & \textbf{\textcolor{red}{Online}} \\
     \midrule
     EmbodiedSAM~\cite{xu2024embodiedsam} & RGB-D & 52.13 & 78.13 & 50.98 & 77.81 & 48.11 & 71.45 & 33.11 & 48.14 & \textbf{\textcolor{red}{Online}} \\
     OpenScene~\cite{peng2023openscene} & Point Cloud, RGB-D & 54.56 & 80.45 & 53.74 & 79.85 & 50.71 & 72.75 & 33.84 & 50.06 & \textbf{\textcolor{blue}{Offline}} \\
     \midrule
     \cellcolor{green_background} EmbodiedSplat (FastSAM~\citep{zhao2023fast} + OpenSeg~\citep{ghiasi2022scaling}) & RGB-D & 57.41 & 82.45 & 55.18 & 80.27 & 52.12 & 75.66 & 34.75 & 52.36 & \textbf{\textcolor{red}{Online}} \\
    \bottomrule
    \end{tabular}}
    \newline
    \caption{\textbf{Additional comparisons on 3D Semantic Segmentation in ScanNet~\cite{dai2017scannet} and ScanNet200~\cite{rozenberszki2022language} datasets.}}
    \label{tab:supple_exp_3dss}
    \vspace{-25pt}
\end{table*}

\vspace{1mm}
\noindent \textbf{SLAM + Semantic 3DGS.} In contrast to aforementioned baselines, this category addresses the online reconstruction of semantic 3DGS by exploting the SLAM pipeline. Online-LangSplat~\cite{katragadda2025online} falls within this category where it combines the language embeddings into MonoGS~\cite{matsuki2024gaussian} which is 3DGS-based SLAM. EA3D~\cite{zhou2025ea3d} further proposes the advanced online framework based on HiCOM~\cite{gao2024hicom} which is the 4DGS-based SLAM. They improve the multi-view consistency among the 2D feature maps by exploiting the matching distributions between two adjacent frames. However, Online-LangSplat and EA3D both still require the per-scene optimization, which prevents them from generalizing to novel scenes in near real time. Furthermore, they distill the feature of 2D models into 3DGS via rendering function, inherently limiting the framework from supporting direct 3D referring. Note that the code of EA3D is not publicly available yet.

\subsection{Limitations}
\label{sec:supple_limitations}

Our \ours is built on top of pretrained FreeSplat++~\citep{wang2025freesplat++}. Hence, it inherits the limitation of FreeSplat++: when the feed-forward 3DGS fails to reconstruct the scene accurately, the resulting semantic Gaussian field becomes correspondingly noisy. We provide several examples where \ours fails to build clean semantic Gaussians due to the inaccurate 3DGS reconstruction.

\vspace{1mm}
\noindent \textbf{Out-of-Distribution Scenarios.} This is shown in Tab.~\textcolor{cvprblue}{2} of the main paper where the \ours trained in ScanNet~\citep{dai2017scannet} dataset fails to outperform the baselines in Replica~\cite{straub2019replica} due to the huge domain gap between real-world scenes and synthetic scenes. Since feed-forward 3DGS is overfitted to the real-world domain, it fails to perform accurate 3D reconstruction in the synthetic domain. Inaccurate 3DGS reconstruction leads to noisy lifting of semantic features to each Gaussian, finally resulting in low 3D segmentation performance.  

\vspace{1mm}
\noindent \textbf{Inaccurate Depth Estimation.} We discuss the inaccurate depth estimation case in Sec.~\ref{sec:exp} of the main paper. If the model faces difficult regions for depth estimation such as \textit{ceilings} or \textit{transparent backgrounds}, and these cases are largely absent from the training dataset, feed-forward 3DGS tends to fail in producing high quality depth predictions. This is shown in the performance drop of Tab.~\textcolor{cvprblue}{2} when the model is trained on ScanNet~\cite{dai2017scannet} and evaluated on ScanNet++~\cite{yeshwanth2023scannet++}. Since \textit{ceiling} parts are largely absent in the multi-view images of ScanNet but frequently appear in ScanNet++, model trained on ScanNet tend to generate noisy depth maps for these regions during evaluation on ScanNet++. Inaccurate depth maps lead to noisy point clouds, which in turn degrade the quality of feature aggregation performed by the 3D U-Net and the memory-based adapter of \ours. If the agent is equipped with depth sensors, this issue can be largely mitigated. 

\section{Additional Experiments.}

Understanding the 3D scene with direct 3D referring is crucial in embodied scenarios for the faster inference and better spatial comprehension. Hence, main paper focuses on 3D Semantic Segmentation by annotating the point clouds without rendering the 2D feature maps. However, as we show in Fig.~\textcolor{cvprblue}{1}, our \ours supports diverse perception tasks such as 2D-rendered segmentation and novel-view synthesis. In this section, we present more various experiments that are not included in the main paper: Sec.~\ref{sec:supple_exp_3dss} presents comparisons against a broader set of baselines and further evaluates \ours with diverse 2D models on 3D semantic segmentation. Sec.~\ref{sec:supple_exp_2dss} explores the comparisons on 2D-rendered segmentation. Sec.~\ref{sec:supple_exp_nvs} conducts the experiments on novel-view synthesis in RGB space. Sec.~\ref{sec:supple_exp_memory} provides deeper ablations on memory efficiency of our sparse coefficient field. Finally, Sec.~\ref{sec:supple_exp_quali} provides more qualitative results of our \ours and \oursfast.

\subsection{3D Semantic Segmentation}
\label{sec:supple_exp_3dss}

Here, we provide additional comparisons on 3D semantic segmentation with broader set of baselines and explores diverse 2D models within the \ours framework. Experimental setting is kept identical with the main paper and experiment is conducted on ScanNet~\citep{dai2017scannet} and ScanNet200~\cite{rozenberszki2022language} datasets with varying number of classes: 10, 15, 19 and 70 classes.

\vspace{1mm}
\noindent \textbf{Comparisons with 3DGS methods.} 1st-7th rows of Tab.~\ref{tab:supple_exp_3dss} presents the additional comparisons with semantic 3DGS which support direct 3D referring. Specifically, LUDVIG~\cite{marrie2025ludvig} and CF$^3$~\cite{lee2025cf3} are further added as a recent baselines. The 5th-7th rows of Tab.~\ref{tab:supple_exp_3dss} further demonstrate that \ours performs well across diverse 2D models, indicating that its compatibility is not restricted to any specific 2D VLM. It is worth noting that widely used SAM+CLIP combination is not suitable for embodied scenarios that require fast inference, as discussed in Sec.~\ref{sec:supple_discussions_2dvlm}. Hence, we adopt FastSam~\cite{zhao2023fast} with pixel-level CLIP model~\cite{radford2021learning} to extract semantic cues from 2D images.

\vspace{1mm}
\noindent \textbf{Comparisons with point-cloud methods.} We further compare \ours with the point-cloud understanding  methods in 8th-10th rows of Tab.~\ref{tab:supple_exp_3dss}. Specifically, we adopt OpenScene~\citep{peng2023openscene} as offline method and EmbodiedSAM~\citep{xu2024embodiedsam} as online method. Since they take RGB-D as inputs, we feed same RGB-D into \ours for the fair evaluation. Our \ours outperforms both methods by exploiting the 3DGS representation which enables smooth feature aggregation onto 3D points using the Mahalanobis distance defined in Eq.~\ref{eq:supple_mahalanobis}.

\subsection{2D-rendered Semantic Segmentation}
\label{sec:supple_exp_2dss}

Here, we explore the 2D-rendered semantic segmentation with our \ours.

\begin{table}[t]
    \renewcommand{\arraystretch}{1.5}
    \centering
    \footnotesize
    \resizebox{\linewidth}{!}{
    \begin{tabular}{l|c|cc|cc|cc}
    \toprule
    \multirow{3}{*}{\textbf{Method}} & \multirow{3}{*}{\textbf{Search Domain}} & \multicolumn{6}{c}{\textbf{ScanNet}~\cite{dai2017scannet}}  \\
    \cline{3-8}
     & & \multicolumn{2}{c|}{10 classes} & \multicolumn{2}{c|}{15 classes} & \multicolumn{2}{c}{19 classes}   \\
    & & mIoU & mACC & mIoU & mACC & mIoU & mACC \\
     \midrule \noalign{\vskip -2pt}
     LangSplat~\cite{qin2024langsplat} & 2D & 45.83 & 73.12 & 42.89 & 69.34 & 44.15 & 70.45   \\
     \midrule
     Occam's LGS~\cite{jun2025dr} & \multirow{2}{*}{3D} & 41.13 & 73.34 & 40.21 & 66.82 & 39.11 & 64.34  \\
     Dr. Splat~\cite{jun2025dr} &  & 40.11 & 71.62 & 38.45 & 66.11 & 39.67 & 65.72   \\
     \midrule
     \cellcolor{green_background} EmbodiedSplat & 3D & 47.44 & 76.95 & 44.11 & 70.12 & 43.75 & 68.16  \\
    \bottomrule
    \end{tabular}}
    \newline
    \caption{\textbf{Quantitative results on 2D-rendered semantic segmentation in ScanNet~\cite{dai2017scannet} dataset.}}
    \label{tab:supple_exp_2dss}
    \vspace{-10pt}
\end{table}

\vspace{1mm}
\noindent \textbf{Rendering 2D Feature Map with \ours.} Rendering the 2D feature maps from 3D Gaussians with high dimensional features incurs huge computational overhead. Hence, we adopt alternative approach which does not require direct rendering of Gaussian features. For pixel $j$ of the target view, we obtain the list of the rendering weights for every Gaussians. Specifically, rendering weight can be easily obtained by leveraging rasterization function of Eq.~\textcolor{cvprblue}{1} where $T_i\tilde{\alpha}_i$ denotes the rendering weight of $i$-th Gaussian. We keep top 5 Gaussians with the highest weight and compute the cosine similarities for each selected Gaussian with given text classes. Finally, cosine similarities of every 5 Gaussians are linearly combined using their corresponding rendering weights, finally outputting the cost map for pixel $j$. Note that rendering weights are renormalized to sum to 1 before performing the linear combination.

We compute two cost maps by using 2D CLIP features and 3D CLIP features, and ensemble them via Eq.~\textcolor{cvprblue}{6} to obtain final costmap in 2D domain.

\vspace{1mm}
\noindent \textbf{Experimental Settings.} We evaluate the 2D-rendered semantic segmentation on interpolated novel views of ScanNet~\citep{dai2017scannet} dataset with 10, 15 and 19 classes. Testing scenes are identical with 3D semantic segmentation setting.

\vspace{1mm}
\noindent \textbf{Experimental Results.} Tab.~\ref{tab:supple_exp_2dss} exhibits the performance on 2D-rendered segmentation performance on ScanNet dataset. Our \ours shows the comparable performance with 2D-specific method such as LangSplat~\citep{qin2024langsplat} even though it is not optimized to specific scene.

\begin{table}[t]
    \renewcommand{\arraystretch}{1.5}
    \centering
    \footnotesize
    \resizebox{\linewidth}{!}{
    \begin{tabular}{l|ccccc|c}
    \toprule
    \textbf{Method} & iPSNR$\uparrow$ & ePSNR$\uparrow$ & SSIM$\uparrow$ & LPIPS$\downarrow$ & $\delta < 1.1\uparrow$ & Type \\
    \midrule
    pixelSplat~\cite{charatan2024pixelsplat} & 15.54 & 13.47 & 0.557 & 0.608 & 0.023 & \textcolor{blue}{\textbf{Offline}} \\
    MVSplat~\cite{chen2024mvsplat} & 16.51 & 13.67 & 0.591 & 0.541 & 0.323 & \textcolor{blue}{\textbf{Offline}} \\
    PixelGaussian~\cite{fei2024pixelgaussian} & 16.33 & 13.40 & 0.601 & 0.549 & 0.282 & \textcolor{blue}{\textbf{Offline}} \\
    FreeSplat++~\cite{wang2025freesplat++} & 23.29 & 19.44 & 0.771 & 0.320 & 0.904 & \textcolor{blue}{\textbf{Offline}} \\
    \midrule
    \cellcolor{green_background} \ours & 22.78 & 19.14 & 0.738 & 0.367 & 0.885 & \textcolor{red}{\textbf{Online}} \\
    \bottomrule
    \end{tabular}}
    \newline
    \caption{\textbf{Whole Scene Reconstruction resutls on ScanNet~\citep{dai2017scannet}.} iPSNR and ePSNR detnoes PSNR on the interpolated views and extrapolated views, respectively.}
    \label{tab:supple_exp_nvs_scannet}
    \vspace{-17pt}
\end{table}

\begin{table}[t]
    \renewcommand{\arraystretch}{1.5}
    \centering
    \footnotesize
    \resizebox{\linewidth}{!}{
    \begin{tabular}{l|ccccc|c}
    \toprule
    \textbf{Method} & iPSNR$\uparrow$ & ePSNR$\uparrow$ & SSIM$\uparrow$ & LPIPS$\downarrow$ & $\delta < 1.1\uparrow$ & Type \\
    \midrule
    pixelSplat~\cite{charatan2024pixelsplat} & 10.70 & 10.37 & 0.497 & 0.663 & 0.000 & \textcolor{blue}{\textbf{Offline}} \\
    MVSplat~\cite{chen2024mvsplat} & 11.10 & 10.62 & 0.497 & 0.648 & 0.028 & \textcolor{blue}{\textbf{Offline}} \\
    PixelGaussian~\cite{fei2024pixelgaussian} & 10.78 & 10.44 & 0.529 & 0.639 & 0.012 & \textcolor{blue}{\textbf{Offline}} \\
    FreeSplat++~\cite{wang2025freesplat++} & 22.63 & 19.51 & 0.829 & 0.261 & 0.890 & \textcolor{blue}{\textbf{Offline}} \\
    \midrule
    \cellcolor{green_background} \ours & 21.54 & 18.62 & 0.780 & 0.330 & 0.925 & \textcolor{red}{\textbf{Online}} \\
    \bottomrule
    \end{tabular}}
    \newline
    \caption{\textbf{Whole Scene Reconstruction resutls on ScanNet++~\citep{yeshwanth2023scannet++}.} iPSNR and ePSNR detnoes PSNR on the interpolated views and extrapolated views, respectively.}
    \label{tab:supple_exp_nvs_scannetpp}
    \vspace{-10pt}
\end{table}

\subsection{Novel View Synthesis}
\label{sec:supple_exp_nvs}

In this section, we evaluate the rendering quality of \ours in the RGB space.

\vspace{1mm}
\noindent \textbf{Experimental Settings.} Given the constructed whole-scene 3DGS, we render the interpolated and extrapolated novel views respectively to evaluate the novel view synthesis, following~\citep{wang2025freesplat++}. PSNR, SSIM~\cite{wang2004image} and LPIPS~\cite{zhang2018unreasonable} are adopted as rendering metric. We further evaluate geometric accuracy by reporting the depth quality. Specifically, threshold tolerance $\delta < 1.1$ on depth difference between rendered depth and ground-truth depth is adopted as metric. We conduct the experiments on ScanNet~\cite{dai2017scannet} and ScanNet++~\cite{yeshwanth2023scannet++} datasets.

\vspace{1mm}
\noindent \textbf{Baselines.} We compare the rendering quality of our \ours with representative feed-forward 3DGS works: pixelSplat~\cite{charatan2024pixelsplat}, MVSplat~\cite{chen2024mvsplat}, PixelGaussian~\cite{fei2024pixelgaussian} and FreeSplat++~\cite{wang2025freesplat++}.

\vspace{1mm}
\noindent \textbf{Experimental Results.} Tab.~\ref{tab:supple_exp_nvs_scannet} and Tab.~\ref{tab:supple_exp_nvs_scannetpp} show that our \ours successfully adapt the inference pipeline of FreeSplat++ into online setting, where it shows the comparable performance to original FreeSplat++ which leverages the entire set of pre-collected images in offline setting. Hence, it inherits the superior performance of FreeSplat++ compared to the previous feed-forward 3DGS models~\citep{fei2024pixelgaussian, charatan2024pixelsplat, chen2024mvsplat} in whole-scene reconstruction setting.

\subsection{Ablations on Memory Compression Rate}
\label{sec:supple_exp_memory}

In this section, we further explore the memory efficiency of our proposed sparse coefficient field with CLIP global codebook.

\begin{table}[t]
    \renewcommand{\arraystretch}{1.5}
    \centering
    \footnotesize
    \resizebox{\linewidth}{!}{
    \begin{tabular}{l|cc|cc}
    \toprule
    \textbf{Scene} & Gaussians Num & Codebook Size & Total Size (MB) & Compression Ratio \\
    \midrule
    \texttt{scene0000\_01} & 3.2M & 8.7K & 148 & \textcolor{green}{\textbf{$\times$ 63 efficient}} \\
    \texttt{scene0046\_00} & 2.4M & 5.3K & 106 & \textcolor{green}{\textbf{$\times$ 65 efficient}} \\
    \texttt{scene0079\_00} & 1.5M & 2.8K & 64 & \textcolor{green}{\textbf{$\times$ 67 efficient}} \\
    \texttt{scene0158\_00} & 1.1M & 1.8K & 48 & \textcolor{green}{\textbf{$\times$ 68 efficient}} \\
    \texttt{scene0316\_00} & 0.6M & 0.6K & 23 & \textcolor{green}{\textbf{$\times$ 70 efficient}} \\
    \texttt{scene0389\_00} & 1.9M & 2.8K & 82 & \textcolor{green}{\textbf{$\times$ 69 efficient}} \\
    \texttt{scene0406\_00} & 0.9M & 2.1K & 41 & \textcolor{green}{\textbf{$\times$ 65 efficient}} \\
    \texttt{scene0521\_00} &1.1M & 2.0K & 49 & \textcolor{green}{\textbf{$\times$ 67 efficient}} \\
    \texttt{scene0553\_00} & 0.7M & 0.8K & 31 & \textcolor{green}{\textbf{$\times$ 70 efficient}} \\
    \texttt{scene0616\_00} & 2.3M & 3.4K & 98 & \textcolor{green}{\textbf{$\times$ 68 efficient}} \\
    \midrule
    Average & 1.57M & 3.0K & 69 & \textcolor{green}{\textbf{$\times$ 67 efficient}} \\
    \bottomrule
    \end{tabular}}
    \newline
    \caption{\textbf{Ablations on memory efficiency of sparse coefficient field in ScanNet~\cite{dai2017scannet} dataset.}}
    \label{tab:supple_exp_memory}
    \vspace{-10pt}
\end{table}

\vspace{1mm}
\noindent \textbf{Experimental Settings.} We report the number of generated Gaussians and the number of CLIP features stored in the CLIP global codebook for each testing scene. Based on this, we estimate the total memory size of semantic Gaussians stored by sparse coefficient field. Specifically, we combine the size of the CLIP codebook with the sizes of the index and weight caches attached to each Gaussian. Finally, we report the memory compression ratio gained from our sparse coefficient field compared to naively storing every per-gaussian original CLIP features with 768 dimension.

\begin{figure*}[t]
\centering
\includegraphics[width=0.8\linewidth]{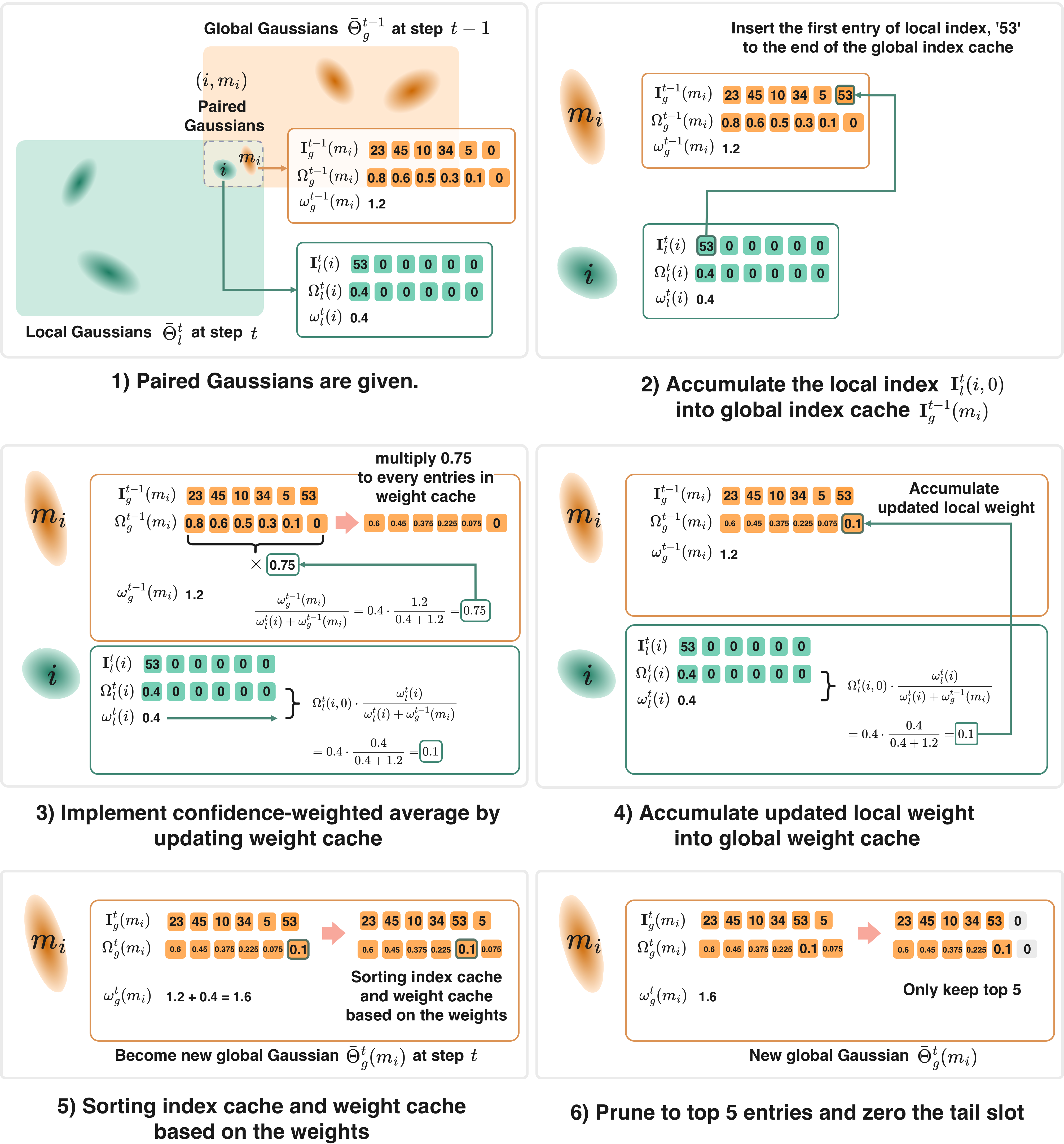}
  \caption{\textbf{Toy example of online fusion algorithm with sparse coefficient field.} \textbf{1)} Sparse coefficient field of paired Gaussians $(i, m_i)$ are fused by our online fusion Algorithm.~\textcolor{cvprblue}{1}. The local Gaussians (\textcolor{green}{\textbf{\textit{Green}}}) which do not have valid match with global Gaussians are just appended to the global set without update. \textbf{2) Line 1 of Algorithm.~\textcolor{cvprblue}{1}:} First entry of local index, $\mathbf{I}^t_i(i, 0)$ is inserted to the last entry of global index cache $\mathbf{I}^{t-1}_g(m_i, -1)$. In the above example, index 53 is appended to $\mathbf{I}^{t-1}_g(m_i)$, such that $\mathbf{I}^{t-1}_g(m_i, -1) \leftarrow 53$. \textbf{3-4) Lines 3-4 of Algorithm.~\textcolor{cvprblue}{1}:} Both the local weight cache $\Omega^t_l(i)$ and global weight cache $\Omega^{t-1}_g(m_i)$ are updated based on the confidence-weighted average. Specifically, all of the entries in $\Omega^{t-1}_g(m_i)$ are multiplied by 0.75, while the first entry of local weight cache $\Omega^t_l(i, 0)$ is scaled by 0.25. Scaled local weight value $0.25 \cdot \Omega^t_l(i, 0)$ is then inserted to the last entry of global weight cache, such that: $\Omega^{t-1}_g(m_i, -1) \leftarrow 0.25 \cdot \Omega^t_l(i, 0)$. Since both the index cache and weight cache of local Gaussian $\bar{\Theta}^{t}_l(i)$ are incorporated into global gaussian $\bar{\Theta}^{t-1}_g(m_i)$ from the previous stages, $\bar{\Theta}^{t-1}_g(m_i)$ becomes new $m_i$-th global Gaussian at step $t$: $\bar{\Theta}^t_g(m_i) \leftarrow \bar{\Theta}^{t-1}_g(m_i)$. $i$-th local Gaussian is simply discarded. \textbf{5-6) Lines 5-6 of Algorithm.~\textcolor{cvprblue}{1}:} We sort both the global weight cache and index cache based on the weight values $\Omega^t_g(m_i)$ in descending order. Then, we keep first $L-1=5$ entires and discard the last values by overwriting them with zero: $\Omega^t_g(m_i, -1) \leftarrow 0, \; \mathbf{I}^t_g(m_i, -1) \leftarrow 0$. This keeps each cache size fixed at $L$ along the exploration, effectively improving the memory efficiency.}
  \label{fig:supple_online_fuse_toy-5}
  \vspace{-5pt}
\end{figure*}

\vspace{1mm}
\noindent \textbf{Observations.} Tab.~\ref{tab:supple_exp_memory} shows that the number of CLIP features stored in codebook is far smaller than the total number of Gaussians, yielding an average 67× improvement in memory efficiency compared to storing per-Gaussian original CLIP features. Our sparse coefficient field is highly practical since it doesn't require any pretraining stage or per-scene optimization compared to the existing memory compression method such as Auto-encoder~\cite{qin2024langsplat}, PQ index~\cite{jun2025dr} and per-scene optimzied codebook~\cite{wu2024opengaussian, li2025instancegaussian}. Instead, the sparse coefficient field is constructed on the fly alongside the semantic 3DGS reconstruction and supports real-time online updates through Algorithm.~\textcolor{cvprblue}{1} (\cf Fig.~\ref{fig:supple_online_fuse_toy-5}), making it highly suitable for online settings.

\subsection{Qualitative Results}
\label{sec:supple_exp_quali}

In this section, we explain additional qualitative results of our \ours and \oursfast.

\begin{figure*}[t]
\captionsetup[subfigure]{}
  \centering  
  \begin{subfigure}{0.22\linewidth}\includegraphics[width=1.0\linewidth]{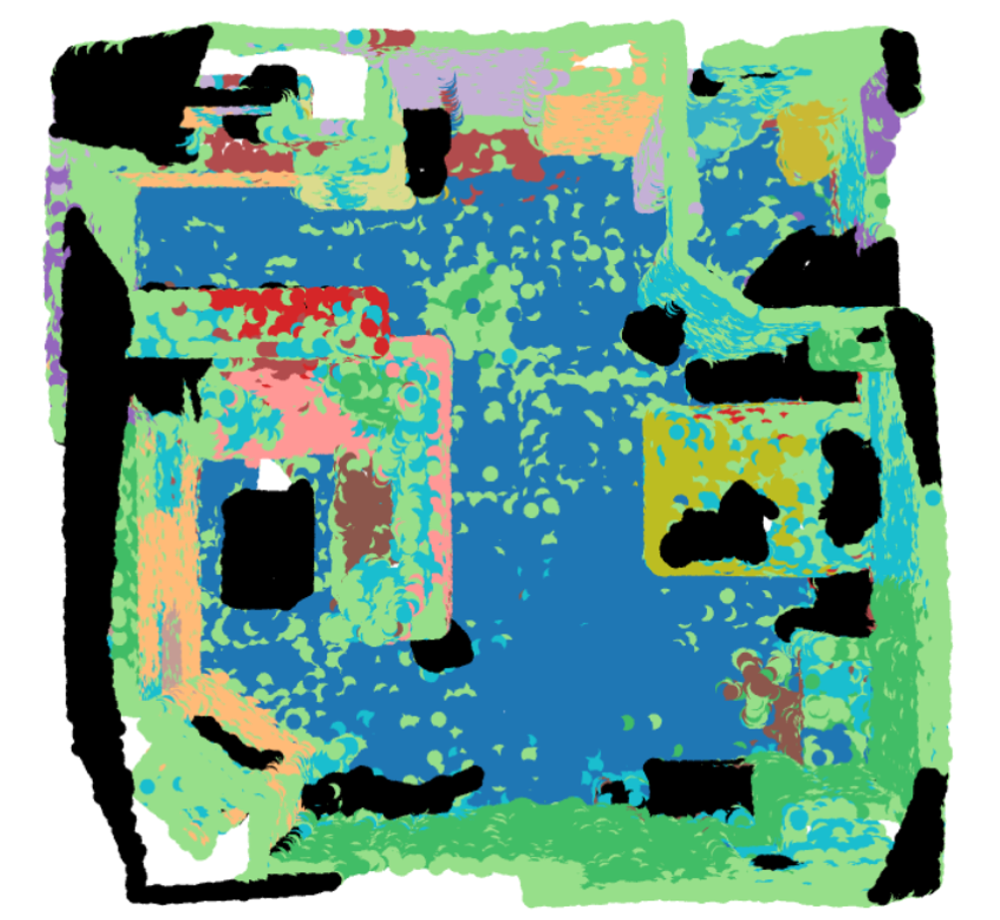}
  \subcaption*{OpenGaussian~\cite{wu2024opengaussian}}
  \end{subfigure}
  \begin{subfigure}{0.22\linewidth}\includegraphics[width=1.0\linewidth]{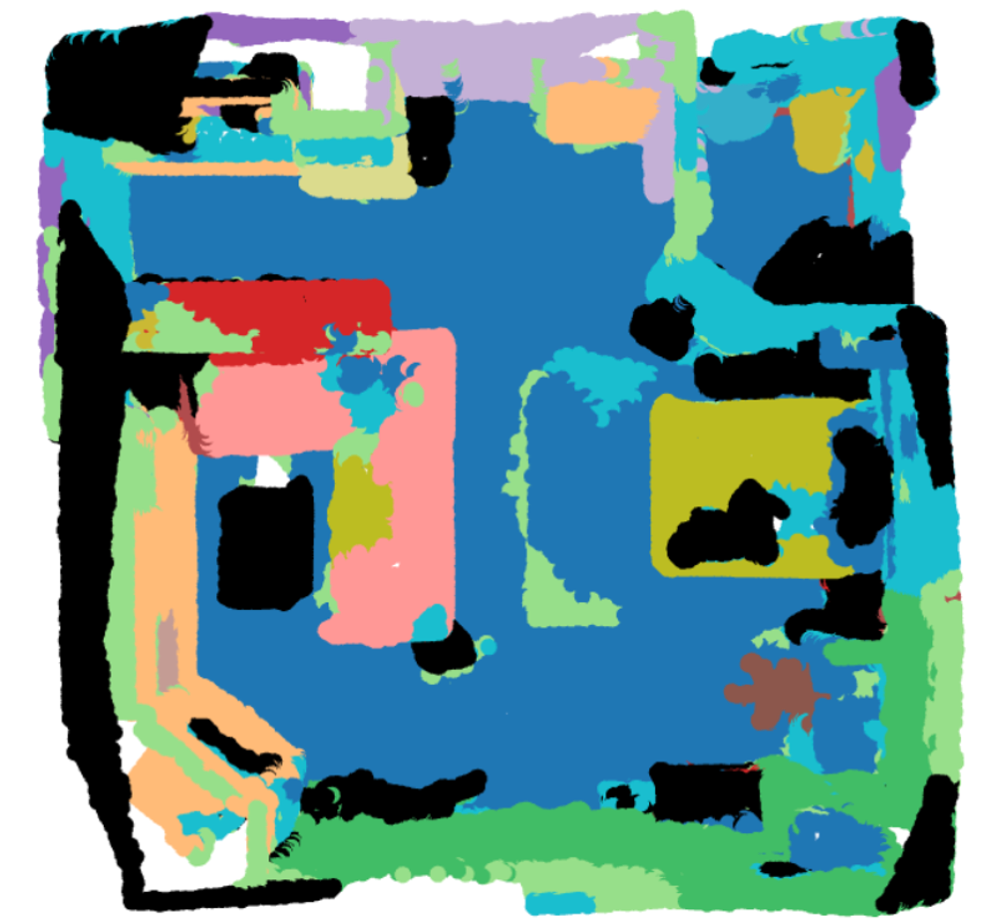}
  \subcaption*{InstanceGaussian~\cite{li2025instancegaussian}}
  \end{subfigure}
  \begin{subfigure}{0.22\linewidth}\includegraphics[width=1.0\linewidth]{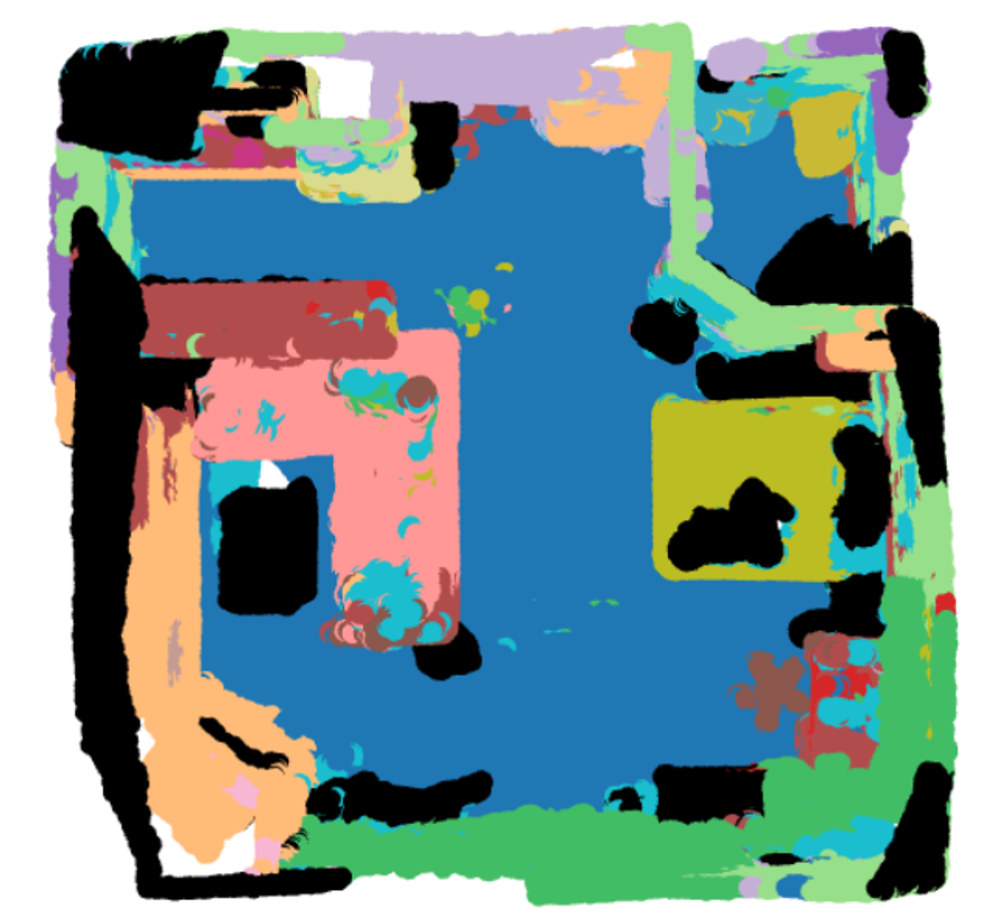}
  \subcaption*{Dr. Splat~\cite{jun2025dr}}
  \end{subfigure}
  \begin{subfigure}{0.22\linewidth}\includegraphics[width=1.0\linewidth]{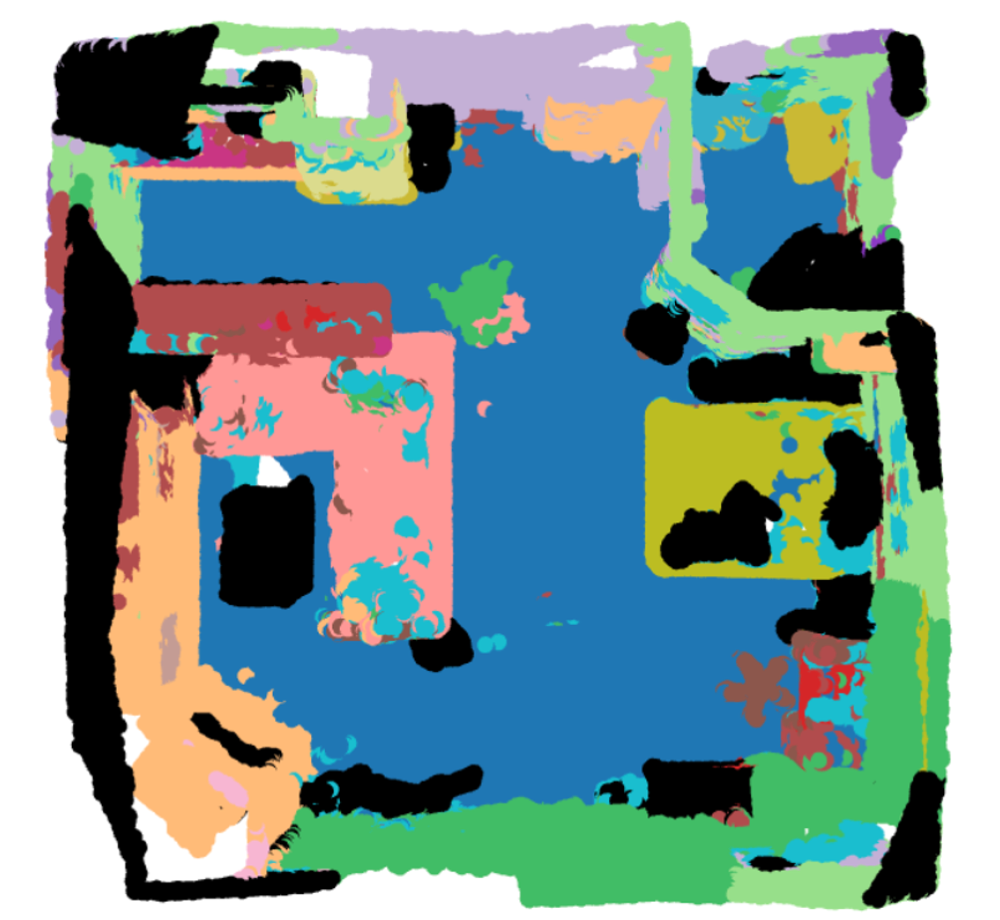}
  \subcaption*{Occam's LGS~\cite{cheng2024occam}}
  \end{subfigure}
  \begin{subfigure}{0.31\linewidth}\includegraphics[width=1.0\linewidth]{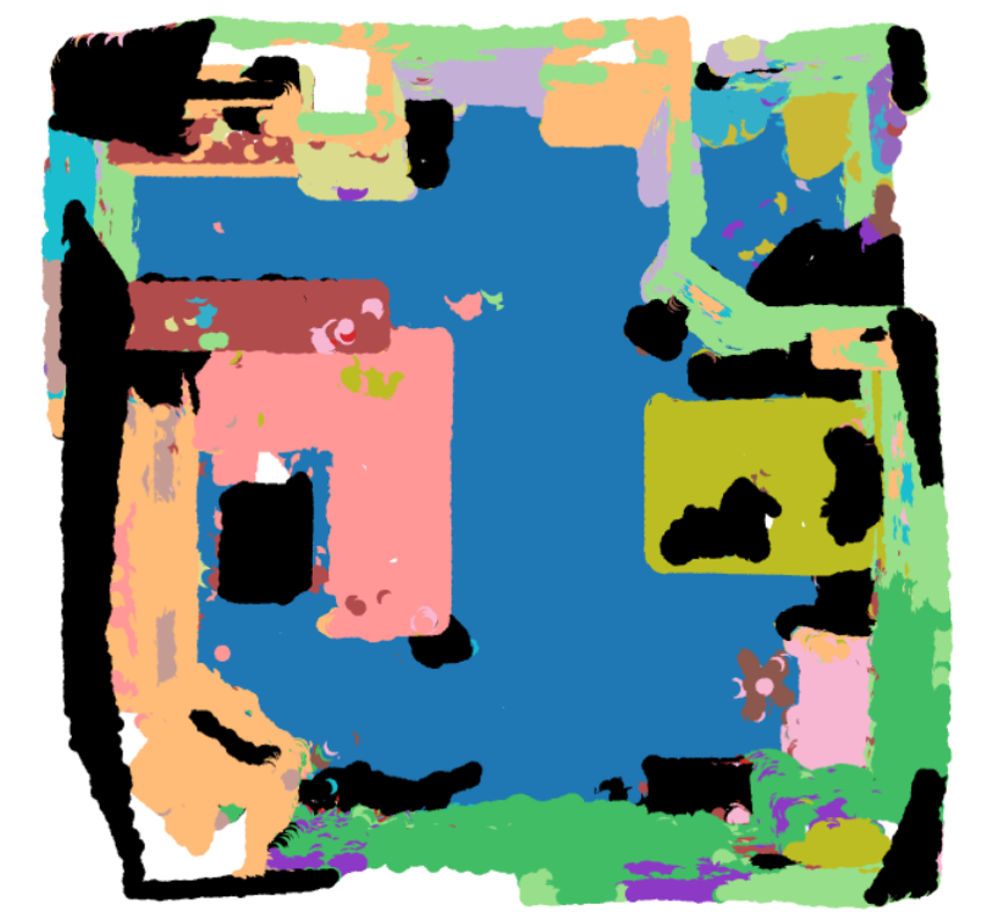}
  \subcaption*{\oursfast}
  \end{subfigure}
  \begin{subfigure}{0.31\linewidth}\includegraphics[width=1.0\linewidth]{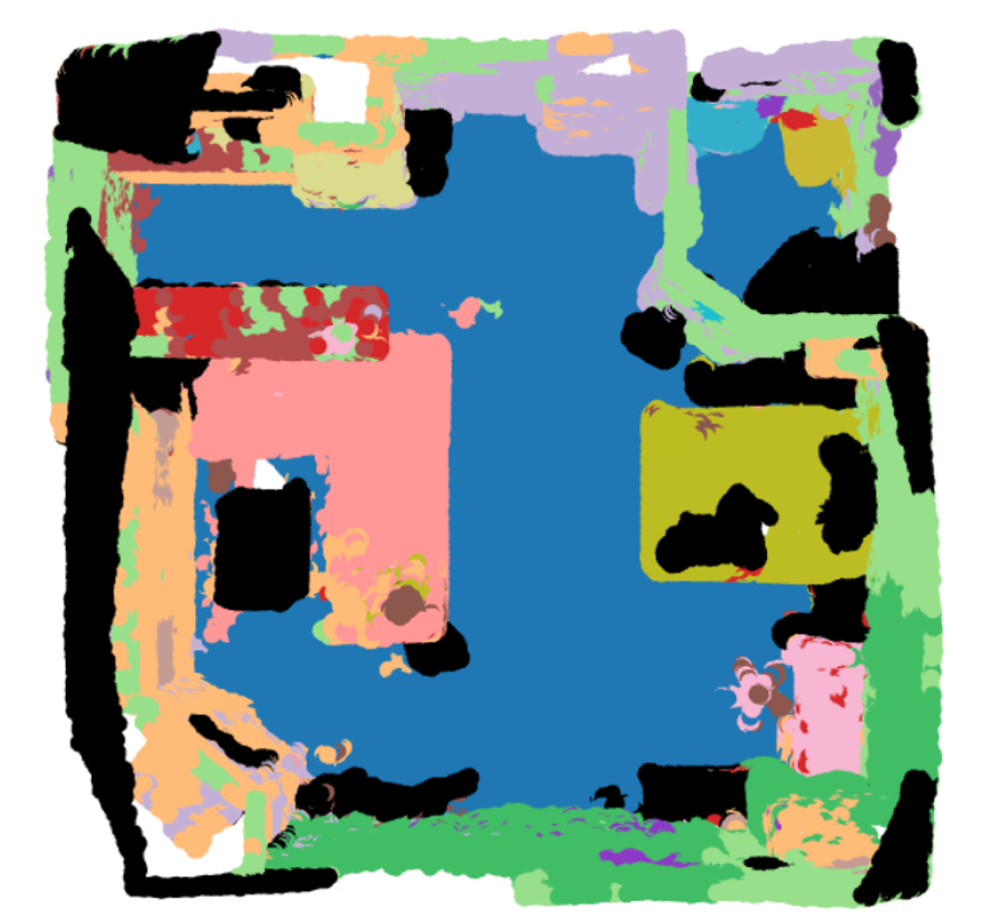}
  \subcaption*{\ours}
  \end{subfigure}
  \begin{subfigure}{0.31\linewidth}\includegraphics[width=1.0\linewidth]{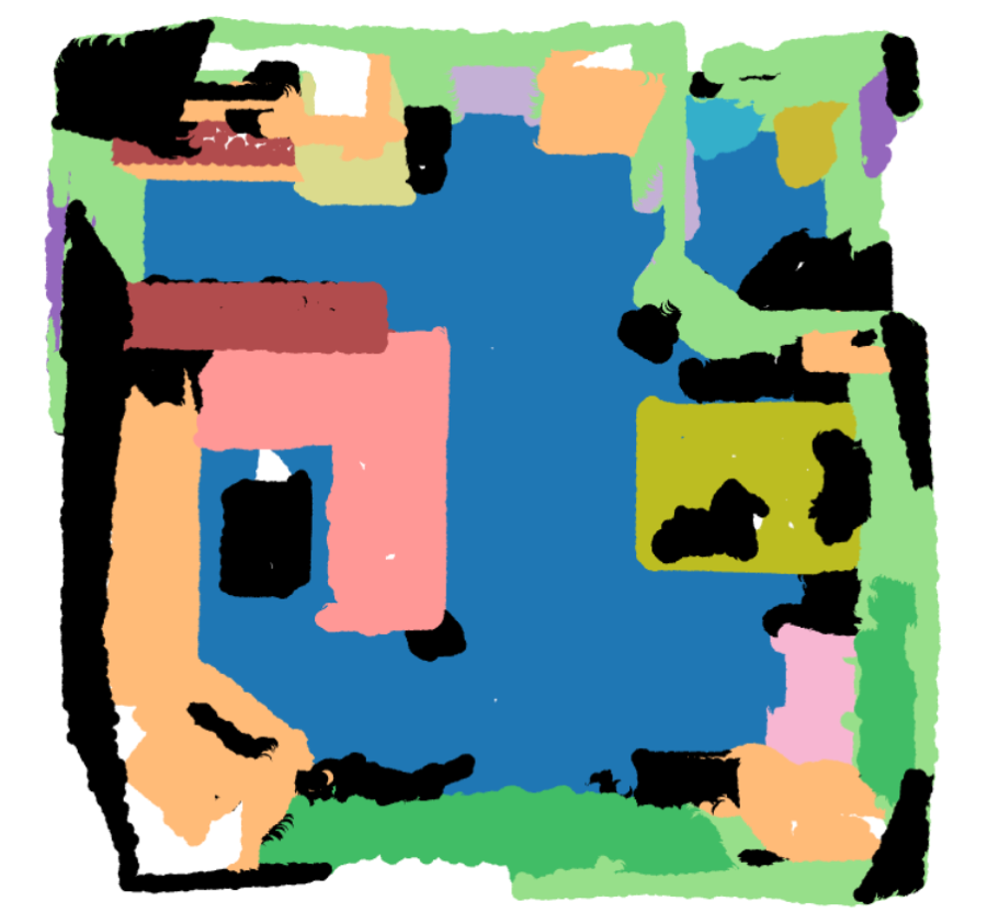}
  \subcaption*{GT}
  \end{subfigure}
  \vspace{30pt}

  \begin{subfigure}{0.22\linewidth}\includegraphics[width=1.0\linewidth]{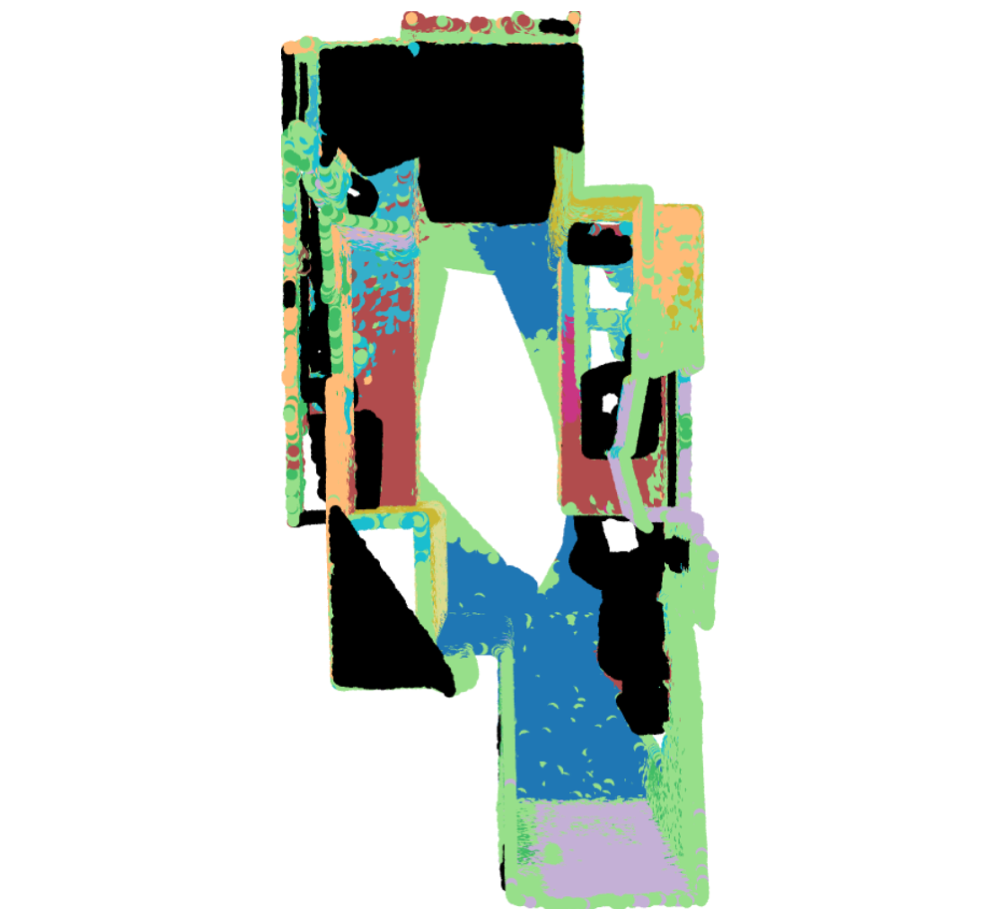}
  \subcaption*{OpenGaussian~\cite{wu2024opengaussian}}
  \end{subfigure}
  \begin{subfigure}{0.22\linewidth}\includegraphics[width=1.0\linewidth]{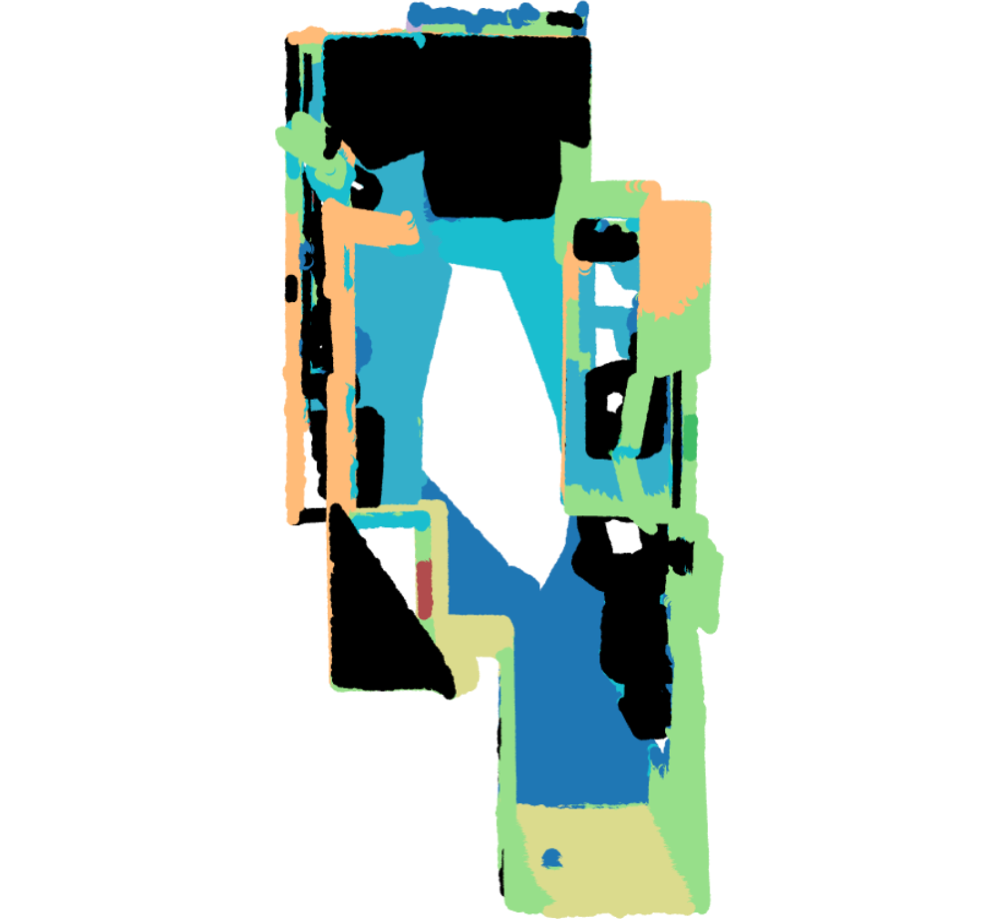}
  \subcaption*{InstanceGaussian~\cite{li2025instancegaussian}}
  \end{subfigure}
  \begin{subfigure}{0.22\linewidth}\includegraphics[width=1.0\linewidth]{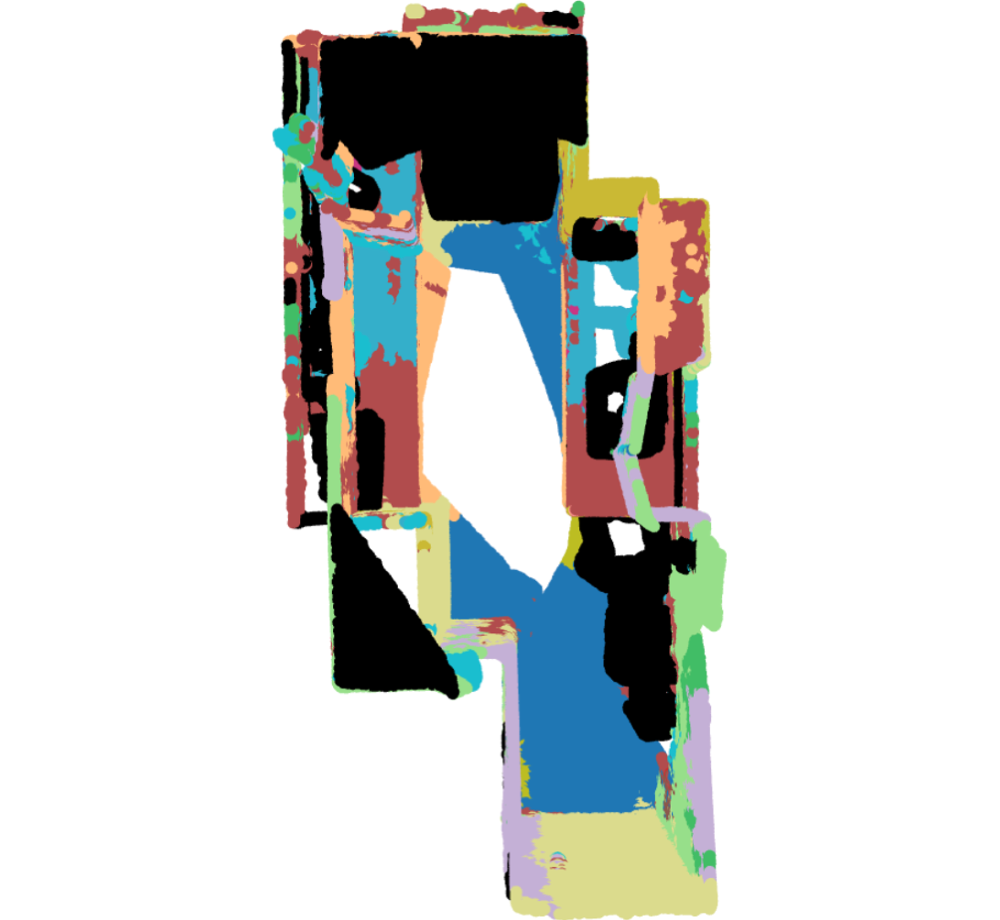}
  \subcaption*{Dr. Splat~\cite{jun2025dr}}
  \end{subfigure}
  \begin{subfigure}{0.22\linewidth}\includegraphics[width=1.0\linewidth]{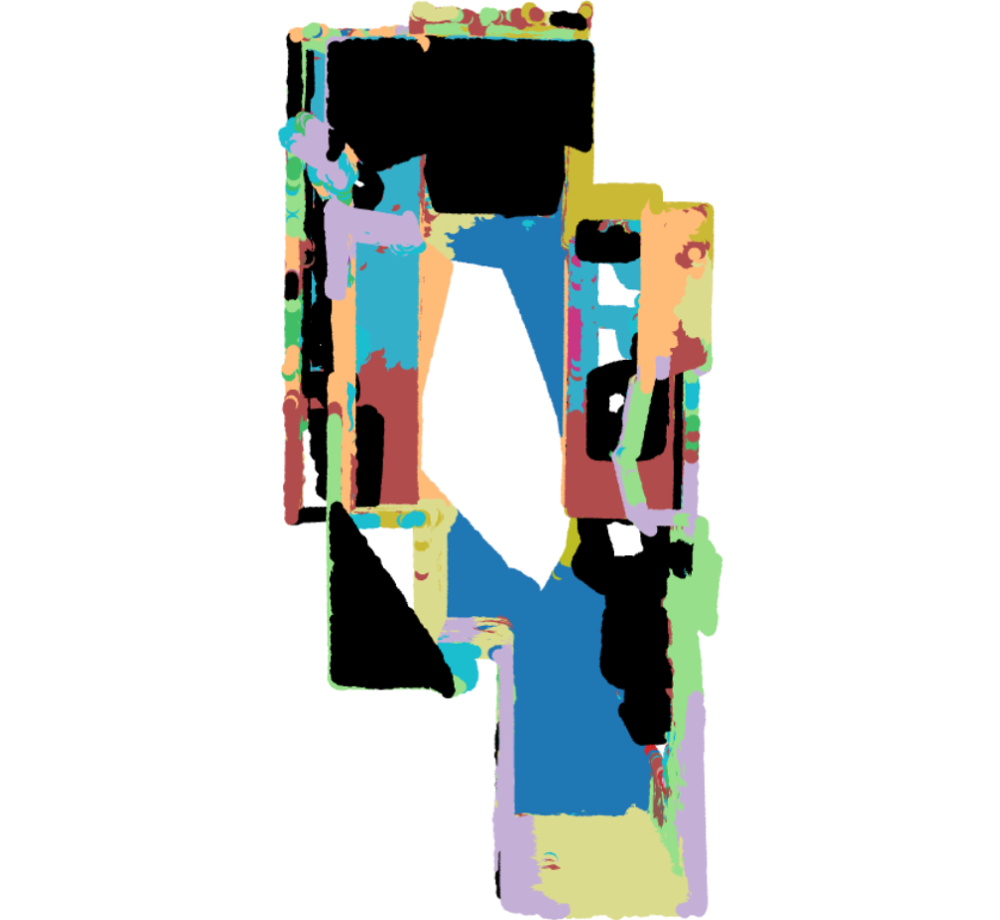}
  \subcaption*{Occam's LGS~\cite{cheng2024occam}}
  \end{subfigure}
  \begin{subfigure}{0.31\linewidth}\includegraphics[width=1.0\linewidth]{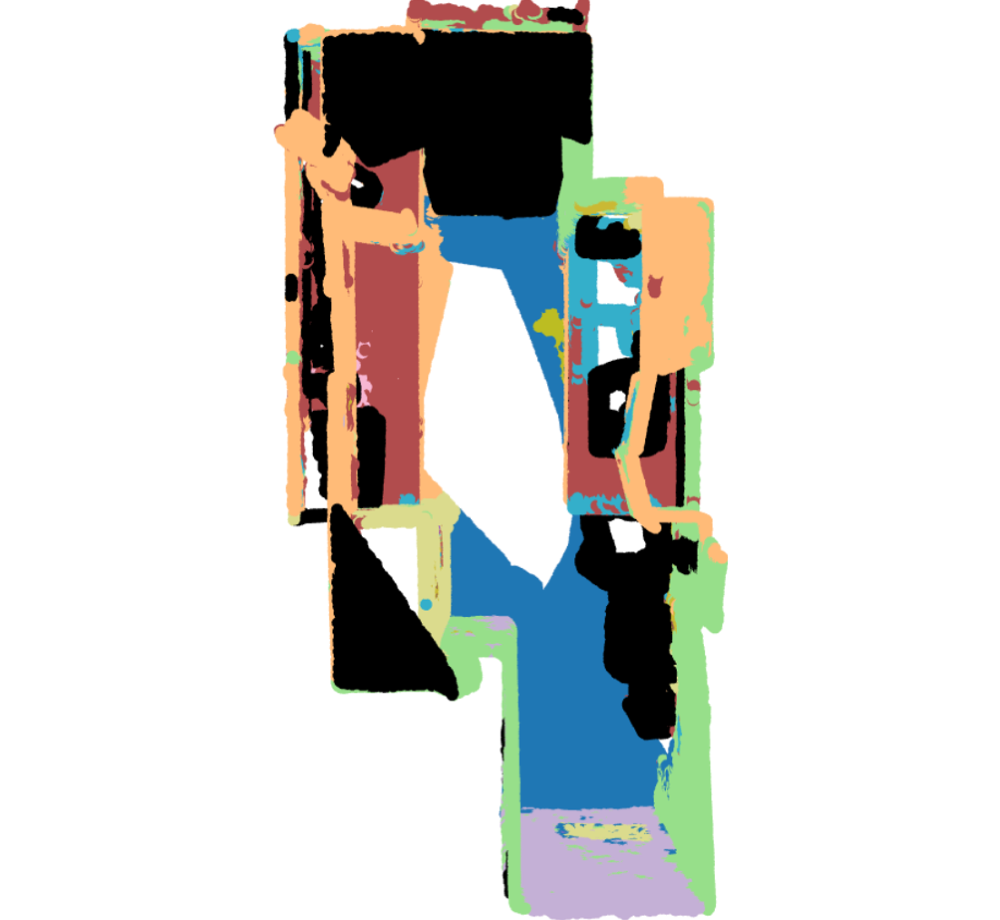}
  \subcaption*{\oursfast}
  \end{subfigure}
  \begin{subfigure}{0.31\linewidth}\includegraphics[width=1.0\linewidth]{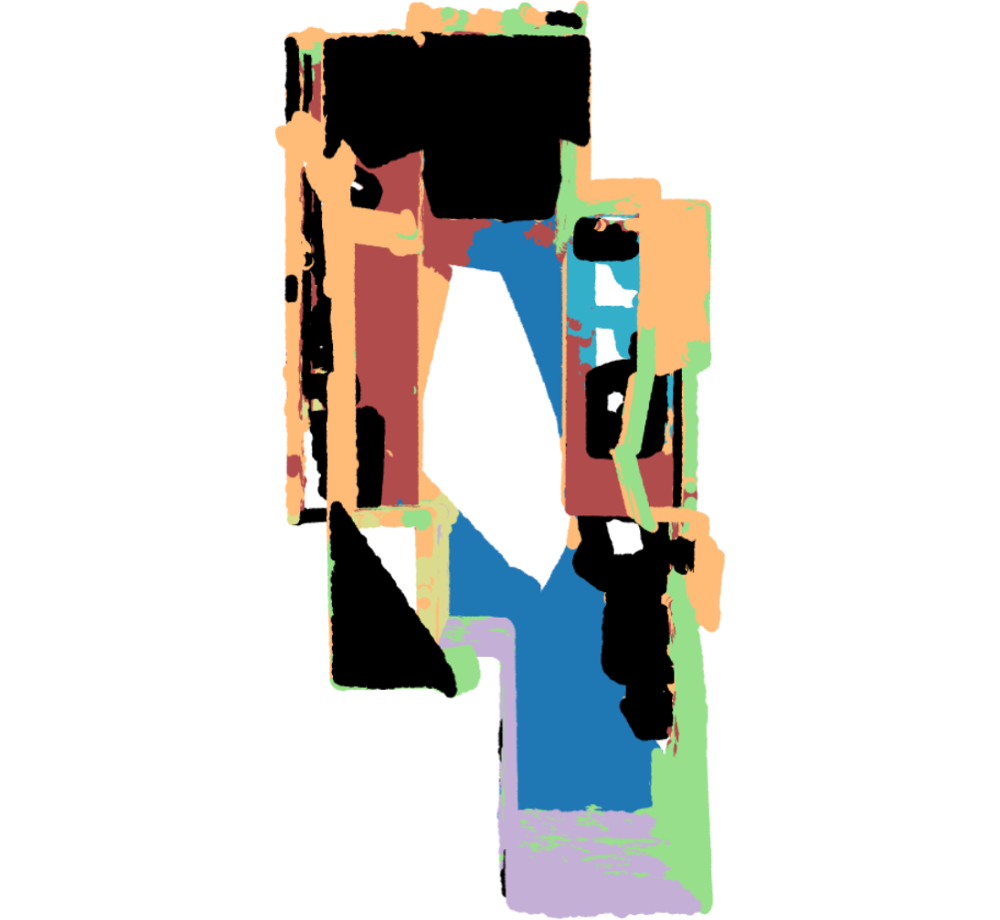}
  \subcaption*{\ours}
  \end{subfigure}
  \begin{subfigure}{0.31\linewidth}\includegraphics[width=1.0\linewidth]{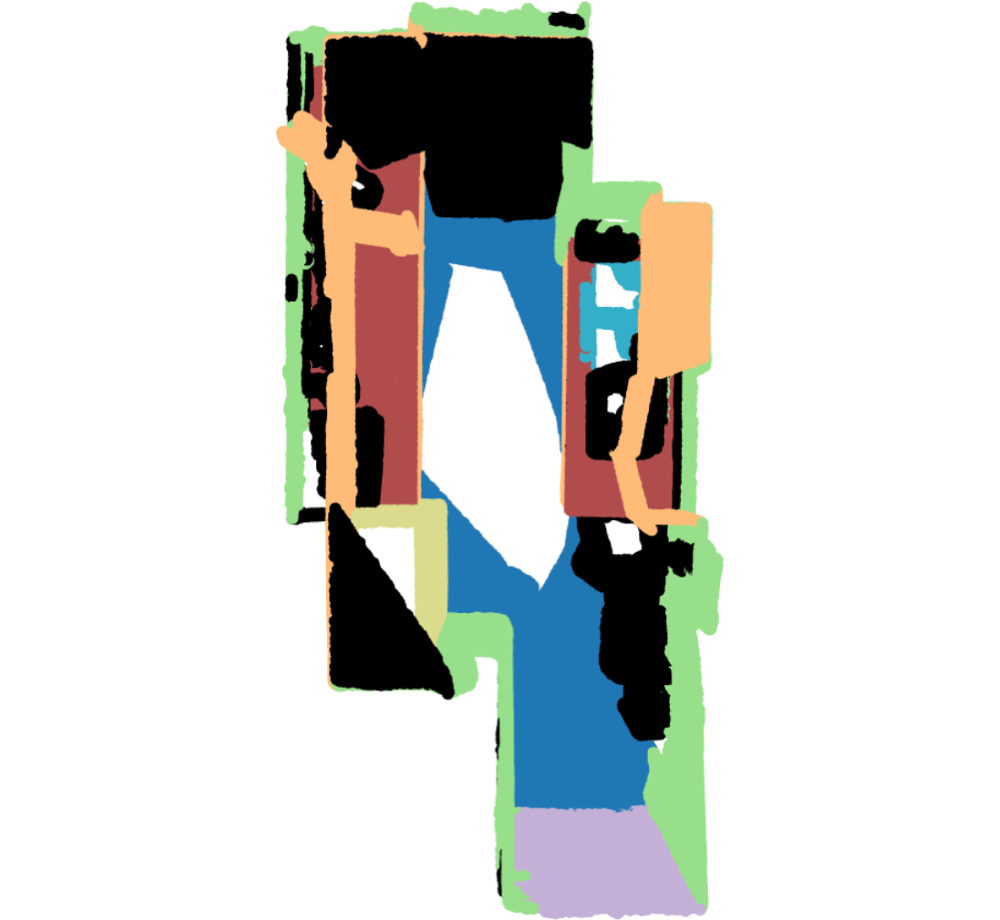}
  \subcaption*{GT}
  \end{subfigure}

  \caption{\textbf{More qualitative comparisons on 3D semantic segmentaiton - (1)}}
  \label{fig:supple_qualitative_3d_1}
\end{figure*}

\begin{figure*}[t]
\captionsetup[subfigure]{}
  \centering  
  \begin{subfigure}{0.22\linewidth}\includegraphics[width=1.0\linewidth]{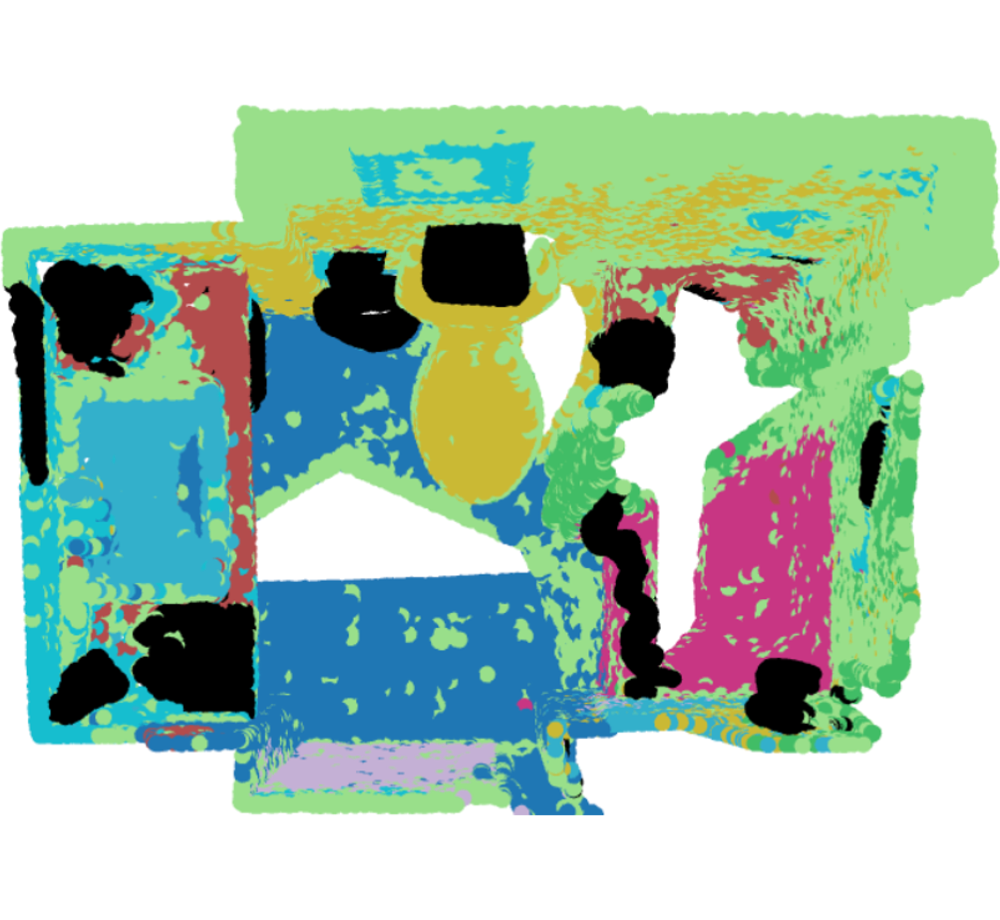}
  \subcaption*{OpenGaussian~\cite{wu2024opengaussian}}
  \end{subfigure}
  \begin{subfigure}{0.22\linewidth}\includegraphics[width=1.0\linewidth]{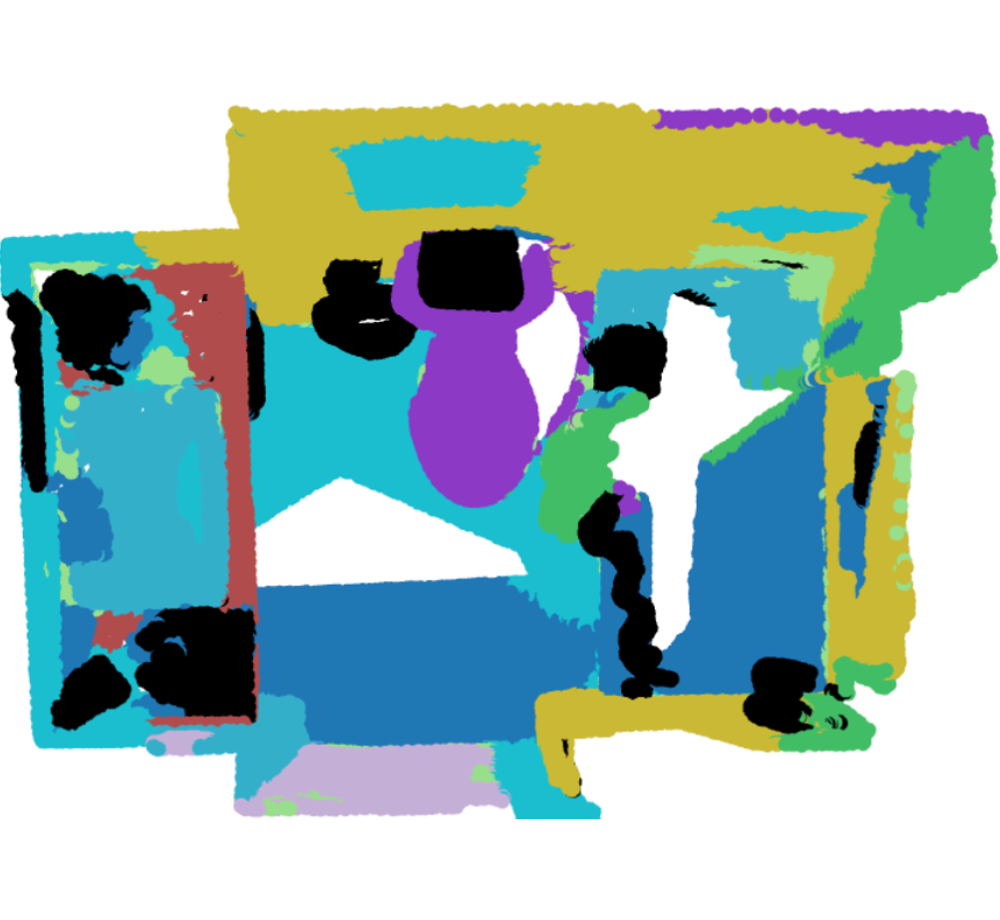}
  \subcaption*{InstanceGaussian~\cite{li2025instancegaussian}}
  \end{subfigure}
  \begin{subfigure}{0.22\linewidth}\includegraphics[width=1.0\linewidth]{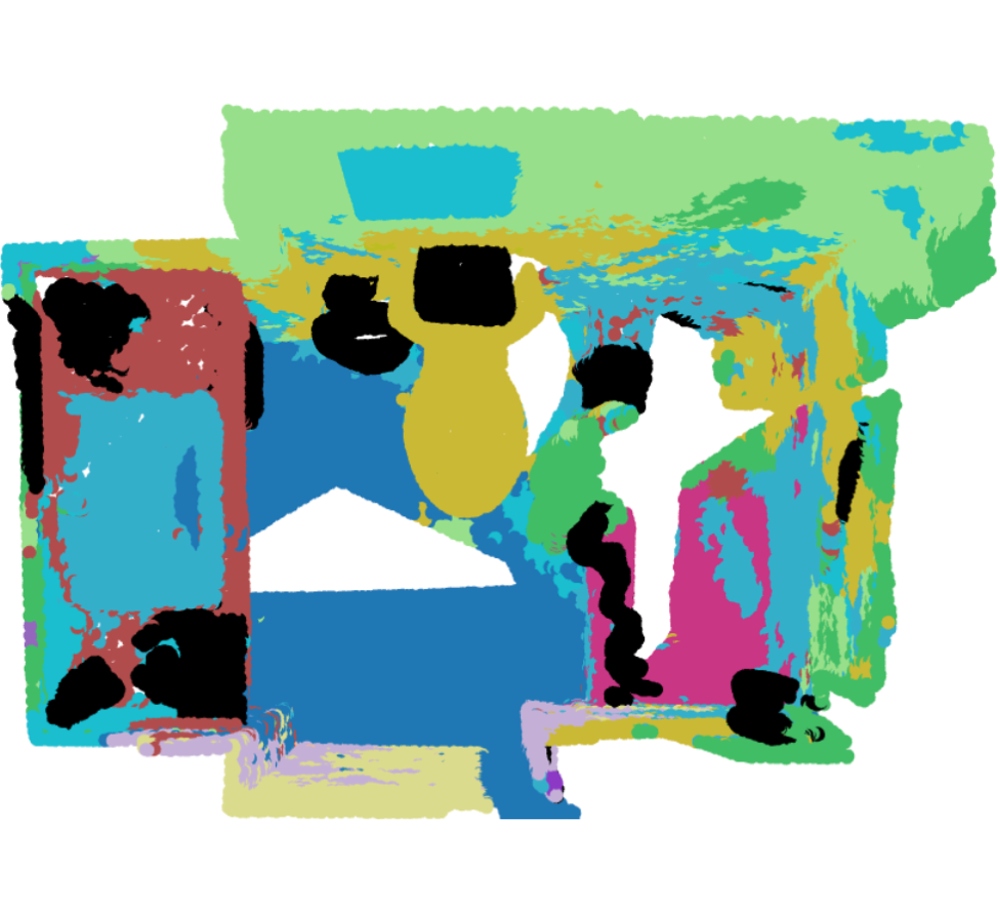}
  \subcaption*{Dr. Splat~\cite{jun2025dr}}
  \end{subfigure}
  \begin{subfigure}{0.22\linewidth}\includegraphics[width=1.0\linewidth]{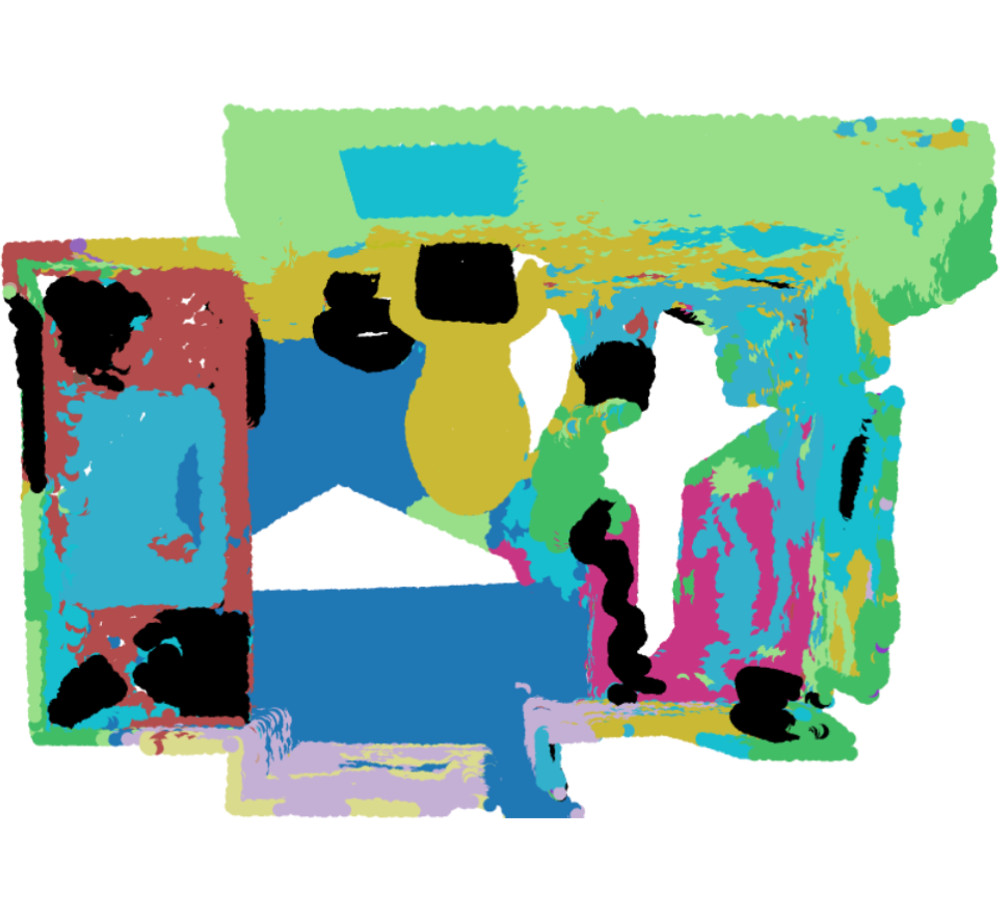}
  \subcaption*{Occam's LGS~\cite{cheng2024occam}}
  \end{subfigure}
  \begin{subfigure}{0.31\linewidth}\includegraphics[width=1.0\linewidth]{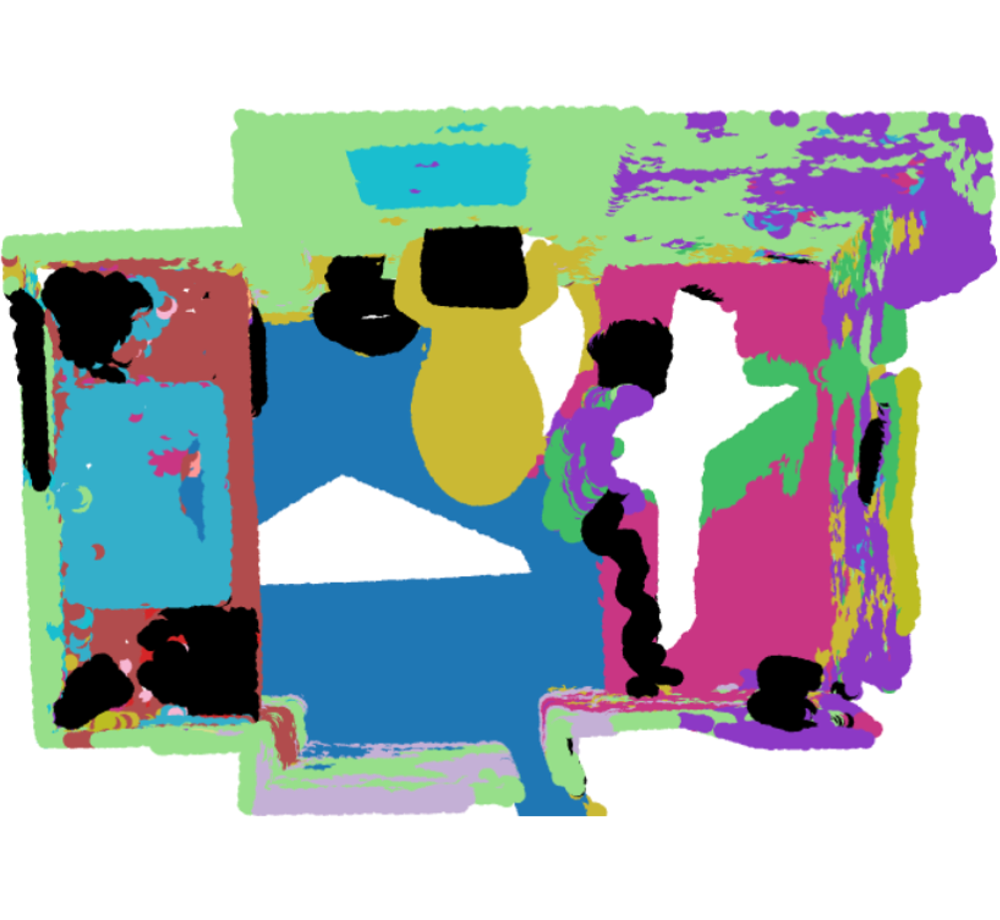}
  \subcaption*{\oursfast}
  \end{subfigure}
  \begin{subfigure}{0.31\linewidth}\includegraphics[width=1.0\linewidth]{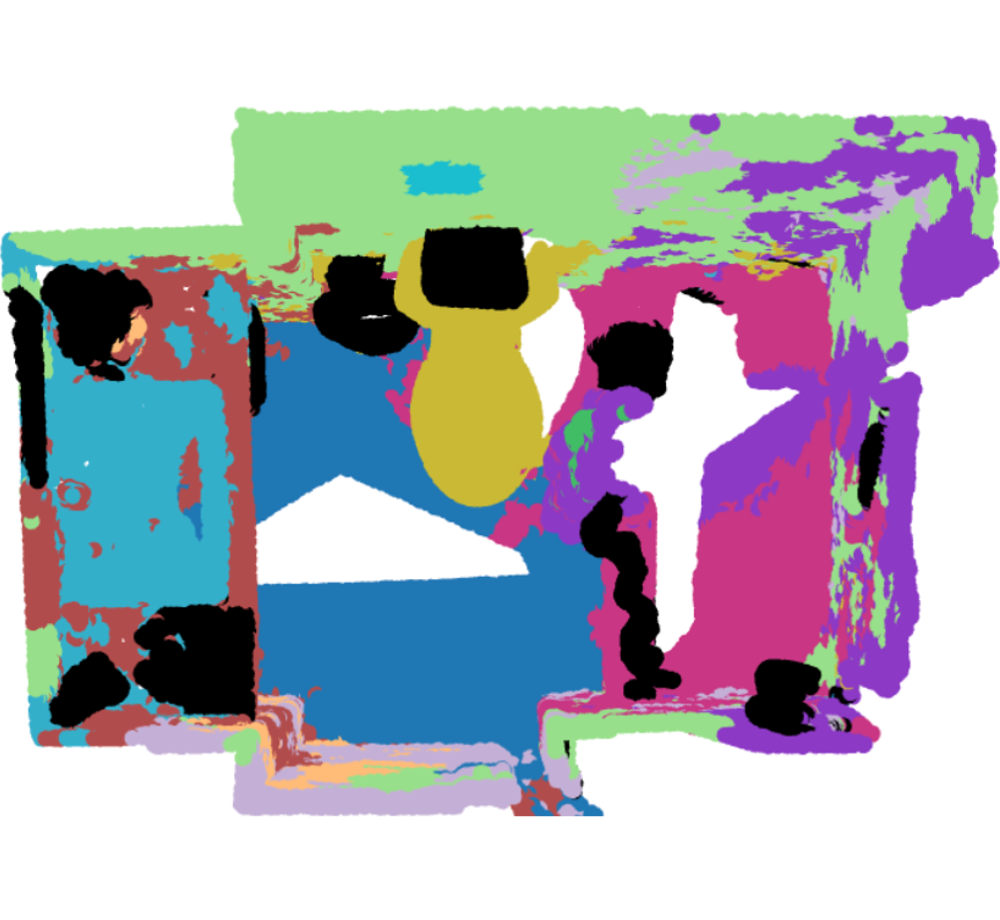}
  \subcaption*{\ours}
  \end{subfigure}
  \begin{subfigure}{0.31\linewidth}\includegraphics[width=1.0\linewidth]{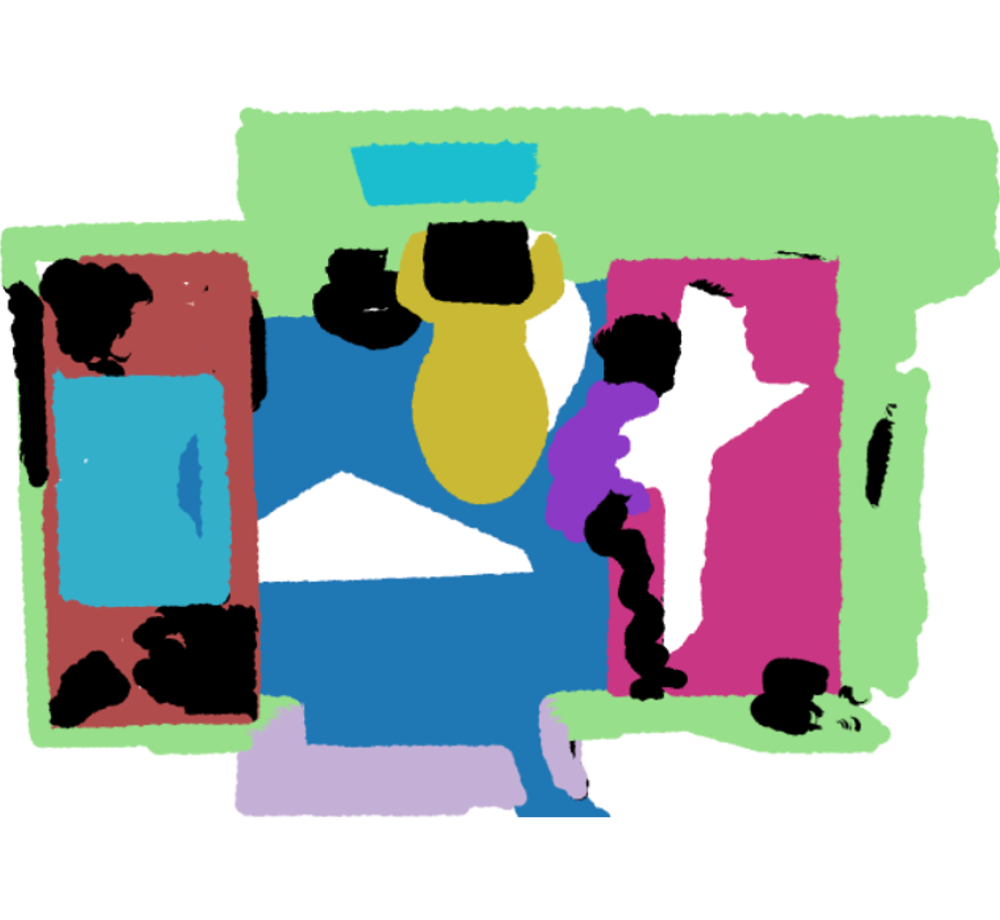}
  \subcaption*{GT}
  \end{subfigure}
  \vspace{30pt}

  \begin{subfigure}{0.22\linewidth}\includegraphics[width=1.0\linewidth]{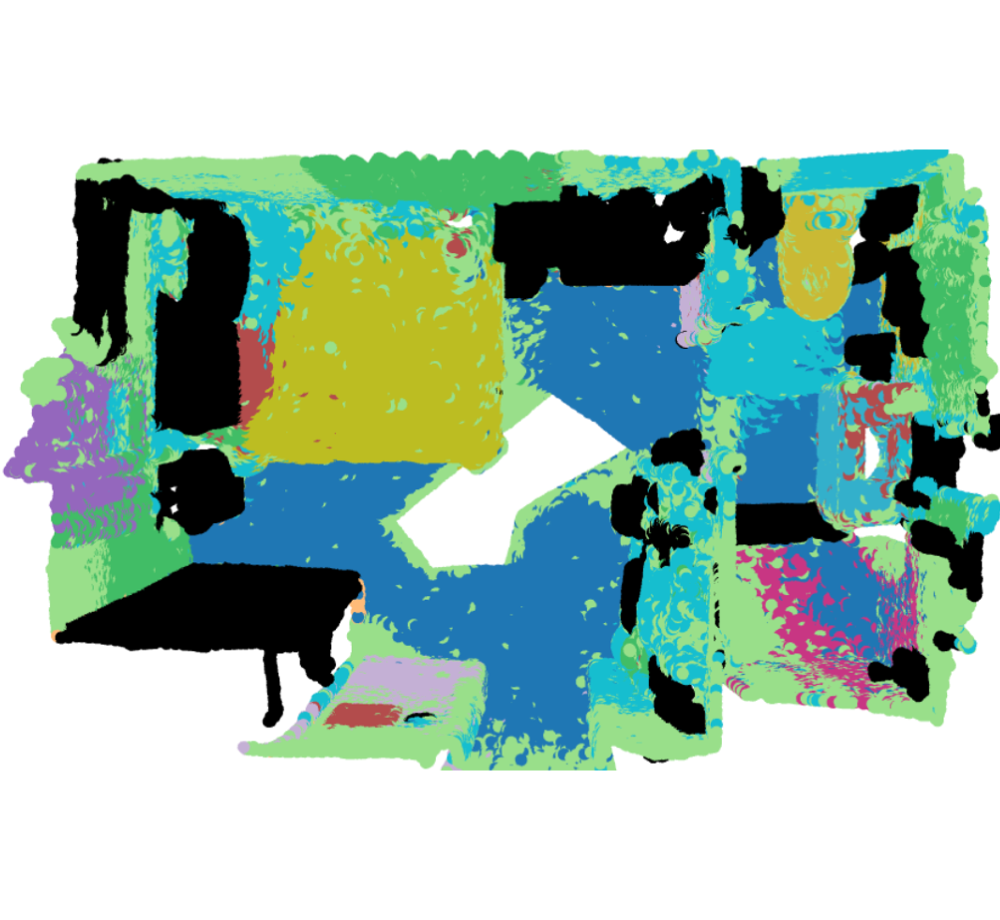}
  \subcaption*{OpenGaussian~\cite{wu2024opengaussian}}
  \end{subfigure}
  \begin{subfigure}{0.22\linewidth}\includegraphics[width=1.0\linewidth]{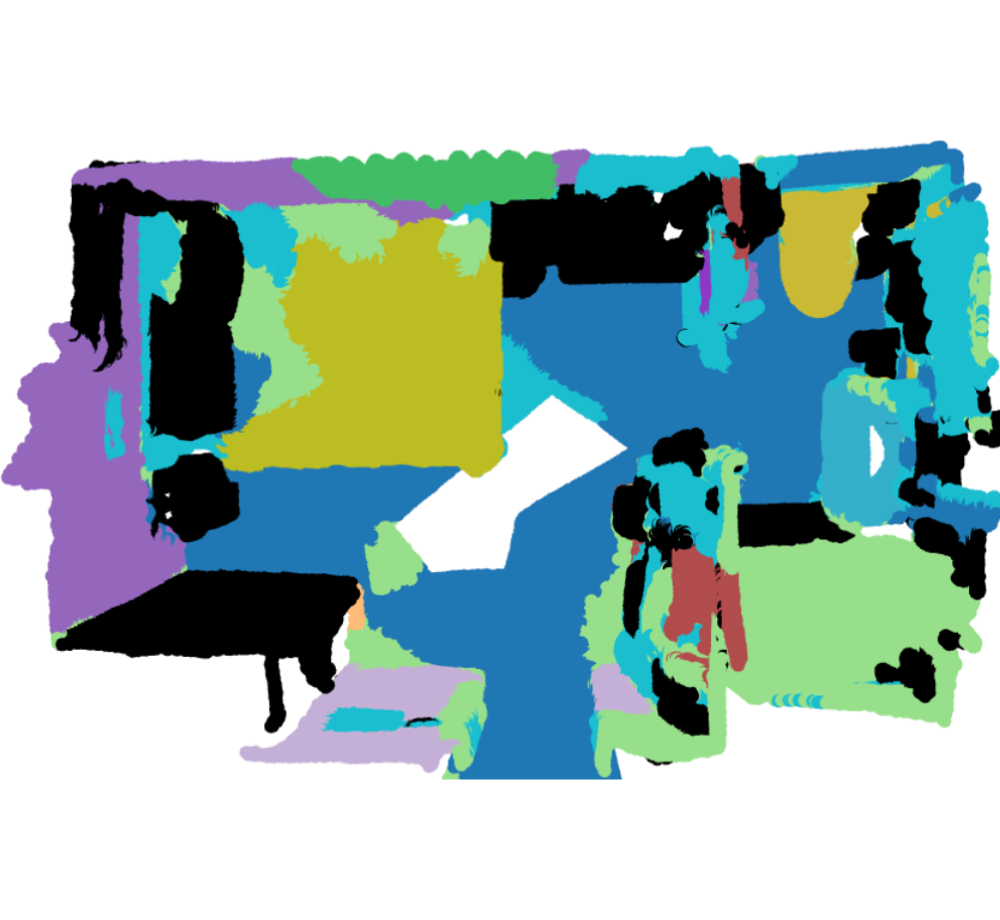}
  \subcaption*{InstanceGaussian~\cite{li2025instancegaussian}}
  \end{subfigure}
  \begin{subfigure}{0.22\linewidth}\includegraphics[width=1.0\linewidth]{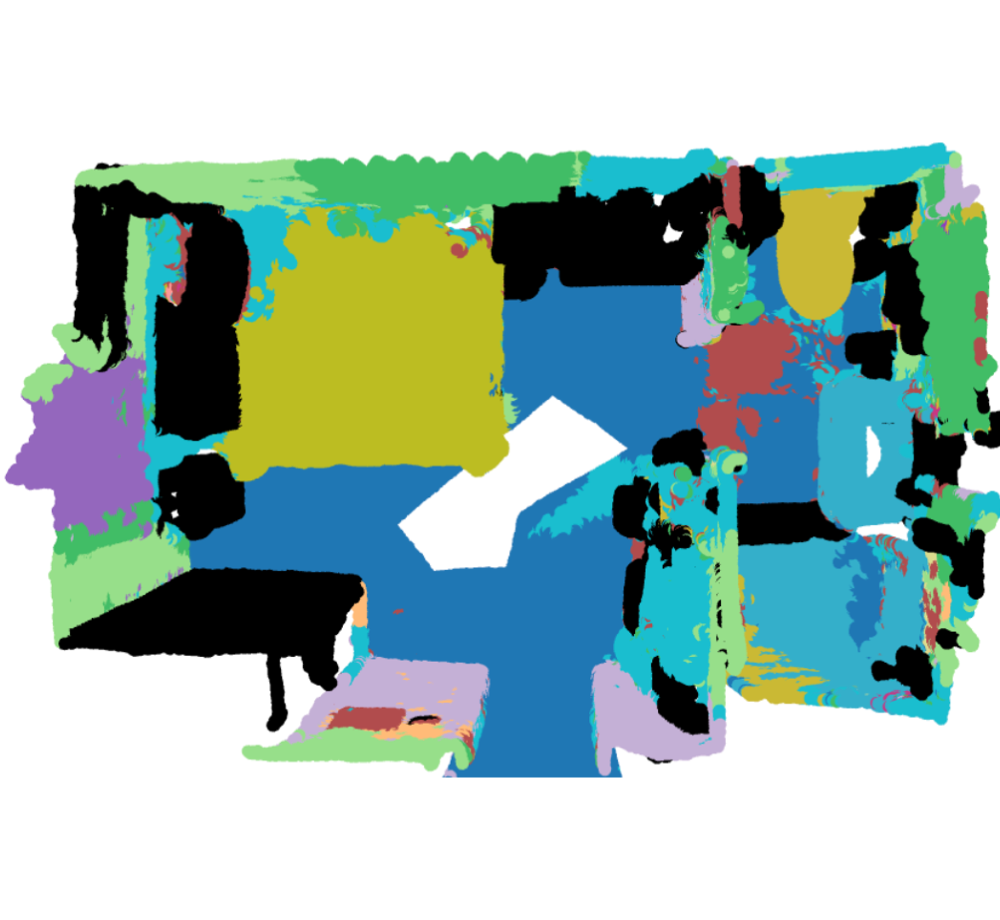}
  \subcaption*{Dr. Splat~\cite{jun2025dr}}
  \end{subfigure}
  \begin{subfigure}{0.22\linewidth}\includegraphics[width=1.0\linewidth]{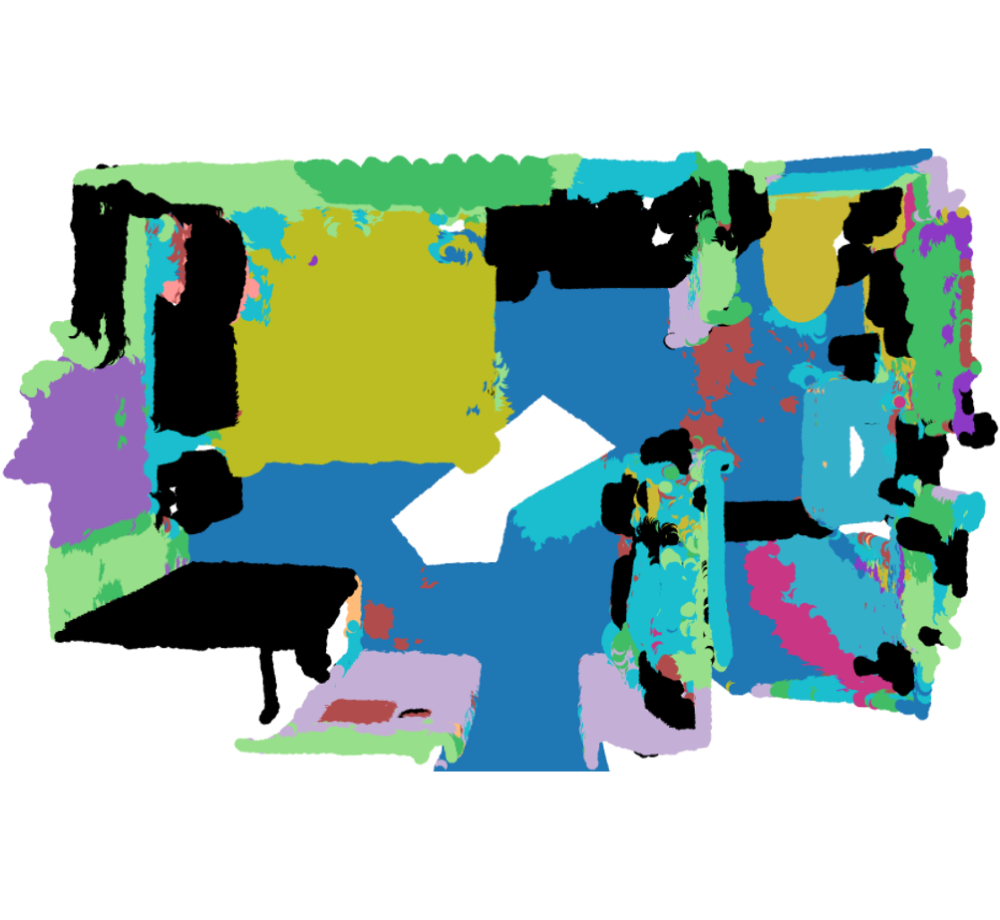}
  \subcaption*{Occam's LGS~\cite{cheng2024occam}}
  \end{subfigure}
  \begin{subfigure}{0.31\linewidth}\includegraphics[width=1.0\linewidth]{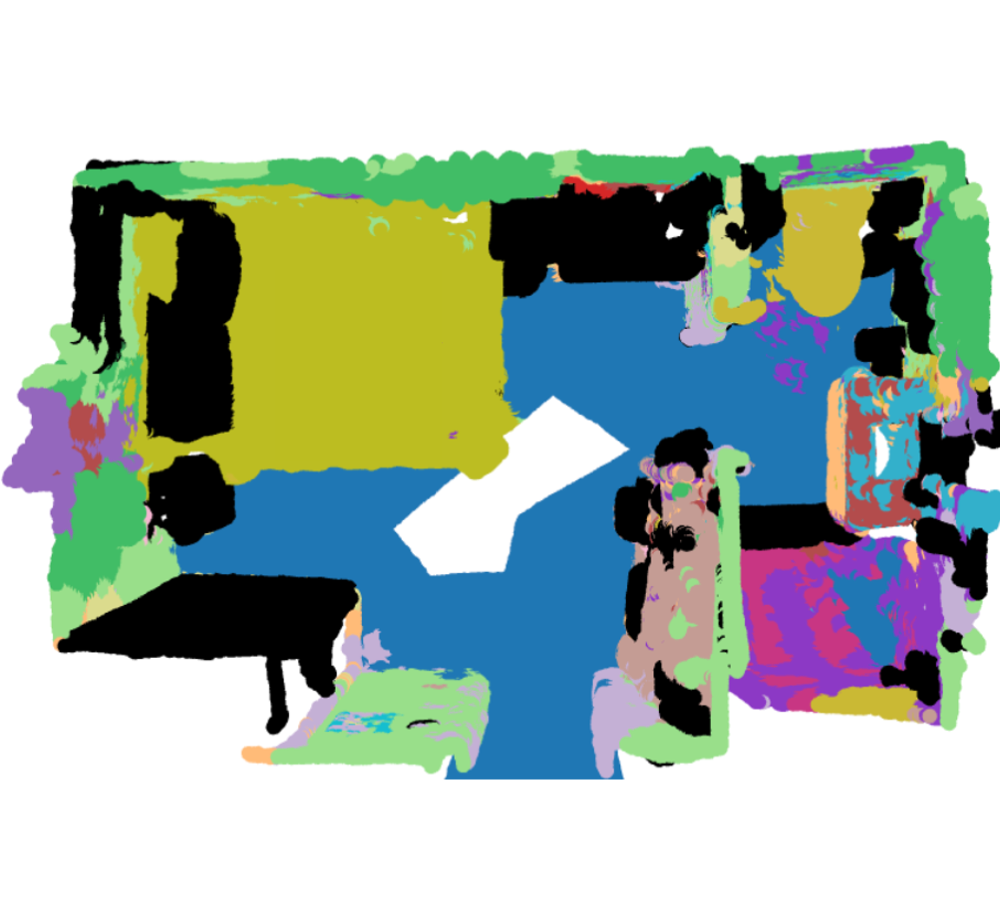}
  \subcaption*{\oursfast}
  \end{subfigure}
  \begin{subfigure}{0.31\linewidth}\includegraphics[width=1.0\linewidth]{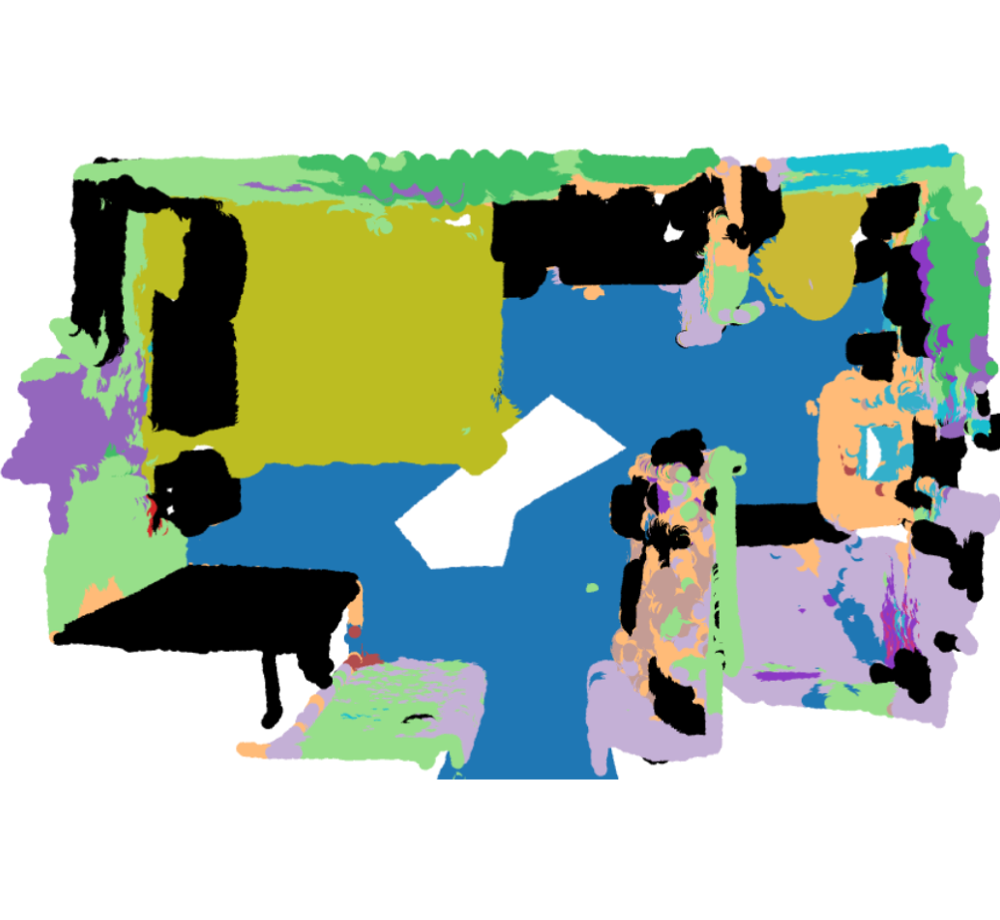}
  \subcaption*{\ours}
  \end{subfigure}
  \begin{subfigure}{0.31\linewidth}\includegraphics[width=1.0\linewidth]{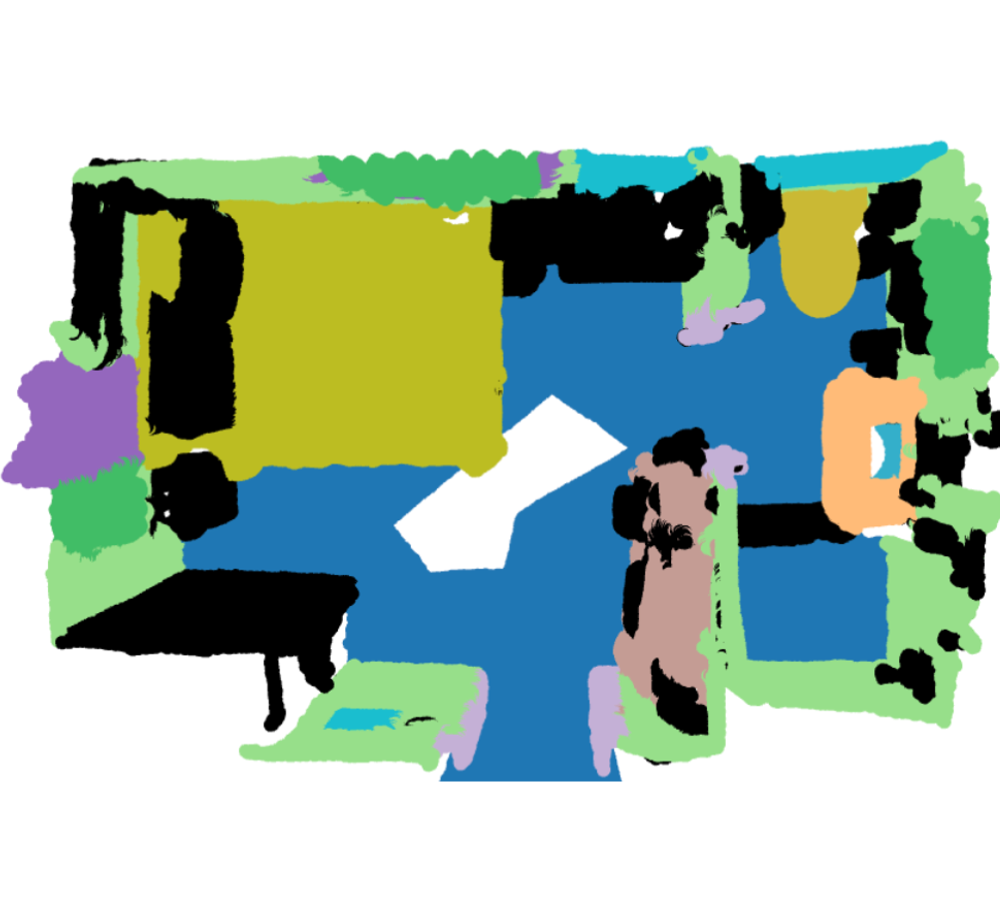}
  \subcaption*{GT}
  \end{subfigure}

  \caption{\textbf{More qualitative comparisons on 3D semantic segmentaiton - (2)}}
  \label{fig:supple_qualitative_3d_2}
\end{figure*}

\vspace{1mm}
\noindent \textbf{Qualitative results on 3D Semantic Segmentation.} Fig.~\ref{fig:supple_qualitative_3d_1} and Fig.~\ref{fig:supple_qualitative_3d_2} present the additional qualitative comparison on 3D semantic segmentation. Our \ours and \oursfast output more clear segmentation mask with more accurate semantic classification compared to the \textit{3D} baselines~\citep{wu2024opengaussian, li2025instancegaussian, jun2025dr, cheng2024occam}.

\begin{figure*}[t]
\captionsetup[subfigure]{}
  \centering  
  \begin{subfigure}{0.195\linewidth}\includegraphics[width=1.0\linewidth]{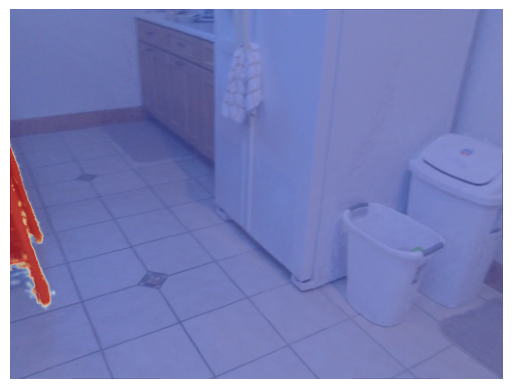}
  \end{subfigure}
  \begin{subfigure}{0.195\linewidth}\includegraphics[width=1.0\linewidth]{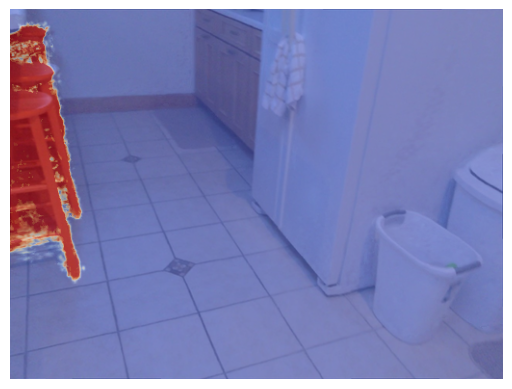}
  \end{subfigure}
  \begin{subfigure}{0.195\linewidth}\includegraphics[width=1.0\linewidth]{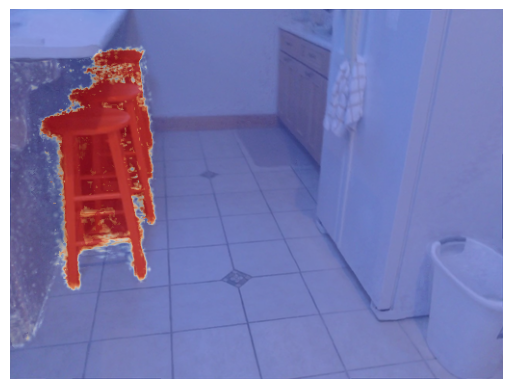}
  \end{subfigure}
  \begin{subfigure}{0.195\linewidth}\includegraphics[width=1.0\linewidth]{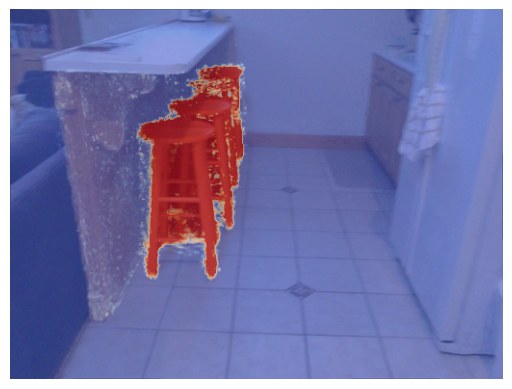}
  \end{subfigure}
  \begin{subfigure}{0.195\linewidth}\includegraphics[width=1.0\linewidth]{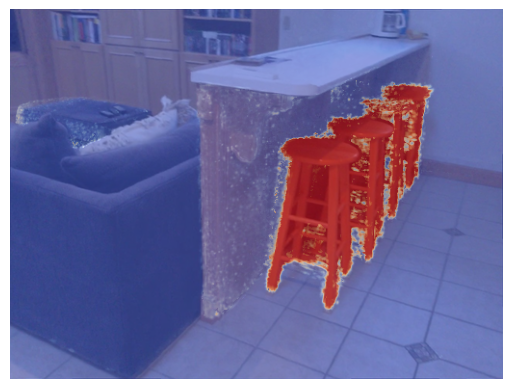}
  \end{subfigure}
  \begin{subfigure}{0.195\linewidth}\includegraphics[width=1.0\linewidth]{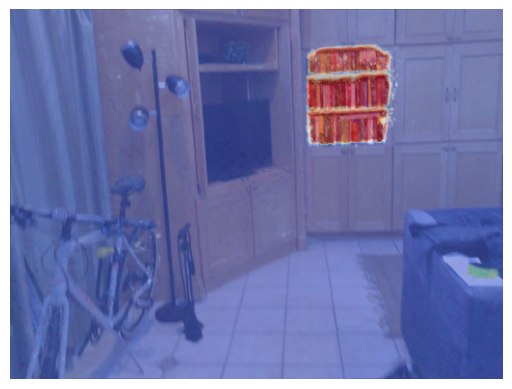}
  \end{subfigure}
  \begin{subfigure}{0.195\linewidth}\includegraphics[width=1.0\linewidth]{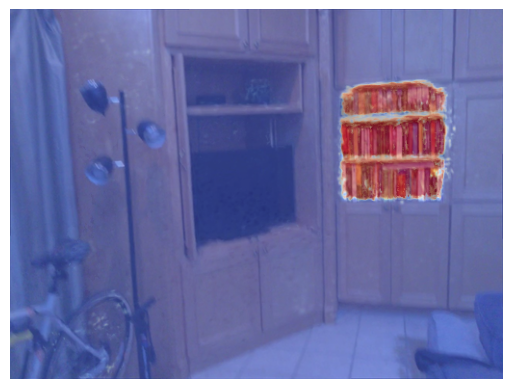}
  \end{subfigure}
  \begin{subfigure}{0.195\linewidth}\includegraphics[width=1.0\linewidth]{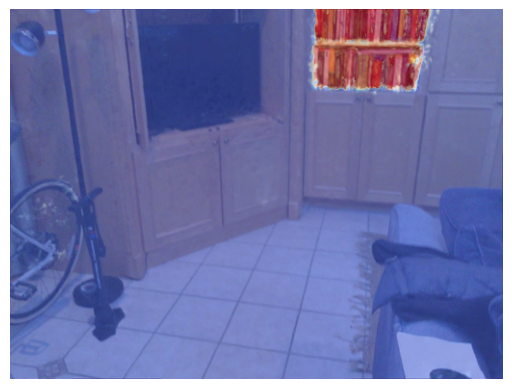}
  \end{subfigure}
  \begin{subfigure}{0.195\linewidth}\includegraphics[width=1.0\linewidth]{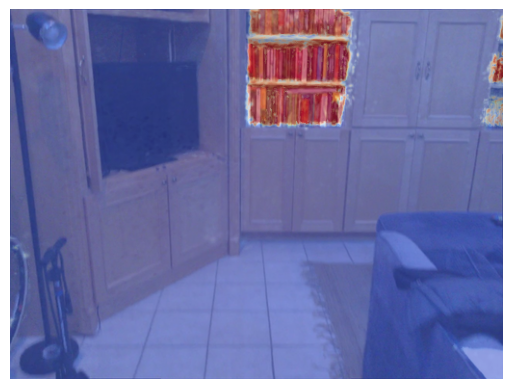}
  \end{subfigure}
  \begin{subfigure}{0.195\linewidth}\includegraphics[width=1.0\linewidth]{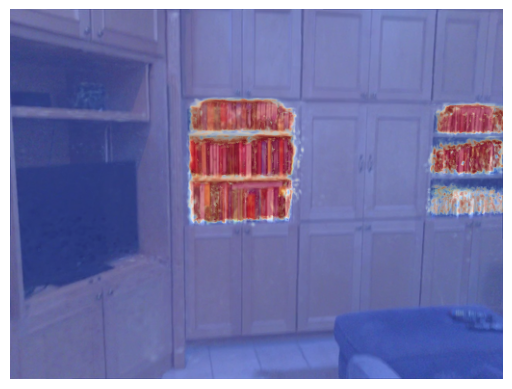}
  \end{subfigure}
  \caption{\textbf{Qualitative results on 2D-rendered object search of our \ours.}}
  \label{fig:supple_qualitative_2dseg}
\end{figure*}

\vspace{1mm}
\noindent \textbf{Qualitative results on 2D-rendered object search.} 
Fig.~\ref{fig:supple_qualitative_2dseg} showcases the additional visualizations for 2D-rendered object search with our \ours. Text queries ``\textit{stool}'' and ``\textit{book}'' are given to each row. Our model outputs multi-view consistent segmentation results by exploiting the 3DGS representation.

\begin{figure*}[t]
\captionsetup[subfigure]{}
  \centering  
  \begin{subfigure}{0.16\linewidth}\includegraphics[width=1.0\linewidth]{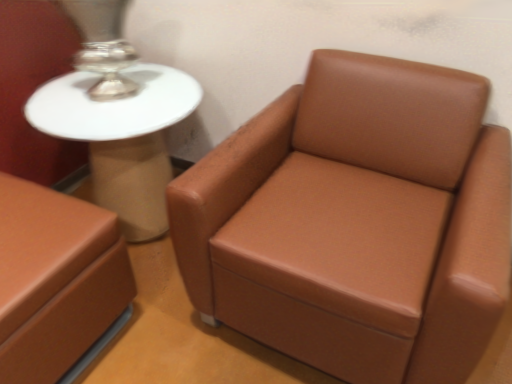}
  \end{subfigure}
  \begin{subfigure}{0.16\linewidth}\includegraphics[width=1.0\linewidth]{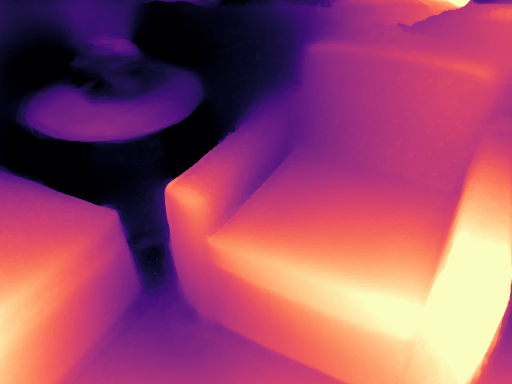}
  \end{subfigure}
  \begin{subfigure}{0.16\linewidth}\includegraphics[width=1.0\linewidth]{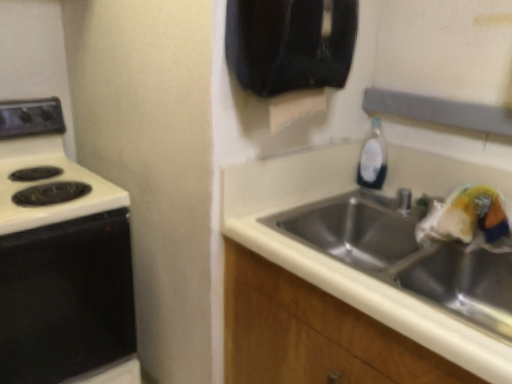}
  \end{subfigure}
  \begin{subfigure}{0.16\linewidth}\includegraphics[width=1.0\linewidth]{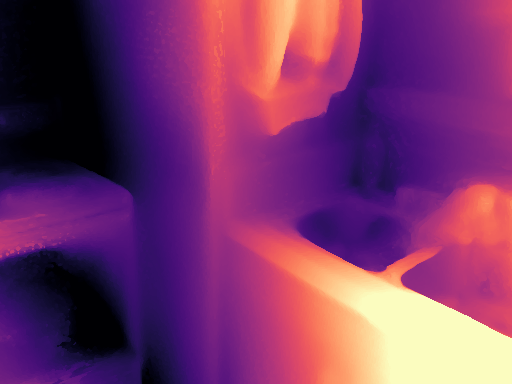}
  \end{subfigure}
  \begin{subfigure}{0.16\linewidth}\includegraphics[width=1.0\linewidth]{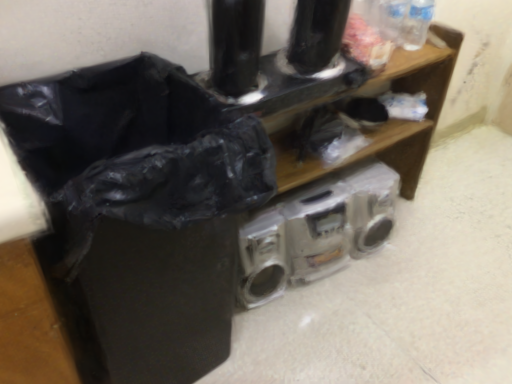}
  \end{subfigure}
  \begin{subfigure}{0.16\linewidth}\includegraphics[width=1.0\linewidth]{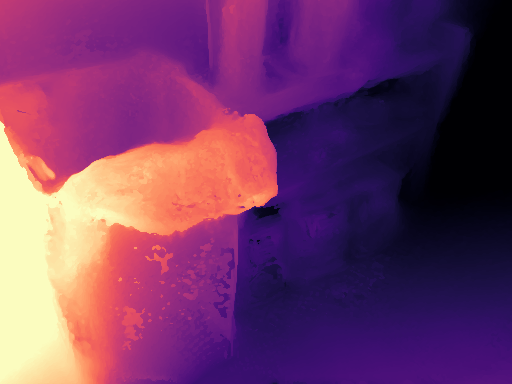}
  \end{subfigure}
  \begin{subfigure}{0.16\linewidth}\includegraphics[width=1.0\linewidth]{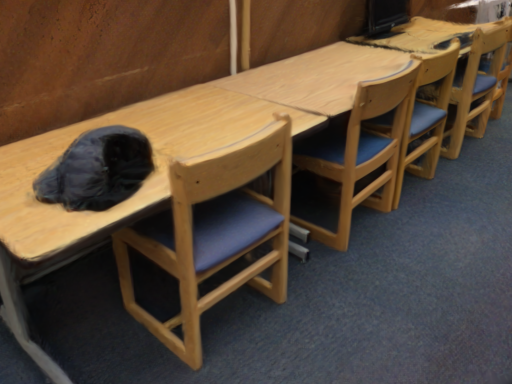}
  \end{subfigure}
  \begin{subfigure}{0.16\linewidth}\includegraphics[width=1.0\linewidth]{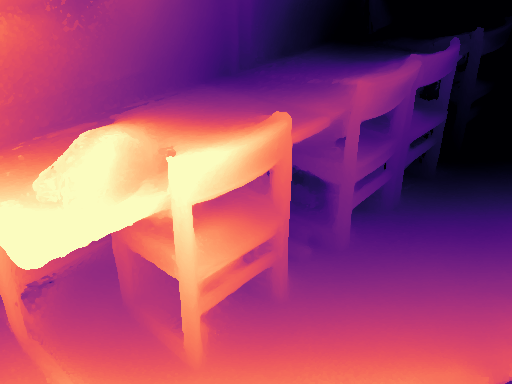}
  \end{subfigure}
  \begin{subfigure}{0.16\linewidth}\includegraphics[width=1.0\linewidth]{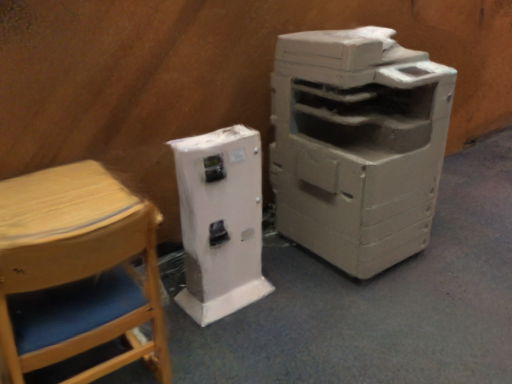}
  \end{subfigure}
  \begin{subfigure}{0.16\linewidth}\includegraphics[width=1.0\linewidth]{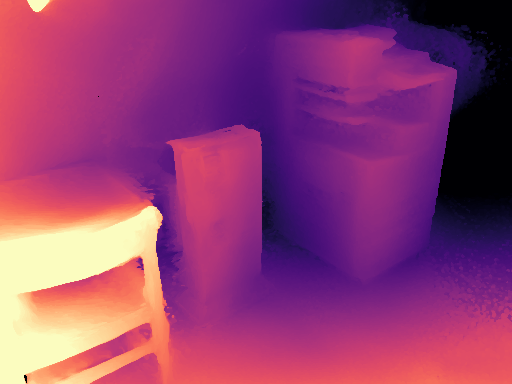}
  \end{subfigure}
  \begin{subfigure}{0.16\linewidth}\includegraphics[width=1.0\linewidth]{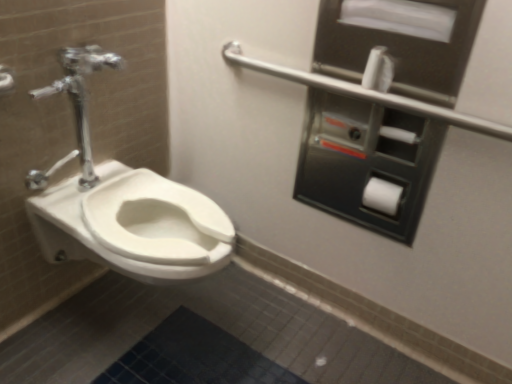}
  \end{subfigure}
  \begin{subfigure}{0.16\linewidth}\includegraphics[width=1.0\linewidth]{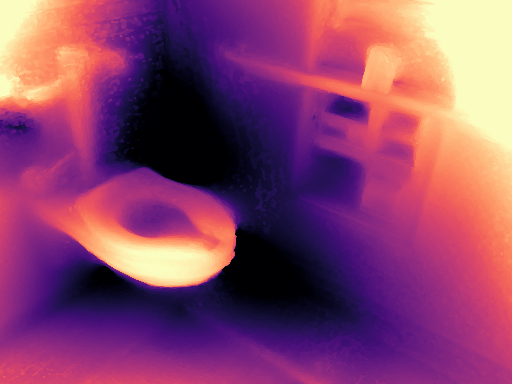}
  \end{subfigure}
  \begin{subfigure}{0.16\linewidth}\includegraphics[width=1.0\linewidth]{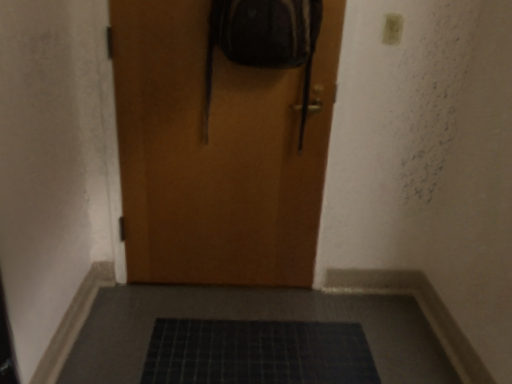}
  \end{subfigure}
  \begin{subfigure}{0.16\linewidth}\includegraphics[width=1.0\linewidth]{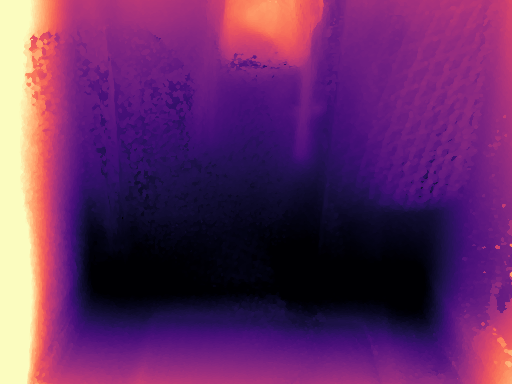}
  \end{subfigure}
  \begin{subfigure}{0.16\linewidth}\includegraphics[width=1.0\linewidth]{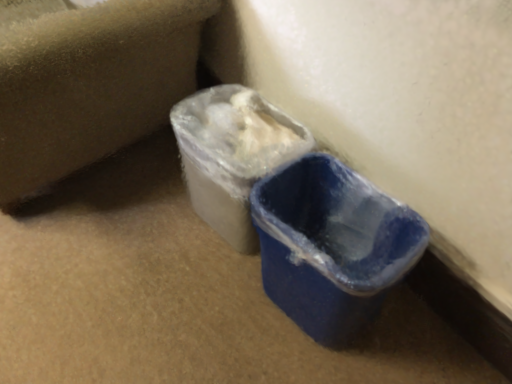}
  \end{subfigure}
  \begin{subfigure}{0.16\linewidth}\includegraphics[width=1.0\linewidth]{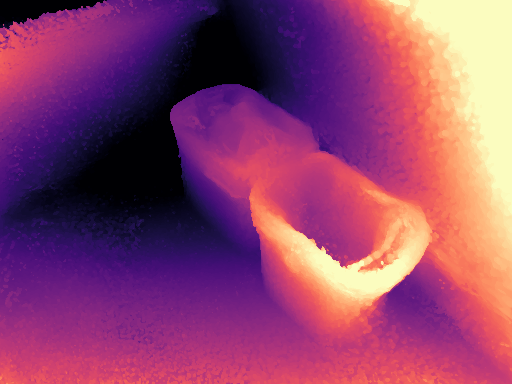}
  \end{subfigure}
  \begin{subfigure}{0.16\linewidth}\includegraphics[width=1.0\linewidth]{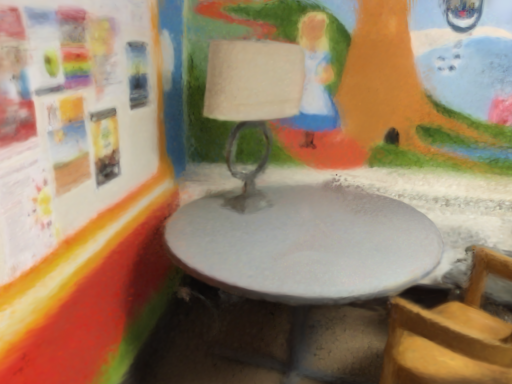}
  \end{subfigure}
  \begin{subfigure}{0.16\linewidth}\includegraphics[width=1.0\linewidth]{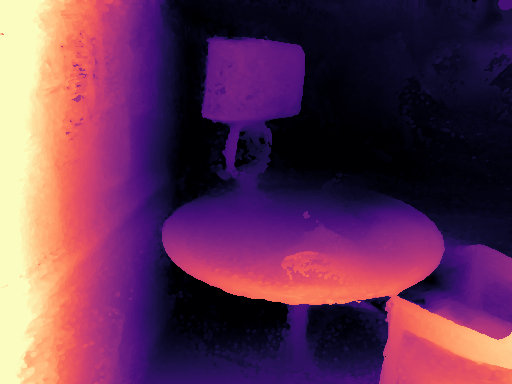}
  \end{subfigure}
  \begin{subfigure}{0.16\linewidth}\includegraphics[width=1.0\linewidth]{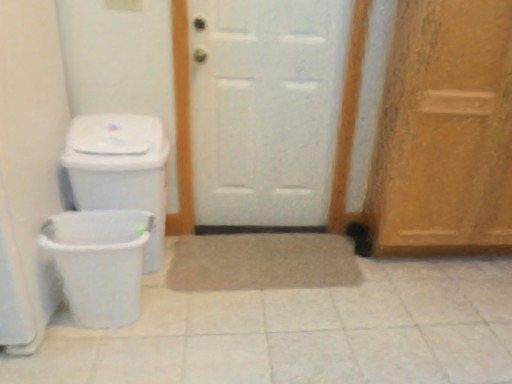}
  \end{subfigure}
  \begin{subfigure}{0.16\linewidth}\includegraphics[width=1.0\linewidth]{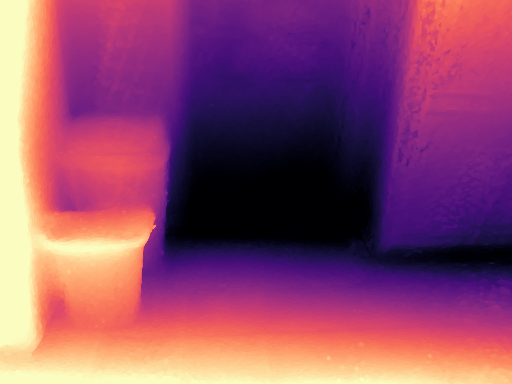}
  \end{subfigure}
  \begin{subfigure}{0.16\linewidth}\includegraphics[width=1.0\linewidth]{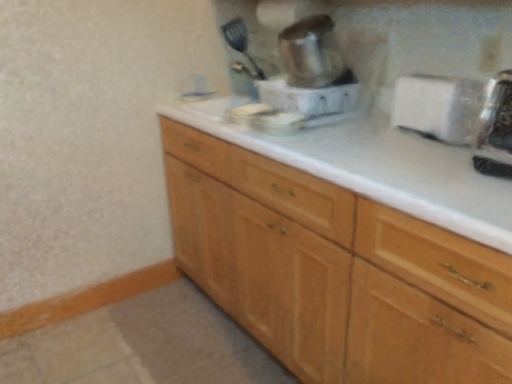}
  \end{subfigure}
  \begin{subfigure}{0.16\linewidth}\includegraphics[width=1.0\linewidth]{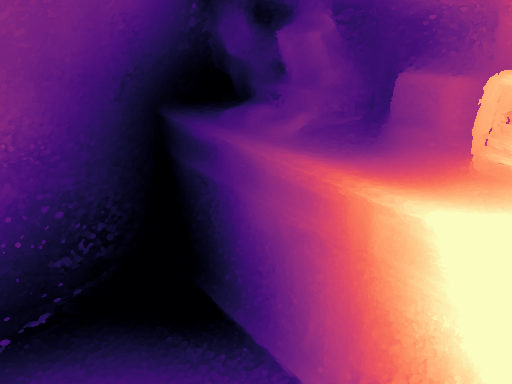}
  \end{subfigure}
  \begin{subfigure}{0.16\linewidth}\includegraphics[width=1.0\linewidth]{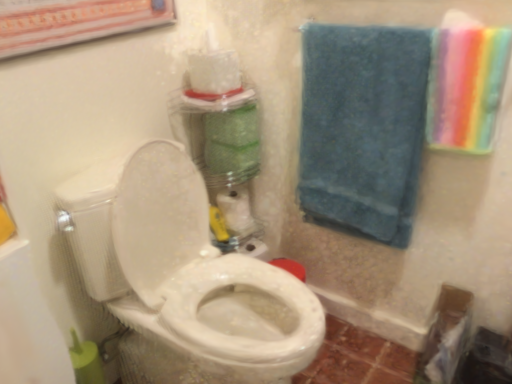}
  \end{subfigure}
  \begin{subfigure}{0.16\linewidth}\includegraphics[width=1.0\linewidth]{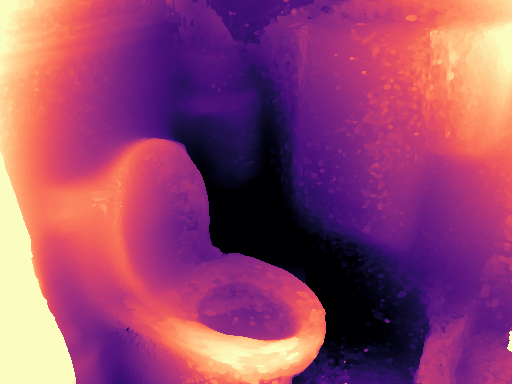}
  \end{subfigure}
  \begin{subfigure}{0.16\linewidth}\includegraphics[width=1.0\linewidth]{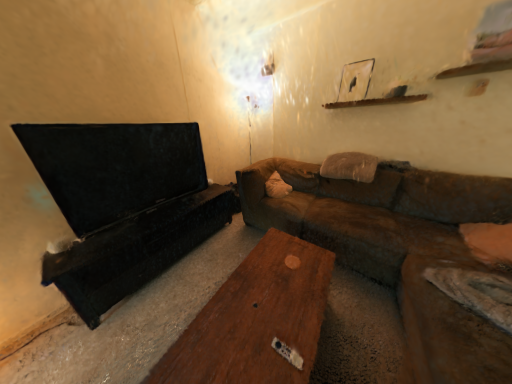}
  \end{subfigure}
  \begin{subfigure}{0.16\linewidth}\includegraphics[width=1.0\linewidth]{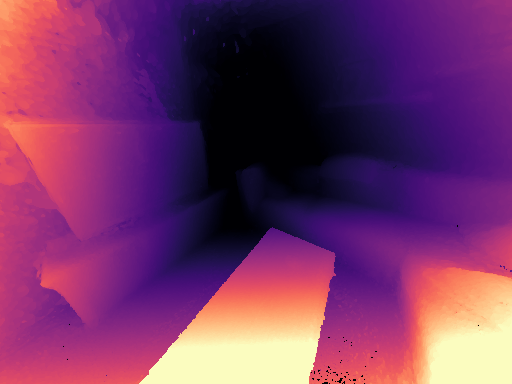}
  \end{subfigure}
  \begin{subfigure}{0.16\linewidth}\includegraphics[width=1.0\linewidth]{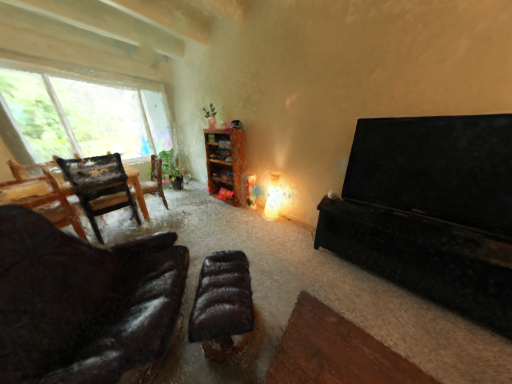}
  \end{subfigure}
  \begin{subfigure}{0.16\linewidth}\includegraphics[width=1.0\linewidth]{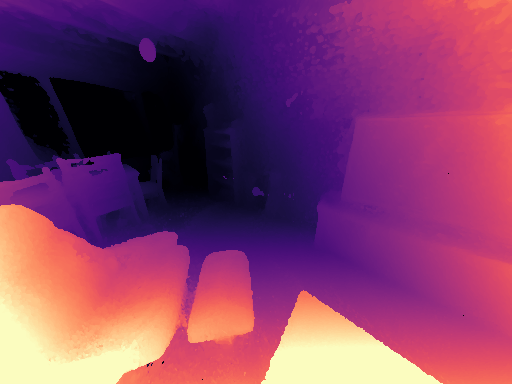}
  \end{subfigure}
  \begin{subfigure}{0.16\linewidth}\includegraphics[width=1.0\linewidth]{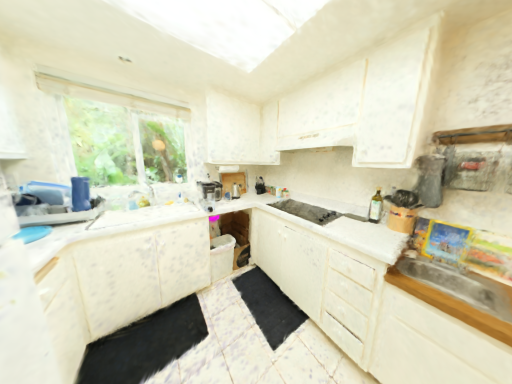}
  \end{subfigure}
  \begin{subfigure}{0.16\linewidth}\includegraphics[width=1.0\linewidth]{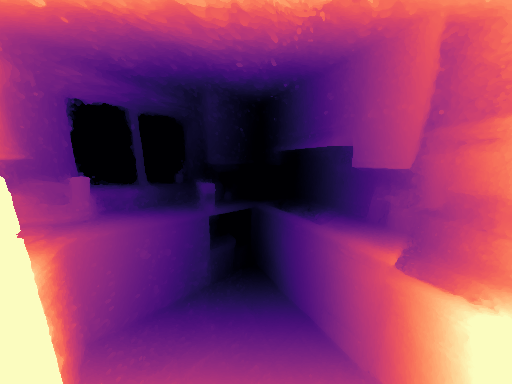}
  \end{subfigure}
  \begin{subfigure}{0.16\linewidth}\includegraphics[width=1.0\linewidth]{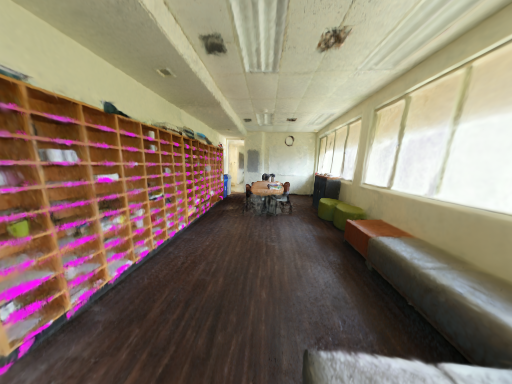}
  \end{subfigure}
  \begin{subfigure}{0.16\linewidth}\includegraphics[width=1.0\linewidth]{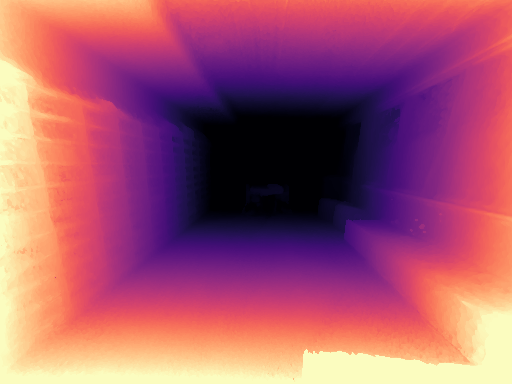}
  \end{subfigure}
  \begin{subfigure}{0.16\linewidth}\includegraphics[width=1.0\linewidth]{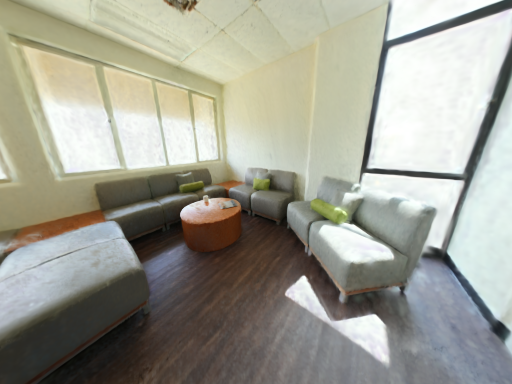}
  \end{subfigure}
  \begin{subfigure}{0.16\linewidth}\includegraphics[width=1.0\linewidth]{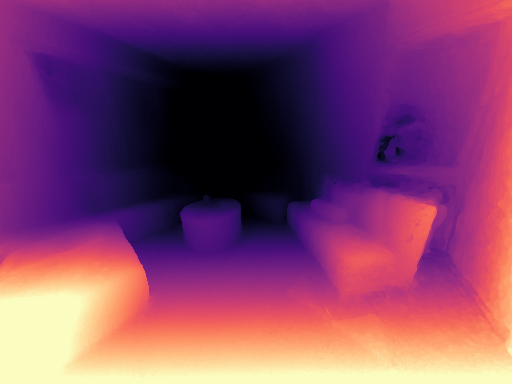}
  \end{subfigure}
  \begin{subfigure}{0.16\linewidth}\includegraphics[width=1.0\linewidth]{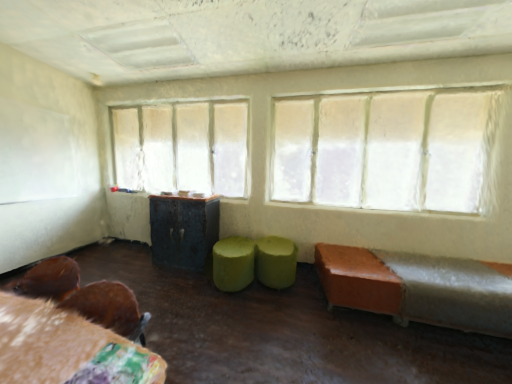}
  \end{subfigure}
  \begin{subfigure}{0.16\linewidth}\includegraphics[width=1.0\linewidth]{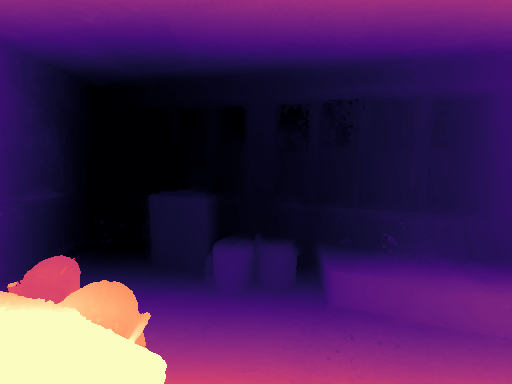}
  \end{subfigure}
  \begin{subfigure}{0.16\linewidth}\includegraphics[width=1.0\linewidth]{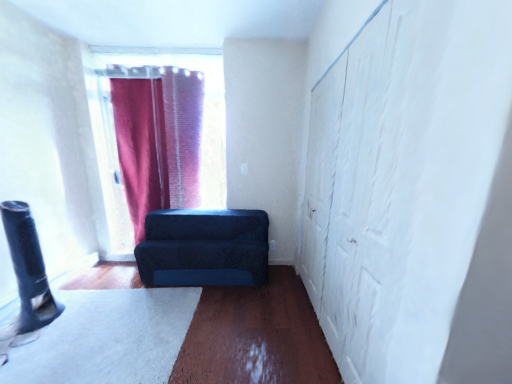}
  \end{subfigure}
  \begin{subfigure}{0.16\linewidth}\includegraphics[width=1.0\linewidth]{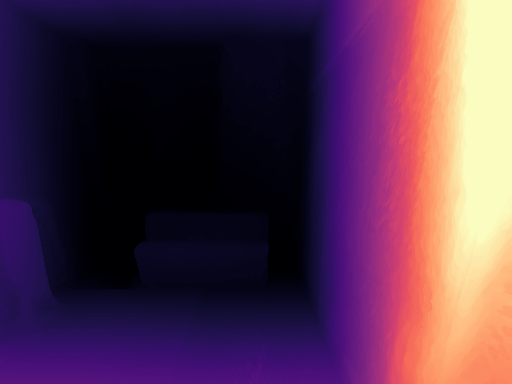}
  \end{subfigure}
  \begin{subfigure}{0.16\linewidth}\includegraphics[width=1.0\linewidth]{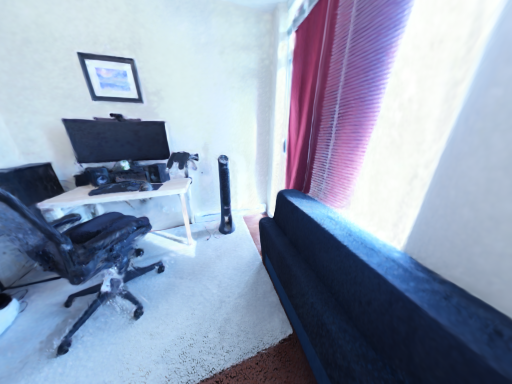}
  \end{subfigure}
  \begin{subfigure}{0.16\linewidth}\includegraphics[width=1.0\linewidth]{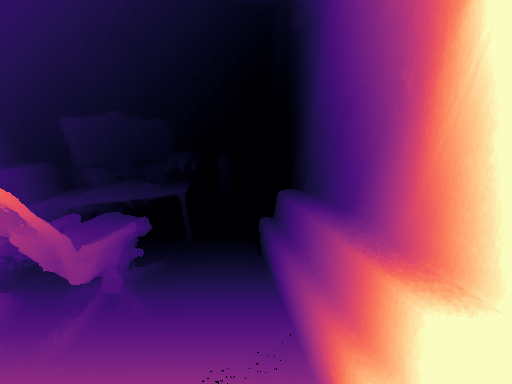}
  \end{subfigure}
  \begin{subfigure}{0.16\linewidth}\includegraphics[width=1.0\linewidth]{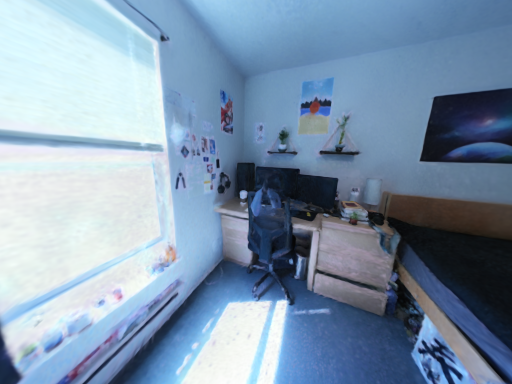}
  \end{subfigure}
  \begin{subfigure}{0.16\linewidth}\includegraphics[width=1.0\linewidth]{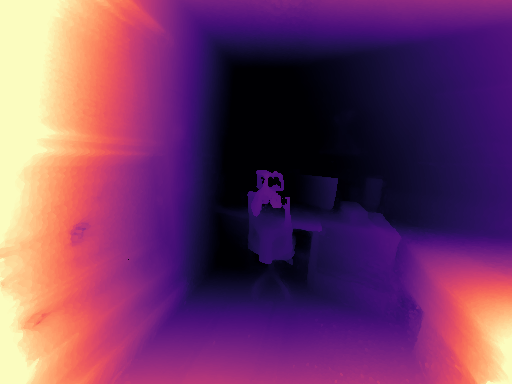}
  \end{subfigure}
  \caption{\textbf{Qualitative results on novel-view synthesis and depth rendering of our \ours.}}
  \label{fig:supple_qualitative_nvs}
\end{figure*}

\vspace{1mm}
\noindent \textbf{Qualitative results on novel-view synthesis.} Fig.~\ref{fig:supple_qualitative_nvs} exhibits the novel-view synthesis results and rendered depths of our \ours. It supports novel-view rendering with high fidelity across the entire scene.

\vspace{1mm}
\noindent \textbf{Video visulization.} Video visualizations in project website demonstrates the online reconstruction process of semantic 3DGS with our \oursfast in Bird's-Eye View. It shows following key aspects of our framework: \textbf{\textit{1) Near real-time reconstruction:}} \ours shows 5-6 FPS of per-frame processing time where it can effectively synchronize the semantic reconstruction process to its online exploration. \textbf{\textit{2) Online 3D perception with free-form language:}} Our \oursfast can localize the 3D objects based on the free-form language along its exploration. For example, video shows that \oursfast progressively detects the ``\textit{guitar}'' which is related to the given text prompt ``\textit{I wanna hear the music}''. Interestingly, our model supports the semantic refinement with re-exploration where it corrects the wrong semantic by exploring the same regions and collecting more views, as shown in the video. Our online fusion algorithm of sparse coefficient field enables this refinement since it accumulates the new evidences from the incoming images into the index and weight cache. Furthermore, it always keeps the top 5 entries with the highest confidence scores at each fusion step, effectively filtering out low-quality semantics signals along the exploration. Video further showcases several 3D localization examples using free-form languages. For instance, \oursfast localizes ``\textit{chair}'', ``\textit{sofa}'' and ``\textit{stool}'' together when queried with the text prompt, ``\textit{where can I sit?}''. \textbf{\textit{3) Supporting diverse 3D perception tasks:}} We also visualize the rendered RGB images along the camera trajectory as well as 2D-rendered PCA visualizations based on the CLIP features of each Gaussian. It demonstrates that our framework supports diverse perception tasks such as RGB reconstruction and semantic understanding in both 2D and 3D modalities.

\end{document}